%% file: main.tex
\pdfoutput=1
\documentclass[12pt,a4paper]{article}

\usepackage[margin=2cm]{geometry}
\usepackage[T1]{fontenc}
\usepackage{lmodern}
\usepackage{amsmath,amssymb,bm}
\usepackage{graphicx}
\usepackage{booktabs}
\usepackage{arydshln}
\usepackage{siunitx}
\usepackage[numbers,sort&compress]{natbib}
\usepackage{xcolor}
\usepackage{tikz}
\usetikzlibrary{positioning,arrows.meta,calc,patterns,decorations.pathreplacing}
\usepackage{pgfplots}
\pgfplotsset{compat=1.18}
\usepackage{algorithm}
\usepackage{algpseudocode}
\usepackage{subcaption}
\usepackage{caption}
\usepackage[section]{placeins}
\usepackage{sectsty}
\allsectionsfont{\raggedright}
\usepackage[colorlinks=true,linkcolor=blue,citecolor=blue,urlcolor=blue]{hyperref}
\usepackage{xurl}
\newcommand{\doi}[1]{\href{https://doi.org/#1}{doi:\,\nolinkurl{#1}}}

\graphicspath{{figures/}}

\definecolor{cblue}{HTML}{1F77B4}
\definecolor{cred}{HTML}{D62728}
\definecolor{cgreen}{HTML}{2CA02C}
\definecolor{corange}{HTML}{FF7F0E}
\definecolor{cpurple}{HTML}{9467BD}
\pgfplotsset{
  paperaxis/.style={
    tick align=inside,
    tick label style={font=\footnotesize},
    label style={font=\small},
    legend style={font=\footnotesize, draw=none, fill=none},
    legend cell align=left,
  }
}

\tikzset{
  spec/.style={fill=cblue!6, draw=black!70, line width=0.7pt},
  crackline/.style={cred, line width=1.2pt, line cap=round},
  clamphatch/.style={pattern=north east lines, pattern color=black!55},
  loadarrow/.style={-{Stealth[length=2.0mm]}, cred, line width=0.8pt},
  heatarrow/.style={-{Stealth[length=2.0mm]}, corange, line width=0.8pt},
  dimline/.style={{Stealth[length=1.5mm]}-{Stealth[length=1.5mm]},
                  black!55, line width=0.5pt},
  dimfont/.style={font=\footnotesize, text=black!75},
}
\input{macros}

\title{A mesh-free deep energy method for phase-field modeling of
thermally driven brittle fracture}
\author{%
  Han Zhang$^{\mathrm{a}}$\thanks{E-mail: han.zhang7@unsw.edu.au}~,
  Mehrisadat Makki Alamdari$^{\mathrm{a}}$,
  Babak Shahbodagh$^{\mathrm{a}}$,\\
  Mohammad Vahab$^{\mathrm{b}}$,
  Cosmin Anitescu$^{\mathrm{c}}$,
  Timon Rabczuk$^{\mathrm{c}}$,\\
  Elena Atroshchenko$^{\mathrm{a}}$\thanks{Corresponding author.
  E-mail: e.atroshchenko@unsw.edu.au}\\[1.2ex]
  {\normalsize $^{\mathrm{a}}$University of New South Wales, Australia}\\
  {\normalsize $^{\mathrm{b}}$Central Queensland University,
  Melbourne, Australia}\\
  {\normalsize $^{\mathrm{c}}$Bauhaus-Universit\"at Weimar, Germany}}
\date{}

\begin{document}
\maketitle

\input{sections/S0_abstract}
\clearpage
\input{sections/S1_intro}
\input{sections/S2_formulation}
\input{sections/S3_method}

\input{sections/S4_results}

\input{sections/S6_conclusions}

\appendix
\renewcommand{\thesection}{Appendix~\Alph{section}}
\renewcommand{\thesubsection}{\Alph{section}.\arabic{subsection}}
\counterwithin*{table}{section}
\counterwithin*{figure}{section}
\renewcommand{\thetable}{\Alph{section}.\arabic{table}}
\renewcommand{\thefigure}{\Alph{section}.\arabic{figure}}
\input{sections/SA_appendix}

\FloatBarrier
\bibliographystyle{unsrtnat}
\bibliography{refs}

\end{document}

%% file: macros.tex
\newcommand{\ub}{\bm{u}}                          %
\newcommand{\xb}{\bm{x}}                          %
\newcommand{\xib}{\bm{\xi}}                       %
\newcommand{\pf}{\phi}                            %

\newcommand{\straintensor}{\bm{\varepsilon}}
\newcommand{\stresstensor}{\bm{\sigma}}
\newcommand{\Jgeo}{\bm{J}}                        %
\newcommand{\detJ}{\lvert \det \Jgeo \rvert}

\newcommand{\Gc}{G_{\mathrm{c}}}
\newcommand{\Gcz}{G_{\mathrm{c},\mathrm{ref}}}               %
\newcommand{\lreg}{l}                             %
\newcommand{\lgrif}{l_{\mathrm{G}}}               %
\newcommand{\kres}{\kappa}                        %

\newcommand{\grad}{\nabla}

\newcommand{\dive}{\operatorname{div}}
\newcommand{\trace}{\operatorname{tr}}
\newcommand{\dV}{\,\mathrm{d}\Omega}
\newcommand{\dA}{\,\mathrm{d}\Gamma}

\newcommand{\Ene}{\Psi}                           %
\newcommand{\EneT}{\Psi_{\mathrm{T}}}             %
\newcommand{\edens}{\psi_{\mathrm{e}}}            %
\newcommand{\psiir}{\psi_{\mathrm{ir}}}           %
\newcommand{\psihyb}{\psi_{\mathrm{hyb}}}         %
\newcommand{\psitot}{\psi_{\mathrm{tot}}}         %
\newcommand{\dmg}{g}                              %

\newcommand{\Tf}{T}                               %
\newcommand{\Tp}{T_{n-1}}                         %
\newcommand{\Tz}{T_{0}}                           %
\newcommand{\Tenv}{T_{\mathrm{b}}}                %
\newcommand{\dT}{\Delta T}                        %
\newcommand{\alphaT}{\alpha}                      %
\newcommand{\ath}{\varepsilon_{\mathrm{th}}}      %
\newcommand{\kcond}{k}                            %
\newcommand{\rhoc}{\rho c}                        %
\newcommand{\hconv}{h}                            %
\newcommand{\Gcov}{\Gamma_{\mathrm{h}}}           %
\newcommand{\GT}{\Gamma_{\mathrm{T}}}             %
\newcommand{\netT}{\mathcal{N}_{\mathrm{T}}}  %
\newcommand{\netM}{\mathcal{N}_{\mathrm{M}}}     %

\newcommand{\R}{\mathbb{R}}
\newcommand{\Ex}{\mathbb{E}}                      %
\newcommand{\sg}[1]{\mathrm{sg}\!\left[#1\right]} %
\newcommand{\pospart}[1]{\left\langle #1 \right\rangle_{+}}

\newcommand{\Fenc}{\bm{F}_{\mathrm{M}}}
\newcommand{\FencT}{\bm{F}_{\mathrm{T}}}                        %
\newcommand{\paperone}{\cite{zhang2026meshfree}}  %

%% file: sections/S0_abstract.tex
\begin{abstract}
Phase-field simulation of thermally driven fracture requires resolving localized crack
bands whose paths are unknown in advance and thermal boundary layers
whose thickness evolves during the transient. Neural energy-based
computations of thermal fracture have so far started from defects
placed in advance. This paper proposes a mesh-free deep energy method
based on multiresolution neural representations that predicts
spontaneous crack nucleation, crack spacing, and selective crack
propagation and arrest under transient thermal loading. One network
represents the displacement and phase fields and a second network the
transient temperature, each with its own multiresolution encoding on
feature grids. The mechanical energy and the incremental functional of the backward Euler heat equation, from which the temperature is computed, are both estimated on integration points redrawn at every iteration. Restricting
irreversibility to the crack set, together with carrying the optimizer
state across load steps and decaying the learning rates within each
step, helps prevent spurious damage accumulation and the widening of
the crack bands. The performance of the proposed method is assessed on
five numerical examples. On two notched plates, the initiation peak
loads agree closely with an independent finite element reference. On a
directionally quenched glass strip under an analytical temperature
field, the crack path changes from straight to oscillatory and then to
branched as the temperature jump grows, with an oscillation threshold close to the linear stability prediction. Under a prescribed boundary
temperature, a quenched ceramic slab forms a crack array of the type
observed in experiment, with a spacing close to that of a published
benchmark computation. On the slab and on a quenched annulus, the
predicted crack arrays are denser than those of a finite element
computation on the same material field. These results demonstrate the
capability of the proposed mesh-free method to predict complex crack
trajectories and spontaneously emerging fracture patterns under thermal
loading without prescribing crack paths or nucleation sites and
without meshes adapted to the cracks.
\end{abstract}


%% file: sections/S1_intro.tex
\section{Introduction}
\label{sec:intro}

Brittle solids subjected to rapid surface cooling form
characteristic fracture patterns. A
ceramic slab quenched into water develops a periodic array of parallel
cracks along its cooled faces, a subset of which propagates into the
depth while the rest arrest, so that the spacing of the array grows
with the depth~\cite{bahr1986,shao2010,jiang2012}. A glass strip
notched at its leading edge and drawn at constant speed from a hot
region into a cold bath develops a single crack, which starts at the
notch and follows the immersion front. As the temperature jump or the
drawing speed grows, the path changes from straight to oscillatory and
then to branched~\cite{yuse1993,yuse1997}. The same mechanism, a contraction
that the surrounding material resists, produces the columnar joints of
cooling lava~\cite{goehring2009}, the crack networks of drying and
cooling in rock, concrete and soil~\cite{jagla2002,bahr2010,sicsic2014}
and the thermal shock fracture of ceramic
components~\cite{lu1998,zeng2026}. The classical theory of thermal
shock and the experiments that tested
it~\cite{hasselman1969,lu1998,nematnasser1978,bazant1979,bahr2010,geyer1982}
describe the initiation of fracture from flaws of given size and the propagation and selective arrest of an initial crack array assumed in advance. A computation of thermal shock has to predict
the pattern itself, where the cracks nucleate, how many form, at what
spacing, and which of them continue.

The sharp-crack description of linear elastic fracture
mechanics represents the crack explicitly and advances it when the
energy release rate~\cite{griffith1921}, or the stress intensity
factor~\cite{irwin1957}, reaches a critical value. Remeshing~\cite{bittencourt1996},
the extended finite element method~\cite{moes1999} and element-free
Galerkin methods~\cite{belytschko1994,rabczuk2004} carry such a crack
with separate criteria for nucleation, for the direction of growth and
for branching, and such cracks have been advanced under thermal
loads~\cite{duflot2008,menouillard2011,chiaramonte2020}. On the thermal
shock problem those criteria would have to supply the quantities the
computation is meant to predict. Cohesive surfaces between all
elements~\cite{xu1994}, peridynamic models~\cite{kilic2009,xu2018peri,li2025pddl}
and lattice models~\cite{hayakawa1994} instead form cracks where bonds
or cohesive surfaces fail.

Of the formulations that produce such patterns without assuming them,
the one adopted here is the variational formulation of fracture of Francfort and
Marigo~\cite{francfort1998} in the regularization of Bourdin et
al.~\cite{bourdin2000,bourdin2008}. In it the sharp crack is replaced
by a phase field that localizes in a band of width proportional to a
regularization length $\lreg$~\cite{ambrosio1990}, and nucleation,
propagation, branching and arrest result from minimizing one energy. The model has since been extended by energy decompositions that treat tension and compression differently~\cite{miehe2010cmame,amor2009,ambati2015}, interpreted as a
gradient damage model with a finite strength~\cite{pham2011,wu2017,tanne2018}, and given a strength surface~\cite{kumar2020}, higher-order densities~\cite{borden2014} and
penalized irreversibility~\cite{gerasimov2019}, as reviewed
in~\cite{wu2020}. On the thermal shock problem Sicsic et
al.~\cite{sicsic2014} showed that the uniform damage band under the
cooled surface bifurcates toward a periodic mode whose wavelength is
proportional to the regularization length. The computations of Bourdin
et al.~\cite{bourdin2014} pass through this bifurcation to an array of
short cracks whose selective propagation follows the spacing law of
Bahr et al.~\cite{bahr2010} and the experiment of Shao et
al.~\cite{shao2011}. The problem tests the model both at nucleation, where the strength governs, and in propagation, where the toughness governs, and in the regularized model the two are not independent, since the strength is set by the toughness, the modulus and the regularization length together~\cite{tanne2018}. Lopez-Pamies et al.~\cite{lopezpamies2025} argue that classical models of this class, without a strength surface, cannot describe nucleation in general, and Kamarei et al.~\cite{kamarei2026} pose a set of challenge problems on which
fracture models can be assessed in this respect. Zeng et
al.~\cite{zeng2025nucleation} treat nucleation in thermal quenching
with a model whose strength is independent of the regularization
length, together with random fields of the strength.

Of the two experiments above, the glass strip is the one in which
quasi-static crack propagation under a thermal load is controlled and
reproducible. Yuse and
Sano~\cite{yuse1993,yuse1997} mapped its straight, oscillatory and
branched regimes and reported the transition to oscillation as a
supercritical Hopf bifurcation. Ronsin et al.~\cite{ronsin1995}
measured the temperature field and the thresholds as functions of the
drawing speed, Ronsin and Perrin~\cite{ronsin1998} the thresholds and
the wavelength at onset, and Yang and Ravi-Chandar~\cite{yang2001}
followed the growth of the crack in real time. Linear stability
analyses~\cite{marder1994,sasa1994,adda1995,corson2009} and an equation of motion for the tip~\cite{bouchbinder2003} gave the threshold and, in
most cases, the wavelength, and the paths were computed with lattice models, finite elements, the extended finite
element method, universal meshes, peridynamics and phase-field
models~\cite{hayakawa1994,bahr1995,menouillard2011,chiaramonte2020,xu2018peri,corson2009,pan2024,zeng2026}.
The computation
of Corson et al.~\cite{corson2009} recovered the transition but not the complex paths
well above the threshold, where the authors suggest that the quasi-static
approximation may no longer hold.

Phase-field computations of thermo-elastic fracture have been made
with finite elements in quasi-static and dynamic settings and with
monolithic, explicit and staggered
solvers~\cite{miehe2015,chu2017,wang2020,mandal2021fracture,tangella2022},
for the hot cracking of the melt pool in additive manufacturing~\cite{ruan2023thermomechanical}, in three
dimensions~\cite{li2022thermal,wang2024thermal}, on meshes adapted
around the growing cracks~\cite{badnava2018,ferro2018,chen2024} and
with a strength independent of the regularization
length~\cite{zeng2026}. Each of them meets the resolution requirement of the model with a mesh, and a strong-form meshless method~\cite{ali2025} meets it with a set of nodes. On the thermal shock problem the mesh has to resolve two
narrow zones that change during the computation. One is the band of width proportional to
$\lreg$, along a path the computation is meant to
discover~\cite{miehe2010cmame,si2023}. The other is the thermal boundary
layer, whose thickness after a cooling time $t$ is $\sqrt{D t}$, with $D$ the thermal diffusivity, and is smallest early in the transient, when the
array nucleates. A refinement that anticipates
either is not neutral, since nonuniform meshes can favor the nucleation
of cracks in their refined zones~\cite{pham2011}. A uniform mesh
avoids this bias at the cost of the finest resolution everywhere, as
in the finite element computations of this paper. In purely elastic
fracture the authors' mesh-free deep energy method~\paperone\ is built so
that no mesh follows the crack.

Neural networks with smooth activations can serve as global trial
functions that are defined without a mesh and differentiated to any
order by automatic differentiation~\cite{lagaris1998,raissi2019,karniadakis2021}. For
problems governed by a minimum principle the deep Ritz
method~\cite{e2018deepritz} and the deep energy
method~\cite{samaniego2020,nguyenthanh2020} minimize the corresponding
functional over the network parameters. Essential conditions can be
imposed exactly through lifts~\cite{sukumar2022}, the distribution of
the training points affects the accuracy of the solution and the
convergence of the training~\cite{wu2023,nabian2021}, and input
encodings let networks represent steep
fields~\cite{tancik2020,mueller2022} and can speed up
training~\cite{huang2024hash}.
Phase-field modeling of brittle fracture has been treated in this way
by Goswami et al.~\cite{goswami2020tl,goswami2020fourth} and Manav et
al.~\cite{manav2024} and in a number of
variants~\cite{chakraborty2022,motlagh2023,feng2025gnn,zhang2026chebykan,plunge2026,dean2026,yu2026pinn}.
An explicit crack has been advanced by the extended deep energy method
of Wang et al.~\cite{wang2026xdem}, networks have served as solution
operators and surrogates~\cite{kiyani2025,hamdi2026,ani2026}, and
networks that minimize a peridynamic energy have propagated
cracks~\cite{zhouyu2024transfer,zhouma2026nonlocal}. The study
of~\paperone\ found that the crack advanced only when the integration points were redrawn during training rather than held fixed.

In the thermal and thermo-elastic applications of
networks~\cite{cai2021,lin2024,mojahedin2022,raj2021pinnfgm,harandi2024},
no crack was present, and where networks have been used for thermal fracture, the crack was mostly
stationary and a stress intensity factor was the
output~\cite{yuzhou2023nonlocal,wu2026,yadav2026mpinn}. Three recent
studies allow the crack to grow under a thermal load. He et
al.~\cite{he2026drm} apply the deep Ritz method to fracture under a steady or
spatially uniform temperature field, on a finite element mesh and from
initial cracks. Kong~\cite{kong2026thesis} grows cracks from notches under a steady temperature held fixed during loading, with an
energy-based network, a second-order phase field and a history field.
The
authors' extended deep energy method~\cite{zhang2026xdem} solves the transient
temperature on the cracked domain and advances an explicit crack from a
notch. In all three the crack starts from a defect placed in advance,
so the nucleation, the spacing and the selection of cracks under a
transient temperature field are not computed.

This paper proposes a mesh-free deep energy method that computes them. It builds on the solver of~\paperone\
(Section~\ref{sec:method}) and is mesh-free in the sense that no
connectivity is built and no point set is held fixed. Two difficulties
arise when a transient temperature drives the cracks. The thermal
boundary layer can be thinner than the crack band at the first time
step and thickens as the transient advances, so the temperature is computed by a second network with a multiresolution encoding of its own, from an incremental functional of the backward Euler heat equation in which the convective condition is a natural term. Under a uniformly cooled
surface the damage is diffuse over a large region before cracks
nucleate, and a stochastic solver that retains every small increase of the phase
field turns the cooled layer into a damaged film. Irreversibility is
therefore imposed on the crack set only (Section~\ref{subsec:thermoelastic}),
and, to keep the bands from widening, the optimizer state is carried across load steps and the learning rates decay within each step (Section~\ref{subsec:staggered}). The resolution is set by the uniform grids of the encoding, and the integration points follow the crack band as it develops, so no mesh is
fitted to the band or to the boundary layer and nothing is remeshed as
the array grows. A curved surface is described exactly by a spline map. Where the heat transfer is little affected by the cracks and the compared computations use the same closed form, as for a strip drawn at constant speed into a bath, an analytical temperature field can be used instead of the computed one. It enters the mechanical problem as a known thermal strain. Where the temperature is computed, each example treats the conductivity as damage-dependent, if the cracks cross the heat flux, or as damage-independent, if they run along it.

The method is assessed on five problems. A
single-edge-notched plate under tension and a functionally graded plate
under shear, each at three thermal loads, verify the thermo-mechanical solver
against an independent staggered finite element reference of the same
phase-field model, up to the differences stated in
Section~\ref{sec:results}. The directional quenching of a glass strip
tests the evolution of a crack path under an analytical temperature
field and is compared with a finite element computation, the phase-field computation of Zeng and Dolbow~\cite{zeng2026},
the linear stability analysis of Corson et al.~\cite{corson2009}, and
the experiments of Yuse and Sano~\cite{yuse1993,yuse1997} and of Ronsin
and Perrin~\cite{ronsin1998}. The quenched ceramic slab combines a
computed thermal transient with the spontaneous nucleation of an array,
and a quenched annulus extends it to a curved domain. Both are compared
with a finite element computation on the same realization of the random
material field of Tang et al.~\cite{tang2016}, which breaks the symmetry
of the uniformly cooled surface. The slab is also compared with the
experiments of Jiang et al.~\cite{jiang2012} and Shao et
al.~\cite{shao2011} and the benchmark computation of Bourdin et
al.~\cite{bourdin2014}.

Section~\ref{sec:formulation} states the model,
Section~\ref{sec:method} the method, Section~\ref{sec:results} the five
examples and Section~\ref{sec:conclusions} the conclusions, with the
numerical parameters and the supporting studies in the appendices.

%% file: sections/S2_formulation.tex
\section{Thermo-elastic phase-field model}
\label{sec:formulation}

Consider a linearly elastic body occupying an open bounded domain
$\Omega \subset \R^{2}$, analyzed under plane strain or plane stress as
each example states, with $z$ normal to the plane. The body is loaded
quasi-statically through prescribed boundary displacements and through a
temperature field $\Tf(\xb, t)$, with $\xb = (x, y)$ the position and
$t$ the time. The temperature is either computed from a heat conduction
problem or given as an analytical field. Cracks are described by a scalar phase field
$\pf : \Omega \to [0,1]$, with $\pf = 0$ in intact material and $\pf = 1$
on a fully developed crack, and small strains
$\straintensor(\ub) = \tfrac{1}{2}(\grad\ub + \grad\ub^{\mathrm{T}})$
are assumed for the displacement $\ub = (u, v)^{\mathrm{T}}$. The
fracture energy density, the strain energy decomposition and the hybrid
formulation are adopted from the purely elastic method of~\paperone.
The fracture energy density is the
second-order density in the approximation of Ambrosio and
Tortorelli~\cite{ambrosio1990}, commonly denoted AT2. The strain energy
is split spectrally into a tensile part that drives the crack and a
compressive part that does not, and in the hybrid formulation the
stress is degraded isotropically while only the tensile part drives the
phase field. These ingredients are stated below with the temperature
included, followed by irreversibility, the heat conduction problem and
the coupling of the two problems.
Table~\ref{tab:notation-te} collects the main symbols of the paper.

\begin{table}[htb]
    \centering
    \caption{Main symbols. Total energies are written with an uppercase
    $\Ene$, densities with a lowercase $\psi$, as in~\paperone. The
    load step index $n$ appears as a subscript, and descriptive
    subscripts are upright ($\ath$, $\pf_{\mathrm{th}}$). $\bm{I}$ is the
    identity tensor, $\trace$ the trace, $\trace_{2}$ and $\trace_{3}$ the
    in-plane and the three-dimensional trace, $\nabla$ the gradient,
    $\mathrm{div}$ the divergence, the colon the double contraction,
    $|\cdot|$ the Euclidean norm, the bold $\bm{n}$ the outward normal of
    the boundary, $\langle\cdot\rangle_{\pm}$ the positive and negative parts, $\Ex$ the
    expectation, a hat on $\mathcal{L}$, $\EneT$ and $S$ an estimate
    from sampled points and a hat on $u$, $v$ and $\pf$ the raw output
    of the mechanical network, and $\sg{\cdot}$ the stop-gradient. On
    $E_{i}$, $\sigma_{\mathrm{t},i}$ and $G_{\mathrm{c},i}$ the subscript
    $i$ marks a cell of the material field, and $\bar{\sigma}_{\mathrm{t}}$
    is the nominal strength.}
    \label{tab:notation-te}
    \scriptsize
    \begin{tabular}{@{}l>{\raggedright\arraybackslash}p{5.2cm}@{\quad}l>{\raggedright\arraybackslash}p{5.0cm}@{}}
        \toprule
        Symbol & Meaning & Symbol & Meaning \\
        \midrule
        $\Omega$, $\widehat{\Omega}$ & physical, parametric domain &
        $\netM$, $\bm{\theta}_{\mathrm{M}}$ & mechanical network, its parameters \\
        $\xb$, $\xib$, $t$ & position, parametric coordinates, time &
        $\netT$, $\bm{\theta}_{\mathrm{T}}$ & temperature network, its parameters \\
        $\ub$, $\straintensor$, $\stresstensor$ & displacement, strain, stress &
        $\Fenc$, $\FencT$, $N_{\mathrm{lev}}$ & feature vectors of the two networks, number of levels \\
        $\rho$, $c$ & density, specific heat &
        $\bm{G}_{\ell,ij}$, $N_{\ell,i}$, $n_{\ell}$ & nodal features of level $\ell$, quadratic B-spline, nodes along $\eta$ \\
        $\straintensor_{\mathrm{e}}$, $\ath$, $\alphaT$ & elastic strain, scalar thermal expansion strain, expansion coefficient &
        $\varepsilon^{*}$ & effective in-plane eigenstrain of plane strain \\
        $E$, $\nu$, $\lambda$, $\mu$, $\lambda'$ & elastic constants, plane-stress constant &
        $\Jgeo$ & Jacobian of the geometry map \\
        $\pf$, $\pf_{\mathrm{notch}}$ & phase field, seeded notch profile &
        $p$, $M$ & sampling density on $\widehat{\Omega}$, points per iteration \\
        $\lreg$, $\Gc$, $\Gcz$ & regularization length, toughness, reference toughness at $T_{\mathrm{ref}}$ &
        $Y_{\mathrm{proc}}$ & driving-force indicator of the sampler \\
        $\Ene$, $\edens$, $\edens^{\pm}$ & total energy, elastic density, its tensile and compressive parts &
        $\omega_{\mathrm{M}}$, $\omega_{\mathrm{T}}$ & envelopes of the displacement and temperature network terms \\
        $\dmg$, $\kres$ & degradation function, residual stiffness &
        $U_{\mathrm{ref}}$, $\dT_{\mathrm{ref}}$, $u_{\mathrm{D}}$ & displacement and temperature scales, displacement lift \\
        $\psihyb$, $\psi_{\mathrm{c}}$, $\psiir$, $\psitot$ & folded hybrid, fracture and penalty densities, their sum &
        $s$ & squash function of the phase field \\
        $\gamma_{\mathrm{ir}}$, $\pf_{\mathrm{th}}$, $\chi$ & penalty weight, threshold and indicator of the crack set &
        $n_{\mathrm{w}}$, $\epsilon_{\mathrm{tol}}$, $\mathcal{W}$, $\widehat{S}$ & plateau window length, tolerance, window, monitored estimate \\
        $\mathcal{L}_{n}$, $n$, $n_{\max}$, $\delta_{n}$ & loss of load step $n$, load step, number of load steps, prescribed displacement &
        $F_{n}$, $R_{\ub}$, $R_{\pf}$ & reaction force, residuals of the mechanical problem \\
        $\Tf$, $\Tz$, $\Tenv$, $\dT$ & temperature, stress-free and initial temperature, bath temperature, thermal load ($T_{\mathrm{edge}} - \Tz$ on the plates, $\Tz - \Tenv$ on the strip, the slab and the annulus) &
        $\lgrif$, $\sigma_{\mathrm{c}}$ & Griffith length $\Gc / (E \alphaT^{2} \dT^{2})$, peak stress of the homogeneous AT2 model \\
        $\kcond$, $\kcond_{\mathrm{d}}$, $\rhoc$, $D$ & conductivity, conductivity of a load step, volumetric heat capacity, diffusivity &
        $m$, $E_{i}$, $\sigma_{\mathrm{t},i}$, $G_{\mathrm{c},i}$, $\bar{\sigma}_{\mathrm{t}}$ & Weibull modulus, modulus, strength and toughness of cell $i$ of the material field, nominal strength \\
        $\hconv$, $\Gcov$, $\GT$, $T_{\mathrm{D}}$ & film coefficient, convective boundary, boundary with prescribed temperature, prescribed boundary temperature and its extension into the domain, or the analytical field &
        $\EneT$, $\psi_{\mathrm{T}}$, $\Delta t$, $t_{n}$, $M_{e}$, $\zeta$ & incremental thermal functional, its density, time step, time of load step $n$, points on edge $e$ and its parameter \\
        $V$, $b$ & drawing speed and half-width of the strip &
        $P$, $R$, $R_{\mathrm{c}}$ & P\'eclet and thermal-shock numbers, threshold of oscillation \\
        $A$, $\Lambda$, $Q$, $(x_{q}, y_{q})$ & amplitude and measured wavelength of an oscillating path, number and position of its turning points &
        $d$, $a$ & crack spacing, depth from the cooled face \\
        $K_{\mathrm{c}}$ & critical stress intensity factor &
        $b_{1}$, $b_{2}$, $T_{\mathrm{ref}}$, $T_{\mathrm{max}}$, $\Theta$ & constants, reference, normalizing and normalized temperature of the toughness law \\
        $w_{\mathrm{u}}$, $w_{\mathrm{c}}$, $w_{\mathrm{p}}$, $\beta_{0}$, $\beta_{\mathrm{d}}$ & stratum weights, process-zone weights of the sampler &
        $t_{0}$, $t_{\mathrm{s}}$, $t_{\mathrm{f}}$ & initial, first and last time of the quench time grid \\
        $\mathcal{U}_{\mathrm{T},n}$, $\mathcal{U}_{\mathrm{M},n}$, $\tau$ & thermal and mechanical updates of load step $n$, iteration index &
        $r$, $r_{1}$, $r_{2}$, $r_{\mathrm{T}}$ & distance to the notch, distances to the cooled surfaces, decay length of the temperature lift \\
        \bottomrule
    \end{tabular}
\end{table}

\subsection{Energy functional}
\label{subsec:thermoelastic}

The temperature enters the mechanics through the scalar thermal
expansion strain $\ath$ and the elastic strain
$\straintensor_{\mathrm{e}}$,
\begin{equation}
    \ath = \alphaT\, (\Tf - \Tz) ,
    \qquad
    \straintensor_{\mathrm{e}} = \straintensor(\ub) - \ath\, \bm{I} ,
    \label{eq:eps-e}
\end{equation}
where $\alphaT$ is the coefficient of thermal expansion and $\Tz$ the
stress-free temperature at which the thermal strain vanishes.

The total energy of the body at a given temperature field reads
\begin{equation}
    \Ene(\ub, \pf;\, \Tf)
    =
    \int_{\Omega}
    \bigl[
        \dmg(\pf)\, \edens^{+}(\straintensor_{\mathrm{e}})
        + \edens^{-}(\straintensor_{\mathrm{e}})
    \bigr]
    \dV
    +
    \int_{\Omega}
    \frac{\Gc}{2\lreg}
    \bigl( \pf^{2} + \lreg^{2}\, |\grad\pf|^{2} \bigr)
    \dV ,
    \label{eq:pi-te}
\end{equation}
in which $\dmg(\pf) = (1 - \pf)^{2} + \kres$ is the quadratic
degradation function with the residual stiffness $\kres$. The second
integral is the fracture energy in the form of Bourdin et
al.~\cite{bourdin2000,bourdin2008}, with the toughness $\Gc$ and the
regularization length $\lreg$. The densities
$\edens^{\pm}$ are the tensile and compressive parts, in the spectral
decomposition of Miehe et al.~\cite{miehe2010,miehe2010cmame}, of the strain
energy density of the undamaged material,
\begin{equation}
    \edens(\straintensor_{\mathrm{e}})
    =
    \tfrac{1}{2}\lambda\, \bigl( \trace\straintensor_{\mathrm{e}} \bigr)^{2}
    + \mu\, \straintensor_{\mathrm{e}} : \straintensor_{\mathrm{e}} ,
    \qquad
    \edens^{\pm}(\straintensor_{\mathrm{e}})
    =
    \tfrac{1}{2}\lambda\, \bigl\langle \trace\straintensor_{\mathrm{e}} \bigr\rangle_{\pm}^{2}
    + \mu \sum_{i} \bigl\langle \varepsilon_{\mathrm{e},i} \bigr\rangle_{\pm}^{2} ,
    \label{eq:psi-e}
\end{equation}
evaluated on the elastic strain, with $\varepsilon_{\mathrm{e},i}$ its
principal values and the Lam\'e constants $\lambda$ and $\mu$, which
follow from Young's modulus $E$ and Poisson's ratio $\nu$. Under plane strain the total
out-of-plane strain vanishes, so the elastic strain contains the
restrained component $\varepsilon_{\mathrm{e},zz} = -\ath$, which enters Eq.~\eqref{eq:psi-e} through the three-dimensional trace $\trace_{3}\straintensor_{\mathrm{e}}$ and through the principal values. Under plane stress the split is the
common two-dimensional approximation, in which $\lambda$ is replaced by
$\lambda' = 2\lambda\mu / (\lambda + 2\mu)$ and the trace and the
principal values are those of the in-plane elastic strain.
The toughness $\Gc = \Gc(\xb, \Tf)$ may vary in space and with the
temperature. Where it depends on the temperature, it follows the
quadratic law of Ruan et al.~\cite{ruan2023thermomechanical},
\begin{equation}
    \Gc(\Tf) = \Gcz\,
    \bigl( 1 - b_{1}\,\Theta + b_{2}\,\Theta^{2} \bigr) ,
    \qquad
    \Theta = \frac{\Tf - T_{\mathrm{ref}}}{T_{\mathrm{max}}} ,
    \label{eq:gcT}
\end{equation}
in which $\Gcz$, which may vary in space, is the toughness at the
temperature $T_{\mathrm{ref}}$, $b_{1}$ and $b_{2}$ are material
constants, $T_{\mathrm{max}}$ is the normalizing temperature of that
law and $\Theta$ is the normalized temperature.

Equation~\eqref{eq:pi-te} defines the fracture energy and the tensile part that drives the crack and is not itself minimized. The equations solved are those of the hybrid formulation of
Ambati et al.~\cite{ambati2015}, in the folded form of~\paperone\ and
applied to the elastic strain in place of the total strain. In this
formulation the stress of the damaged material is degraded
isotropically,
\begin{equation}
    \stresstensor
    =
    \dmg(\pf)\, \frac{\partial \edens}{\partial \straintensor_{\mathrm{e}}} ,
    \label{eq:stress-te}
\end{equation}
while the phase field is driven by $\edens^{+}$ alone. The stress and
the driving force are both carried by a single density, folded in the
sense that its derivative with respect to $\ub$ gives the degraded
stress and its derivative with respect to $\pf$ the driving force of
$\edens^{+}$ alone,
\begin{equation}
    \psihyb
    =
    \dmg\bigl(\sg{\pf}\bigr)\, \edens(\straintensor_{\mathrm{e}})
    +
    \Bigl( \dmg(\pf) - \sg{\dmg(\pf)} \Bigr)\,
    \sg{\edens^{+}(\straintensor_{\mathrm{e}})} ,
    \label{eq:hybrid-te}
\end{equation}
where $\sg{\cdot}$ denotes the stop-gradient operation, whose argument
is treated as a constant during differentiation. These two derivatives
give the equilibrium and the phase-field equations of the hybrid
formulation, as shown in~\paperone, and the value of $\psihyb$ is the degraded
strain energy density $\dmg(\pf)\, \edens(\straintensor_{\mathrm{e}})$,
so the computed energies and the reaction force keep their mechanical
meaning. The constraint of the formulation of Ambati et al., $\pf = 0$ where $\edens^{+} < \edens^{-}$, is not imposed.

The load is applied in steps $n = 1, 2, \dots$, and irreversibility
between successive steps is enforced by the one-sided penalty
\begin{equation}
    \psiir
    =
    \gamma_{\mathrm{ir}}\,
    \chi\bigl(\pf_{n-1} \geq \pf_{\mathrm{th}}\bigr)\,
    \pospart{\pf_{n-1} - \pf}^{2} ,
    \label{eq:penalty-te}
\end{equation}
with the penalty weight $\gamma_{\mathrm{ir}}$, the trial phase field
$\pf$ of step $n$, written $\pf_{n}$ once the step is accepted, the
phase field $\pf_{n-1}$ accepted at the end of the previous step, and
the indicator
$\chi$ of the crack set
$\{\pf_{n-1} \geq \pf_{\mathrm{th}}\}$ with the crack-set threshold
$\pf_{\mathrm{th}}$. The indicator restricts the penalty to the
developed cracks, so that outside the crack set the phase field may
decrease. The construction follows the crack-set irreversibility of
Bourdin et al.~\cite{bourdin2000,bourdin2008}, in which partially
damaged regions may heal~\cite{gerasimov2019,linse2017}, except that
the penalty acts against a decrease below the previous value rather
than setting $\pf = 1$ on the crack set.

With $\pf_{\mathrm{th}} = 0$ the penalty acts on the whole domain, the
form of~\paperone. Where diffuse damage covers a large region, as under
a uniformly cooled surface before cracks nucleate, each stochastic
update of the optimizer (Section~\ref{sec:method}) perturbs the phase field slightly, and a penalty on the whole domain retains every increase, so that the increases accumulate into a damaged film. The threshold $\pf_{\mathrm{th}} = 1/2$ counters this, and two options of the mechanical update (Section~\ref{subsec:staggered}) keep the bands from widening (\ref{app:arrays}). Since the weight
$\gamma_{\mathrm{ir}}$ is finite, the constraint holds approximately
(\ref{app:hyper}).

At load step $n$, with the temperature $\Tf_{n}$ and, where the example
has one, the prescribed boundary displacement $\delta_{n}$, the
mechanical problem is stated on the loss
\begin{equation}
    \mathcal{L}_{n}(\ub, \pf;\, \Tf_{n}, \pf_{n-1}, \delta_{n})
    =
    \int_{\Omega}
    \bigl[ \psihyb + \psi_{\mathrm{c}} + \psiir \bigr] \dV ,
    \label{eq:incremental-te}
\end{equation}
with $\psi_{\mathrm{c}}$ the fracture energy density of
Eq.~\eqref{eq:pi-te}. The loss is written $\mathcal{L}$ below where
the step is clear. The solution of the step is the pair $(\ub_{n}, \pf_{n})$, with
$\ub$ admissible for $\delta_{n}$, at which the two residuals
\begin{equation}
    R_{\ub}(\ub, \pf) = \frac{\partial \mathcal{L}}{\partial \ub} = \bm{0} ,
    \qquad
    R_{\pf}(\ub, \pf) = \frac{\partial \mathcal{L}}{\partial \pf} = 0
    \label{eq:residuals-te}
\end{equation}
vanish, both derivatives taken with the arguments of $\sg{\cdot}$ held
constant. By the construction of Eq.~\eqref{eq:hybrid-te}, $R_{\ub}$ is
the weak form of the equilibrium equation with the stress of
Eq.~\eqref{eq:stress-te}, and $R_{\pf}$ that of the phase-field
equation driven by $\edens^{+}$, with the penalty added. As
in~\paperone, the two fields are advanced together by one
gradient-based update per optimizer iteration on these derivatives. This update, like the hybrid formulation itself, does not minimize a single energy~\cite{ambati2015,wu2020}, so the value of $\mathcal{L}$ need not
decrease from one iteration to the next. In the isothermal limit $\Tf = \Tz$
with $\pf_{\mathrm{th}} = 0$ the formulation reduces to that
of~\paperone. Without the penalty, the strong forms of
the two conditions are
\begin{equation}
    \dive \stresstensor = \bm{0} ,
    \qquad
    \frac{\Gc}{\lreg}\, \pf
    - \lreg\, \dive \bigl( \Gc\, \grad\pf \bigr)
    =
    2 (1 - \pf)\, \edens^{+}(\straintensor_{\mathrm{e}}) ,
    \label{eq:strong-te}
\end{equation}
with the natural condition $\Gc\, \grad\pf \cdot \bm{n} = 0$ on the
boundary.

\FloatBarrier
\subsection{Heat conduction}
\label{subsec:heat}

Where the temperature is computed from the heat conduction problem below, each example treats the conductivity in one of two ways. Where the cracks cross the heat
flux, their faces are taken as adiabatic through a damage-dependent conductivity~\cite{tangella2022,ruan2023thermomechanical}. The conductivity of load step $n$ is degraded by the phase field of the previous step, so that the thermal problem of a step can be solved before its mechanical problem,
\begin{equation}
    \kcond_{\mathrm{d},n} = \kcond\, \dmg(\pf_{n-1})
    = \kcond \bigl[ (1 - \pf_{n-1})^{2} + \kres \bigr] ,
    \label{eq:kdeg}
\end{equation}
with the conductivity $\kcond$ of the intact material and the residual
stiffness $\kres$. For a finite $\lreg$ and $\kres > 0$ the crack band
retains a small conductance, so Eq.~\eqref{eq:kdeg} is a phase-field
approximation of adiabatic crack faces. Where the cracks run along the
heat flux, the conductivity is damage-independent~\cite{bourdin2014,chu2017,mandal2021fracture,tangella2022}, $\kcond_{\mathrm{d},n} = \kcond$, so that the crack band conducts heat as intact material and the temperature does not depend on the cracks. Table~\ref{tab:hyper} lists the treatment of each example.

The temperature obeys the transient heat conduction equation. The
boundary of $\Omega$ is divided into a part $\GT$ with prescribed temperature, a
convective part $\Gcov$ through which heat passes to the bath, and an
insulated remainder,
\begin{equation}
    \begin{aligned}
    \rhoc\, \frac{\partial \Tf}{\partial t}
    &= \dive \bigl( \kcond_{\mathrm{d}}\, \grad\Tf \bigr)
    && \text{in } \Omega ,
    \\
    \Tf &= T_{\mathrm{D}}
    && \text{on } \GT ,
    \\
    \kcond_{\mathrm{d}}\, \grad\Tf \cdot \bm{n}
    &= -\hconv\, (\Tf - \Tenv)
    && \text{on } \Gcov ,
    \end{aligned}
    \label{eq:heat-strong}
\end{equation}
with the volumetric heat capacity $\rhoc$, the conductivity
$\kcond_{\mathrm{d}}$ of the load step, the prescribed boundary temperature $T_{\mathrm{D}}$, the film coefficient $\hconv$ and the
bath temperature $\Tenv$. The thermal diffusivity is $D = \kcond / (\rho c)$, formed with the conductivity $\kcond$ of the intact material, with $\rhoc$ the product of the density $\rho$ and the specific heat $c$. The
backward Euler difference over a step $\Delta t = t_{n} - t_{n-1}$, with
$t_{n}$ the time of load step $n$, turns Eq.~\eqref{eq:heat-strong} into the
stationarity condition of the incremental functional
\begin{equation}
    \Psi_{\mathrm{T},n}(\Tf;\, \Tp, \pf_{n-1})
    =
    \int_{\Omega}
    \Bigl[
        \tfrac{1}{2}\, \kcond_{\mathrm{d},n}\, |\grad\Tf|^{2}
        + \frac{\rhoc}{2\,\Delta t}\, (\Tf - \Tp)^{2}
    \Bigr]
    \dV
    +
    \int_{\Gcov}
    \tfrac{1}{2}\, \hconv\, (\Tf - \Tenv)^{2}
    \dA ,
    \label{eq:piT}
\end{equation}
with $\Tp$ the temperature of the previous step and
$\kcond_{\mathrm{d},n}$ the conductivity of the step. The functional is
written $\EneT$ where the step is clear, with $\psi_{\mathrm{T}}$ its
domain integrand. In its stationarity condition the convective and the
insulated conditions are natural and need no constraint on the field, like
traction-free edges in the mechanical problem, so only the condition on
$\GT$ is imposed. The functional is strictly convex in $\Tf$ for $\rhoc > 0$ and
$\Delta t > 0$, so the temperature at $t_{n}$ is its unique minimizer
over the fields that carry the prescribed data on $\GT$, and it is
computed by the same estimator and optimizer as the mechanical fields.
The convexity does not carry over to the parameters of the network that
represents the field. Incremental minimum principles for coupled heat
conduction~\cite{yang2006,kamagate2025} and implicit time steps
computed as minimizations over
networks~\cite{georgoulis2023,park2023,zhang2026xdem} precede this
formulation.

\FloatBarrier
\subsection{Coupling of the two problems}
\label{subsec:oneway}

Within a load step the coupling is one way. The temperature acts on the
mechanics through the thermal strain of Eq.~\eqref{eq:eps-e} and, where
the toughness depends on the temperature, through Eq.~\eqref{eq:gcT}
evaluated at $\Tf_{n}$ and held fixed during the mechanical problem. The mechanics acts on the temperature only through the damage-dependent conductivity of Eq.~\eqref{eq:kdeg}, evaluated on $\pf_{n-1}$, and not at all where the conductivity is damage-independent. The temperature of a step is therefore computed first and
passed to the mechanical problem as data
(Section~\ref{subsec:staggered}). Where the temperature is an
analytical field, as on the strip, no thermal problem is solved and the
thermal strain is a known eigenstrain. The thermo-elastic heat source~\cite{huang2025gradient} and the heat released by fracture~\cite{miehe2015} are neglected, as in the
computations of Mandal et al.~\cite{mandal2021fracture} and Bourdin et
al.~\cite{bourdin2014}.

%% file: sections/S3_method.tex
\section{Mesh-free deep energy method}
\label{sec:method}

The incremental problems of Section~\ref{sec:formulation} are solved
with the mesh-free deep energy method of~\paperone. Its representation
of the displacement and the phase field by one network on a
multiresolution encoding of B-spline feature grids, its stratified
Monte Carlo estimator on integration points redrawn at every iteration
and its training by Adam~\cite{kingma2015} are used unchanged and are restated only as
far as the temperature requires. For the thermal problem the method adds a second network for the temperature on feature grids of its own
(Section~\ref{subsec:fields-te}). The thermal functional is estimated on points
drawn from the same density as those of the mechanical loss
(Section~\ref{subsec:estimator-te}). Each load step obtains the temperature first, from the thermal update or the analytical field, and passes it to the mechanical update, for which two options are introduced (Section~\ref{subsec:staggered}). The numerical parameters
of every run are tabulated in \ref{app:hyper}.

\FloatBarrier
\subsection{Field representation and boundary conditions}
\label{subsec:fields-te}

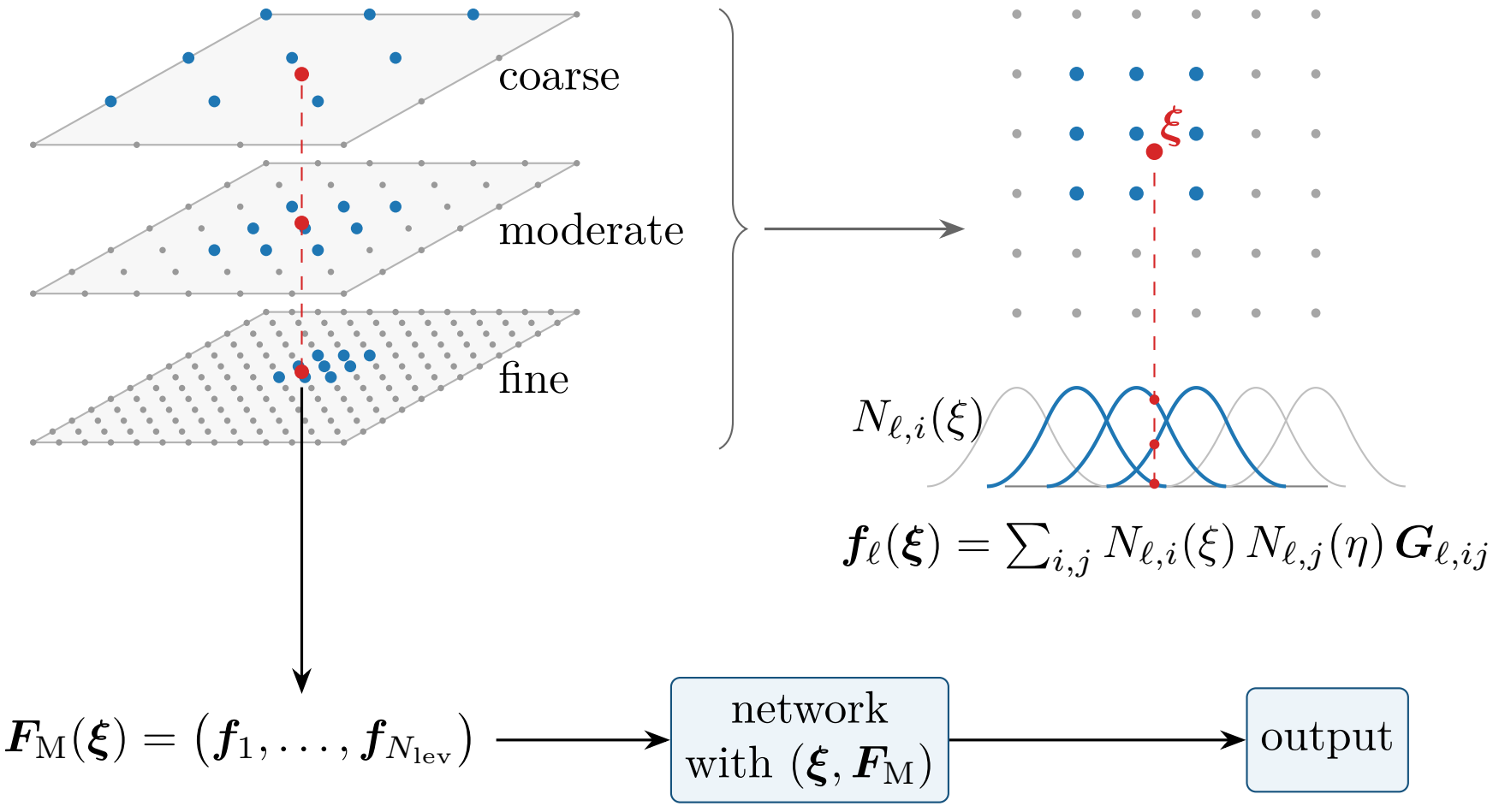
\begin{figure}[htb]
    \centering
    \begin{tikzpicture}[
        box/.style = {draw, rounded corners=2pt, align=center,
                      font=\footnotesize, inner sep=3pt, minimum height=8mm,
                      fill=cblue!8, draw=cblue!70!black},
        arr/.style = {-{Stealth[length=2mm]}, semithick}
    ]
        \foreach \k/\n/\lab in {0/13/fine, 1/7/moderate, 2/4/coarse} {
            \begin{scope}[shift={(0,\k*1.15)}, yscale=0.42, xslant=0.75]
                \draw[black!30, fill=black!3] (0,0) rectangle (2.4,2.4);
                \pgfmathsetmacro{\hh}{2.4/(\n-1)}
                \foreach \i in {0,...,\the\numexpr\n-1\relax} {
                    \foreach \j in {0,...,\the\numexpr\n-1\relax} {
                        \node[circle, fill=black!40, inner sep=0pt,
                              minimum size=1.4pt] at (\i*\hh, \j*\hh) {};
                    }
                }
                \pgfmathtruncatemacro{\qi}{round(1.1/\hh)}
                \pgfmathtruncatemacro{\qj}{round(1.3/\hh)}
                \foreach \a in {-1,0,1} {
                    \foreach \b in {-1,0,1} {
                        \node[circle, fill=cblue, inner sep=0pt,
                              minimum size=2.6pt]
                            at ({(\qi+\a)*\hh}, {(\qj+\b)*\hh}) {};
                    }
                }
                \node[circle, fill=cred, inner sep=0pt,
                      minimum size=3.2pt] at (1.1,1.3) {};
                \coordinate (q\k) at (1.1,1.3);
                \coordinate (r\k) at (2.4,1.2);
            \end{scope}
            \node[font=\footnotesize, anchor=west] at ($(r\k) + (0.15,0)$) {\lab};
        }
        \draw[cred, dashed] (q0) -- (q2);
        \begin{scope}[shift={(7.6,1.0)}, scale=0.924]
            \foreach \i in {0,...,5} {
                \foreach \j in {0,...,5} {
                    \fill[black!35] (\i*0.5, \j*0.5) circle (1.1pt);
                }
            }
            \foreach \i in {1,2,3} {
                \foreach \j in {2,3,4} {
                    \fill[cblue] (\i*0.5, \j*0.5) circle (1.7pt);
                }
            }
            \fill[cred] (1.15,1.35) circle (2.0pt);
            \coordinate (qd) at (1.15,1.35);
            \node[font=\footnotesize, cred, anchor=south west, inner sep=1pt]
                at (1.15,1.35) {$\xib$};
            \def\yb{-1.45}
            \draw[black!50] (-0.1,\yb) -- (2.6,\yb);
            \foreach \c in {0.0,2.0,2.5} {
                \pgfmathsetmacro{\ca}{\c-0.75}
                \pgfmathsetmacro{\cb}{\c+0.75}
                \draw[black!25, domain=\ca:\cb, samples=61, smooth]
                    plot (\x, {\yb + 1.1*((abs(\x-\c)/0.5 < 0.5) ?
                        (0.75 - (abs(\x-\c)/0.5)^2) :
                        (0.5*max(0, 1.5 - abs(\x-\c)/0.5)^2))});
            }
            \foreach \c in {0.5,1.0,1.5} {
                \pgfmathsetmacro{\ca}{\c-0.75}
                \pgfmathsetmacro{\cb}{\c+0.75}
                \draw[cblue, thick, domain=\ca:\cb, samples=61, smooth]
                    plot (\x, {\yb + 1.1*((abs(\x-\c)/0.5 < 0.5) ?
                        (0.75 - (abs(\x-\c)/0.5)^2) :
                        (0.5*max(0, 1.5 - abs(\x-\c)/0.5)^2))});
            }
            \draw[cred, dashed] (1.15,\yb) -- (1.15,1.35);
            \foreach \w in {0.66,0.32,0.02} {
                \fill[cred] (1.15, \yb + 1.1*\w) circle (1.2pt);
            }
            \node[font=\footnotesize, anchor=east] at (-0.1, \yb + 0.55) {$N_{\ell,i}(\xi)$};
            \node[font=\footnotesize, anchor=north] at (1.25, \yb - 0.15)
                {$\bm{f}_{\ell}(\xib) = \sum_{i,j} N_{\ell,i}(\xi)\, N_{\ell,j}(\eta)\, \bm{G}_{\ell,ij}$};
        \end{scope}
        \draw[decorate, decoration={brace, amplitude=6pt}, black!60]
            (5.3, 3.35) -- (5.3, -0.05);
        \draw[-{Stealth[length=2mm]}, semithick, black!60] (5.65, 1.65) -- (7.2, 1.65);
        \node[font=\footnotesize] (F) at (1.6,-2.3)
            {$\Fenc(\xib) = \bigl(\bm{f}_{1}, \dots, \bm{f}_{N_{\mathrm{lev}}}\bigr)$};
        \draw[arr] ($(q0) + (0,-0.12)$) -- (F.north -| q0);
        \node[box] (N) at (6.0,-2.3) {network\\with $(\xib, \Fenc)$};
        \node[box] (out) at (10.0,-2.3) {output};
        \draw[arr] (F) -- (N);
        \draw[arr] (N) -- (out);
    \end{tikzpicture}
    \caption{Field representation of the proposed method. The feature
    grids of the levels, from coarse to fine (top left), are read at the
    same point $\xib$ (red). On each level (top right) the feature at the
    point is the quadratic B-spline blend of the $3 \times 3$ nodes around
    it (blue), weighted by products of the one-dimensional basis functions
    $N_{\ell,i}$ of Eq.~\eqref{eq:bspline-te}. The features of all levels
    form the vector $\Fenc(\xib)$, which the network takes together with
    $\xib$ (bottom). The mechanical and the temperature networks share
    this structure, each with feature grids of its own.}
    \label{fig:encoding}
\end{figure}

The displacement and the phase field are represented as
in~\paperone\ by one network on the parametric domain
$\widehat{\Omega} = [0,1]^{2}$ with the coordinates
$\xib = (\xi, \eta)$,
\begin{equation}
    \netM : \bigl(\xib,\, \Fenc(\xib);\, \bm{\theta}_{\mathrm{M}}\bigr)
    \;\longmapsto\;
    \bigl(\hat{u},\, \hat{v},\, \hat{\pf}\bigr) ,
    \qquad
    \Fenc(\xib)
    =
    \bigl( \bm{f}_{1}(\xib), \dots, \bm{f}_{N_{\mathrm{lev}}}(\xib) \bigr) ,
    \label{eq:net-te}
\end{equation}
a multilayer perceptron with smooth Gaussian error linear unit (GELU) activations~\cite{hendrycks2016gelu} and the parameters $\bm{\theta}_{\mathrm{M}}$. The perceptron returns the raw outputs $\hat{u}$,
$\hat{v}$ and $\hat{\pf}$ and is fed with the coordinates and with the
feature vector $\Fenc$ of $N_{\mathrm{lev}}$ levels of trainable
feature grids. The subscript M marks the mechanical network and T the
temperature network introduced below. The parameters comprise the
weights of the perceptron and the feature values of the grids. Level $\ell$ stores a feature vector
$\bm{G}_{\ell,ij}$ at each node $(i, j)$ of a uniform grid with
$n_{\ell}$ nodes along $\eta$ (\ref{app:hyper}). The feature
vector of the level at a point is the quadratic B-spline blend of the nodal vectors at the $3 \times 3$ nodes around it,
\begin{equation}
    \bm{f}_{\ell}(\xib)
    =
    \sum_{i,j} N_{\ell,i}(\xi)\, N_{\ell,j}(\eta)\, \bm{G}_{\ell,ij} ,
    \label{eq:bspline-te}
\end{equation}
where $N_{\ell,i}$ and $N_{\ell,j}$ are the quadratic B-splines of level
$\ell$ centered on node $i$ along $\xi$ and on node $j$ along $\eta$,
three of which are nonzero in each direction at any point. Fig.~\ref{fig:encoding} shows the
construction. The blend makes every feature $C^{1}$ in the coordinates, so with the smooth activation the strains and the gradient of the phase field taken by automatic differentiation are continuous (a linear blend gives $C^{0}$ features). The spacing of the finest grid is the smallest scale the
representation can express, uniform over the parametric domain and
fixed before training. What concentrates on the crack is the density
of the integration points (Section~\ref{subsec:estimator-te}). The
grids are a tensor-product parameterization of the field, on which no
connectivity is built and to which no integration rule is tied. Curved
domains enter through an exact non-uniform rational
B-spline (NURBS) map of the parametric square, as in isogeometric
analysis~\cite{hughes2005} and in~\paperone.

The raw outputs become fields that satisfy the essential conditions
exactly. Each displacement component is written as
$U_{\mathrm{ref}}\, \omega_{\mathrm{M}}\, \hat{u} + u_{\mathrm{D}}$, with $\hat{u}$
the raw output of that component, $U_{\mathrm{ref}}$ a displacement
scale, $\omega_{\mathrm{M}}$ an envelope that
vanishes where the displacement is prescribed and $u_{\mathrm{D}}$ a
lift, a function that carries the prescribed values, so the essential
conditions hold for any network parameters. A displacement fixed at
a single point is imposed instead by subtracting the network value
there. The phase field is
assembled around a seeded notch,
\begin{equation}
    \pf(\xib)
    =
    \pf_{\mathrm{notch}}(\xib) + \bigl[ 1 - \pf_{\mathrm{notch}}(\xib) \bigr]\,
    s\bigl(\hat{\pf}(\xib)\bigr) ,
    \qquad
    s(\hat{\pf}) = \frac{1}{1 + e^{-\hat{\pf}}} ,
    \label{eq:phi-te}
\end{equation}
where $s$ is the squash function and $\pf_{\mathrm{notch}} = \exp(-r / \lreg)$ is
the optimal profile of the AT2 density about the notch at the distance
$r$ from it, with $\pf_{\mathrm{notch}} = 0$ where no notch is seeded.
Since $s$ is positive, the intact state $\pf = 0$ is approached but
not represented exactly (\ref{app:hyper}).

The temperature varies steeply within a boundary layer whose thickness may change by more than an order of magnitude over a transient, while the crack
band has the fixed width set by $\lreg$. The temperature is therefore
represented by a second network $\netT$ with a narrower perceptron,
parameters $\bm{\theta}_{\mathrm{T}}$ disjoint from
$\bm{\theta}_{\mathrm{M}}$ and feature grids of its own. Without feature
grids the early layer may not be resolved (\ref{app:sampler}). With their
feature vector $\FencT$ the temperature of step $n$ is
\begin{equation}
    \Tf_{n}(\xib)
    =
    T_{\mathrm{D}}(\xib, t_{n})
    + \omega_{\mathrm{T}}(\xib)\, \dT_{\mathrm{ref}}\,
    \netT\bigl(\xib, \FencT(\xib);\, \bm{\theta}_{\mathrm{T},n}\bigr) ,
    \label{eq:T-ansatz}
\end{equation}
where the lift $T_{\mathrm{D}}$ extends the prescribed temperatures of $\GT$
into the domain and $\omega_{\mathrm{T}}$ vanishes on $\GT$. The
temperature scale $\dT_{\mathrm{ref}}$ normalizes the raw output to
order one (\ref{app:hyper}). The network takes neither the time nor an explicit crack geometry as input, and the
temperature of step $n$ follows from the lift at $t_{n}$ and from the
parameters $\bm{\theta}_{\mathrm{T},n}$ obtained in that step. Where no boundary temperature is prescribed, $T_{\mathrm{D}} = \Tz$ and $\omega_{\mathrm{T}} = 1$. On the notched plates $T_{\mathrm{D}}$ interpolates linearly between the prescribed temperatures of the bottom and the top edge (\ref{app:hyper}). Where a surface is held at $\Tenv$ from a uniform initial temperature $\Tz$,
\begin{equation}
    T_{\mathrm{D}} = \Tenv + (\Tz - \Tenv)\, \omega_{\mathrm{T}} ,
    \qquad
    \omega_{\mathrm{T}}
    = \tanh\bigl( r_{1} / r_{\mathrm{T}} \bigr)\,
      \tanh\bigl( r_{2} / r_{\mathrm{T}} \bigr) ,
    \label{eq:T-lift}
\end{equation}
written for two cooled surfaces, with $r_{1}$ and
$r_{2}$ the physical distances to them and the decay length $r_{\mathrm{T}}$ chosen
well below the thickness of the boundary layer at the first time step.
The lift holds the cooled surfaces at $\Tenv$ from the outset, so the
first backward Euler step forms the boundary layer from the initial jump. An analytical temperature field is set as $T_{\mathrm{D}}$ with $\omega_{\mathrm{T}} = 0$.

\FloatBarrier
\subsection{Monte Carlo integration}
\label{subsec:estimator-te}

The loss and the thermal functional are both estimated in the same way as the energy in~\paperone. For any sampling density $p > 0$ on
$\widehat{\Omega}$, an integral over the domain is the expectation of its integrand times $\detJ$ divided by the density, with $\Jgeo$ the Jacobian of the geometry map. With $\psitot = \psihyb + \psi_{\mathrm{c}} + \psiir$ the integrand of the loss in Eq.~\eqref{eq:incremental-te} and $M$ points
drawn from $p$,
\begin{equation}
    \mathcal{L}(\bm{\theta}_{\mathrm{M}})
    =
    \Ex_{\xib \sim p}
    \left[ \frac{\psitot\, \detJ}{p} \right]
    \approx
    \widehat{\mathcal{L}}(\bm{\theta}_{\mathrm{M}})
    =
    \frac{1}{M} \sum_{i=1}^{M}
    \frac{\psitot(\xib_{i}; \bm{\theta}_{\mathrm{M}})\, \detJ(\xib_{i})}
         {p(\xib_{i})} ,
    \qquad
    \xib_{i} \sim p ,
    \label{eq:mc-te}
\end{equation}
where $\Ex_{\xib \sim p}$ denotes the expectation over points drawn from
$p$, and the estimator is
unbiased for any density that is positive and evaluated exactly at the
drawn points. Points drawn from the same density,
independently at every iteration of the thermal solve, give the
estimator of the thermal functional,
\begin{equation}
    \widehat{\EneT}(\bm{\theta}_{\mathrm{T}})
    =
    \frac{1}{M} \sum_{i=1}^{M}
    \frac{\psi_{\mathrm{T}}(\xib_{i}; \bm{\theta}_{\mathrm{T}})\, \detJ(\xib_{i})}
         {p(\xib_{i})}
    +
    \sum_{e \in \Gcov}
    \frac{1}{M_{e}} \sum_{j=1}^{M_{e}}
    \tfrac{1}{2}\, \hconv\,
    \bigl( \Tf(\xib_{j}^{e}) - \Tenv \bigr)^{2}\,
    \Bigl| \frac{\partial \xb}{\partial \zeta} \Bigr|_{j} ,
    \label{eq:mcT}
\end{equation}
where $\psi_{\mathrm{T}}$ is the thermal density of Eq.~\eqref{eq:piT}
and the second sum runs over the convective edges $e$, each with
$M_{e}$ points $\xib_{j}^{e}$ drawn uniformly in its parameter $\zeta$ and
weighted by the arc length per unit parameter. The importance weights
$\detJ / p$ and the weights for the arc length are treated as data and
excluded from differentiation.

The interior points are drawn from the mixture of three strata
of~\paperone,
\begin{equation}
    \begin{gathered}
    p
    =
    w_{\mathrm{u}}\, p_{\mathrm{unif}}
    + w_{\mathrm{c}}\, p_{\mathrm{crack}}
    + w_{\mathrm{p}}\, p_{\mathrm{proc}} ,
    \\
    p_{\mathrm{proc}}
    \propto
    \pf_{n-1} (1 - \pf_{n-1})
    + \beta_{0}\, \pf_{\mathrm{notch}}
    + \beta_{\mathrm{d}}\, Y_{\mathrm{proc}} ,
    \end{gathered}
    \label{eq:mixture-te}
\end{equation}
with stratum weights that sum to one, $w_{\mathrm{u}} + w_{\mathrm{c}}
+ w_{\mathrm{p}} = 1$, and each stratum a normalized density weighted
by $\detJ$, so that the uniform stratum is uniform in the physical
domain. The crack stratum is proportional to
$\max(\pf_{\mathrm{notch}}, \pf_{n-1})$, and where no notch is seeded it is instead a fixed band along the cooled faces. The process-zone stratum, which follows the growing cracks, peaks on the flanks of the crack band and ahead of the tip, with the weights $\beta_{0}$ of the notch profile and $\beta_{\mathrm{d}}$ of the driving-force indicator $Y_{\mathrm{proc}}$. The indicator is a tensile strain energy of
the previous state evaluated on its elastic strain, since under a
thermal load the total strain does not measure the stress
(\ref{app:hyper}). The crack and process-zone strata are piecewise
constant on an auxiliary grid of cells, so their densities are known
exactly at every point, and since the uniform stratum keeps $p > 0$,
the estimator stays unbiased for the mechanical loss and the thermal
functional alike, which \ref{app:sampler} checks on the thermal solve.
All strata are frozen within a
step and rebuilt from the accepted state.

The points are redrawn at every optimizer iteration, as in the deep
Ritz method~\cite{e2018deepritz}, for the thermal problem as for the
mechanical one. On a fixed point set the optimizer lowers the estimated
functional at the points while the field deteriorates between
them~\cite{rivera2022}, as~\paperone\ measured on the purely elastic
problem and \ref{app:sampler} measures on the thermal one.

\FloatBarrier
\subsection{Solution of a load step}
\label{subsec:staggered}

Since the temperature of a step does not depend on its mechanical
solution (Section~\ref{subsec:oneway}), each load step solves the two
problems in sequence,
\begin{equation}
    \begin{aligned}
    \bm{\theta}_{\mathrm{T},n}
    &=
    \mathcal{U}_{\mathrm{T},n}\bigl(\bm{\theta}_{\mathrm{T},n-1};\, \Tp,\, \pf_{n-1}\bigr) ,
    \\
    \bm{\theta}_{\mathrm{M},n}
    &=
    \mathcal{U}_{\mathrm{M},n}\bigl(\bm{\theta}_{\mathrm{M},n-1};\, \Tf_{n},\, \pf_{n-1},\,
    \delta_{n}\bigr) .
    \end{aligned}
    \label{eq:staggered}
\end{equation}
The temperature at $t_{n}$ is found first by the thermal update
$\mathcal{U}_{\mathrm{T},n}$, the iterations of Adam on
$\widehat{\EneT}$ over $\bm{\theta}_{\mathrm{T}}$ at the frozen phase
field $\pf_{n-1}$, starting from the parameters of the previous step,
with the temperature $\Tp$ supplied by a frozen copy of that step's network. The displacement
and the phase field are then advanced by the mechanical update
$\mathcal{U}_{\mathrm{M},n}$, the iterations of Adam from
$\bm{\theta}_{\mathrm{M},n-1}$ on the derivatives of the estimate $\widehat{\mathcal{L}}$ of the loss of Eq.~\eqref{eq:incremental-te}, and its result
is the accepted state of the step. The thermal strain is taken as
\begin{equation}
    \ath = \sg{\alphaT\, (\Tf_{n} - \Tz)} ,
    \label{eq:ath-sg}
\end{equation}
so that the mechanical gradient does not pass through the temperature
network and the temperature enters the mechanical loss as data. No
state is stored at integration points, since $\Tp$, $\pf_{n-1}$ and the
stratum densities are fields that can be evaluated wherever the sampler
draws next.

Each solve stops when the relative range of its estimated loss or
functional over a trailing window of iterations falls below a
tolerance, or when its iteration budget, the maximum number of
iterations, is spent (\ref{app:hyper}). The test measures the noise of the estimate as well as its drift and is not a residual test, so the result of a step depends on the budget, which is set per example. On
the strip every step uses its full budget, which is chosen in
\ref{app:increment}.

\begin{figure}[tb]
    \centering
    \begin{tikzpicture}[
        box/.style  = {rectangle, rounded corners=2pt, draw, align=center,
                       inner sep=4pt, font=\footnotesize, text width=50mm,
                       minimum height=11mm},
        tbox/.style = {box, fill=corange!8, draw=corange!70!black},
        mbox/.style = {box, fill=cblue!8, draw=cblue!70!black},
        fbox/.style = {box, fill=black!4, draw=black!60, text width=112mm,
                       minimum height=0mm},
        hdr/.style  = {font=\small\bfseries, align=center},
        arr/.style  = {-{Stealth[length=2.2mm]}, semithick},
        lbl/.style  = {font=\scriptsize\itshape, align=center}
    ]
        \def\xt{-3.6}   \def\xm{3.6}   \def\dy{1.95}
        \node[box, fill=black!4, draw=black!60, text width=112mm, minimum height=0mm]
            (start) at (0, 2.85)
            {start: lift temperature $T_{\mathrm{D}}$ and seeded notch at
             $t_{0} = 0$, load step $n = 1$};
        \node[fbox] (frozen) at (0, 1.6)
            {frozen state of step $n{-}1$: temperature $\Tp$, phase field
             $\pf_{n-1}$, stratum densities $p$ and the driving force
             with the thermal strain of $\Tp$};
        \node[tbox] (t1) at (\xt, 0*\dy - 0.55)
            {redraw $\{\xib_{i}\} \sim p$ and the boundary points
             on $\Gcov$};
        \node[tbox] (t2) at (\xt, -1*\dy - 0.55)
            {$\Tf = T_{\mathrm{D}} + \omega_{\mathrm{T}}\, \dT_{\mathrm{ref}}\,
             \netT(\xib, \FencT)$\\
             own feature grids, no crack geometry};
        \node[tbox] (t3) at (\xt, -2*\dy - 0.55)
            {estimate $\EneT$ by Eq.~\eqref{eq:mcT}\\
             Adam step on $\bm{\theta}_{\mathrm{T}}$};
        \node[tbox] (t4) at (\xt, -3*\dy - 0.55)
            {at plateau or budget: freeze $\netT$,
             $\ath = \sg{\alphaT (\Tf_{n} - \Tz)}$};
        \node[mbox] (m1) at (\xm, -1*\dy - 0.55)
            {redraw $\{\xib_{i}\} \sim p$};
        \node[mbox] (m2) at (\xm, -2*\dy - 0.55)
            {$(\ub, \pf)$ from $\netM(\xib, \Fenc)$ with the lifts and
             the seeded profile};
        \node[mbox] (m3) at (\xm, -3*\dy - 0.55)
            {estimate $\mathcal{L}(\ub, \pf;\, \Tf_{n})$ on
             $\straintensor_{\mathrm{e}} = \straintensor - \ath \bm{I}$,
             $\Gc(\Tf_{n})$\\
             Adam step on $\bm{\theta}_{\mathrm{M}}$};
        \node[mbox, below=6mm of m3] (m4)
            {at plateau or budget:\\ reaction force on a dedicated\\
             sample, save the accepted state};
        \draw[arr] (frozen.south -| t1.north) --
            node[hdr, fill=white, inner sep=2pt] {thermal problem, parameters $\bm{\theta}_{\mathrm{T}}$}
            (t1.north);
        \draw[arr] (frozen.south -| m1.north) --
            node[hdr, fill=white, inner sep=2pt, pos=0.62] {mechanical problem, parameters $\bm{\theta}_{\mathrm{M}}$}
            (m1.north);
        \draw[arr] (t1) -- (t2);
        \draw[arr] (t2) -- (t3);
        \draw[arr] (t3) -- (t4);
        \draw[arr] (m1) -- (m2);
        \draw[arr] (m2) -- (m3);
        \draw[arr] (m3) -- (m4);
        \draw[arr] (t3.west) .. controls +(-8mm,3mm) and +(-8mm,-3mm) .. (t1.west);
        \node[lbl, anchor=east] at ($(t1.west)!0.5!(t3.west) + (-7mm,0)$)
            {every\\iteration};
        \draw[arr] (m3.east) .. controls +(8mm,3mm) and +(8mm,-3mm) .. (m1.east);
        \node[lbl, anchor=west] at ($(m1.east)!0.5!(m3.east) + (7mm,0)$)
            {every\\iteration};
        \draw[arr] (t4.east) -- node[lbl, above, yshift=1pt] {$\ath$ as data}
            node[lbl, below, yshift=-1pt] {no gradient} (m3.west);
        \draw[arr] (start) -- (frozen);
        \draw[arr, dashed] (m4.south) -- ++(0,-5mm) -| (-8.3, 1.6) -- (frozen.west);
        \node[lbl, anchor=north] at ($(m4.south) + (-\xm, -5mm)$)
            {if $n < n_{\max}$: $n \leftarrow n{+}1$, warm start of both networks};
        \node[box, fill=black!4, draw=black!60, text width=7mm, minimum height=0mm,
              right=15mm of m4] (stop) {end};
        \draw[arr] (m4.east) -- node[lbl, above] {$n = n_{\max}$} (stop.west);
    \end{tikzpicture}
    \caption{One load step of the thermo-mechanical solver. The thermal
    functional (orange) is minimized and the mechanical fields (blue) are
    advanced, the two updates of Eq.~\eqref{eq:staggered} running in
    sequence, each on its own parameters, on points redrawn at every iteration. The
    temperature enters the mechanical loss as the thermal strain of
    Eq.~\eqref{eq:ath-sg}, through which no gradient passes, and the
    accepted state becomes the frozen state of the next step (dashed),
    until the last load step $n_{\max}$ ends the run.}
    \label{fig:workflow}
\end{figure}
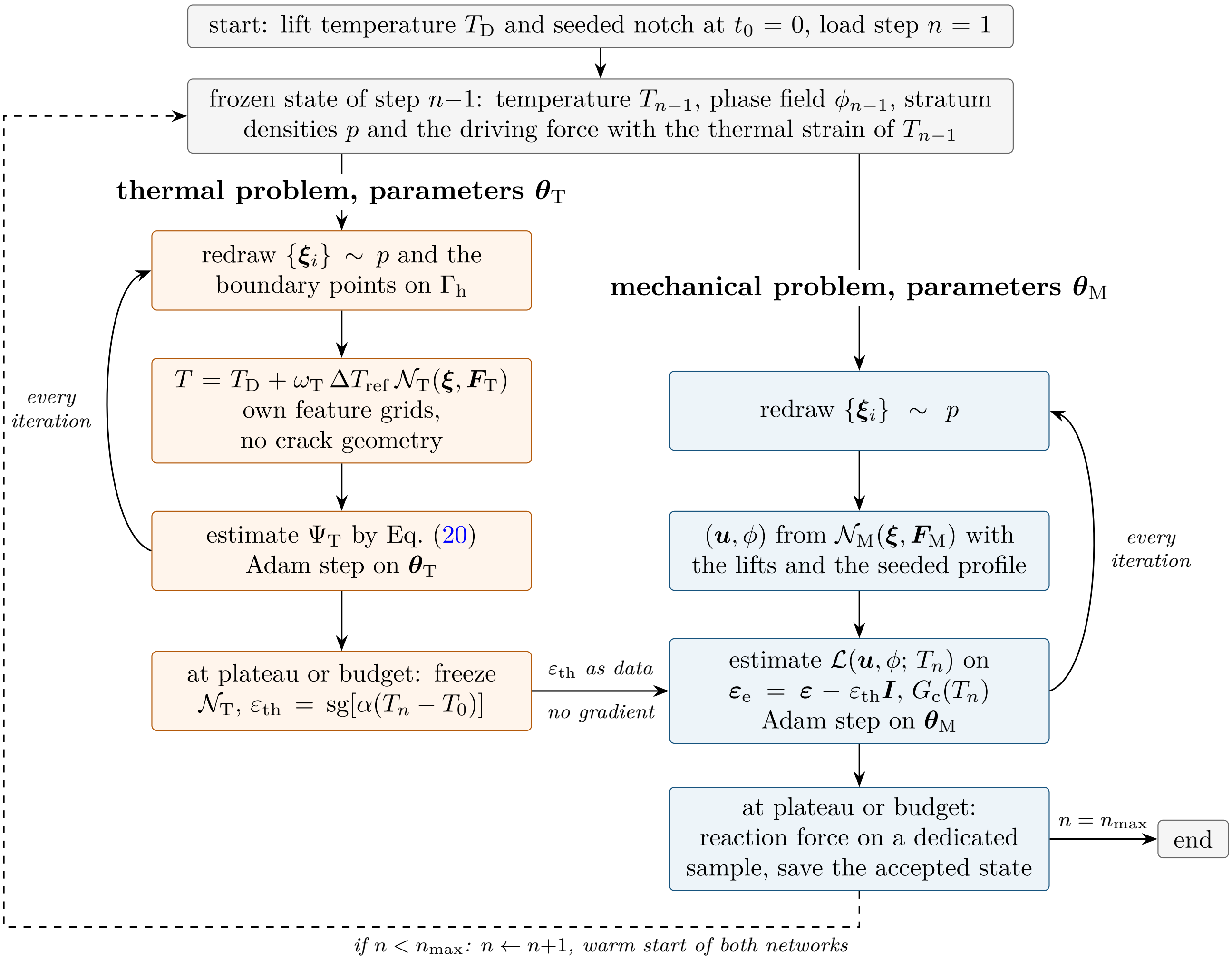

At $t_{0} = 0$ the network $\netT$, with its output layer initialized
at zero, returns zero, so the initial temperature is the lift
$T_{\mathrm{D}}$, which needs no training. The mechanical network is
initialized and trained in the first step as in~\paperone. Both
solves use Adam only, since the redraw changes the estimate of the
objective at every iteration and a line search would act on the noise
of one draw.
Fig.~\ref{fig:workflow} shows one load step, with what each solve
receives from the frozen state and what it passes on.

Two options of the mechanical update are introduced here for problems in
which many cracks nucleate under a uniform load. In the update of~\paperone\ the
network parameters are carried across load steps while Adam is
restarted. With its running averages of the gradient reset, Adam moves
every parameter by about the learning rate in the first iterations,
which perturbs the phase field throughout the domain at the start of
every step. The first option carries the optimizer state
across load steps. In the second, the learning rates decay
exponentially over the budget of each step to $10\%$ of their initial
values, so each step ends with small updates, and the rates return to
their initial values at the next step, a learning rate schedule with warm restarts~\cite{loshchilov2017}. The two options are used together and
with the crack-set threshold of Eq.~\eqref{eq:penalty-te}, the only combination tested on the slab that kept its bands at a nearly constant width and its cooled face free of a damaged film (\ref{app:arrays}).

At the end of a step the reaction force on a loaded edge is the
derivative of the loss with respect to the prescribed displacement,
taken with the network parameters held at their accepted values,
\begin{equation}
    F_{n}
    =
    \left. \frac{\partial \mathcal{L}_{n}(\bm{\theta}_{\mathrm{M}}, \delta)}{\partial \delta}
    \right|_{\bm{\theta}_{\mathrm{M}} = \bm{\theta}_{\mathrm{M},n},\; \delta = \delta_{n}} ,
    \label{eq:observables}
\end{equation}
where $\delta$ is the component of the prescribed displacement that
loads the edge, and $\mathcal{L}_{n}(\bm{\theta}_{\mathrm{M}}, \delta)$
is the loss of Eq.~\eqref{eq:incremental-te} with the fields of the
network and the displacement lift of $\delta$. Since the lift depends
on $\delta$ while the parameters stay fixed, and the stop-gradient
leaves only the degraded strain energy in this derivative, $F_{n}$ is
the work conjugate of $\delta$, which equals the resultant of the hybrid stress in Eq.~\eqref{eq:stress-te} on the edge when equilibrium
holds (\ref{app:hyper}). It is evaluated on a dedicated sample several
times larger than the $M$ points of an iteration, as in~\paperone.

%% file: sections/S4_results.tex
\section{Numerical examples}
\label{sec:results}
\input{figures/geom_layout}

The two notched plates, under tension and under shear, are reported
first, followed by the directionally quenched glass strip, the quenched
ceramic slab and the quenched ceramic annulus, the last two being the
quench problems. The temperature is computed with a damage-dependent
conductivity on the notched plates, given as an analytical field on
the strip, and computed with a damage-independent conductivity on the slab and the annulus (Section~\ref{subsec:heat}). The cooled faces of the slab and the annulus carry a prescribed boundary temperature, and the slab under the convective condition is treated in \ref{app:convective}. The toughness depends on the temperature, through Eq.~\eqref{eq:gcT}, only on the notched plates.

The finite element computations use an independent staggered
phase-field solver built on scikit-fem~\cite{gustafsson2020scikitfem}
with the hybrid formulation, checked against the published results of
Tangella et al.~\cite{tangella2022} (\ref{app:reference}). Where the temperature is computed, the solver advances the heat conduction problem by backward Euler on the same time grid as the proposed method. It enforces irreversibility through the history field
on the notched plates and as a bound on the phase field on the strip
and the quench problems. It serves as the reference on the notched
plates and as a computation of the same model on the strip and the
quench problems, where ``the same model'' means the same phase-field
model up to the treatment of irreversibility and the representation of
the notch. On the notched plates the reference also omits the
temperature-dependent toughness and the corner toughening of
Section~\ref{subsec:sent}.
On the strip and the quench problems the bound prevents any decrease of the phase field, whereas the penalty of the proposed method acts only on the crack set, so where diffuse damage localizes the two solutions may differ. Section~\ref{subsec:slab} tests this difference on the slab.

Each problem is computed with one set of numerical parameters at all of
its loads, and these are listed in Table~\ref{tab:hyper}. The strip, the slab and the
annulus use the crack-set threshold $\pf_{\mathrm{th}} = 1/2$ together
with the optimizer state carried across load steps and the learning
rates decaying within a step (Section~\ref{subsec:staggered}), while the
notched plates use $\pf_{\mathrm{th}} = 0$ and the update
of~\paperone.

\subsection{Single-edge-notched tension test}
\label{subsec:sent}

\begin{figure}[htbp]
    \centering
    \includegraphics[width=\textwidth]{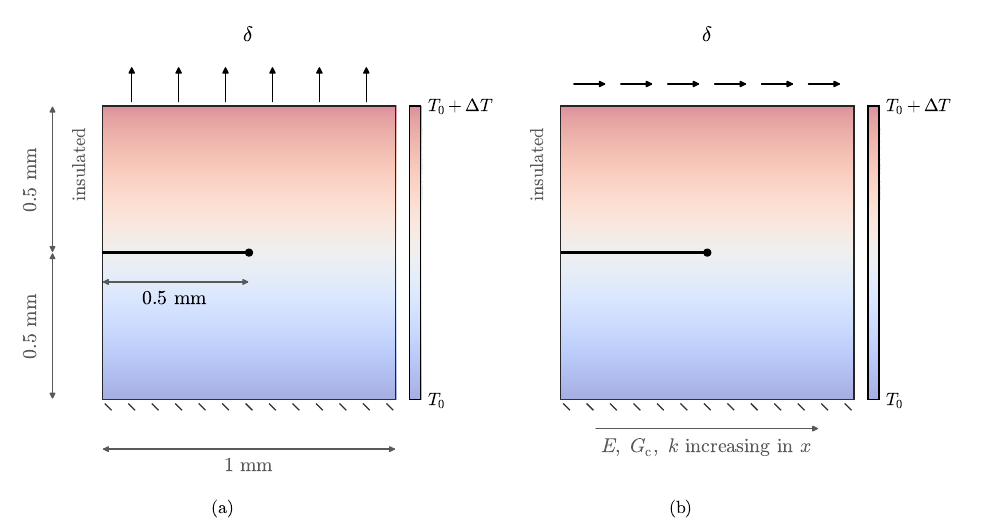}
    \caption{Geometry and boundary conditions of the two notched
    plates. (a)~Homogeneous plate under tension. (b)~Graded plate under
    shear, with stiffness, toughness, conductivity and density increasing from
    left to right. In both the bottom edge is fixed and held at $\Tz$,
    the top edge is displaced and brought to $\Tz + \dT$, and the
    lateral edges are free and insulated. Dimensions are marked on the
    first panel, and the temperature shading is schematic.}
    \label{fig:geom-plates}
\end{figure}

With this problem we verify the thermo-mechanical solver on the tension
specimen of~\paperone, here with a thermal load added. A square plate
of side $1$~mm carries a horizontal notch of length $0.5$~mm from the
left edge at mid-height, Fig.~\ref{fig:geom-plates}(a). The bottom edge is fixed and held at the initial temperature $\Tz = 300$~K, and the top edge is displaced
vertically by $\delta$ in $80$ steps of $10^{-5}$~mm while its
temperature $\Tf_{\mathrm{edge}}$ is ramped linearly to
$\Tz + \dT$ at step $20$ and then held, with $\dT = -50$, $0$ and
$+50$~K. The lateral edges are free and insulated. The material is the
ceramic of Tangella et al.~\cite{tangella2022} under plane strain,
Table~\ref{tab:mat-plates}, with the damage-dependent conductivity of
Eq.~\eqref{eq:kdeg}, since the crack crosses the heat flux, and the toughness law of
Eq.~\eqref{eq:gcT} with the constants of Ruan et
al.~\cite{ruan2023thermomechanical}. As in~\paperone, the toughness is
doubled within about $3\lreg$ of the four corners of the fixed and the displaced edges,
so that the singular driving force there does not nucleate damage. The displacement lift interpolates linearly between the bottom and the top edge (\ref{app:hyper}). Physical time is
$t = 1\,\mathrm{s\,mm^{-1}} \times \delta$, so the thermal boundary
layer under the driven edge reaches a thickness $\sqrt{D t}$ of $0.24$
to $0.33$~mm at the peak loads of the cooled and heated cases.

\begin{table}[htb]
    \centering
    \caption{Material data of the notched plates, after Tangella et
    al.~\cite{tangella2022}. The tension plate is homogeneous. The
    graded shear plate varies linearly in $x$ from the values of the
    homogeneous plate at $x = 0$ to those of the last column at
    $x = 1$~mm, taken from the released code of Tangella et
    al.~\cite{tangella2022code}, with $\nu$, $\alphaT$ and $c$ constant.
    The toughness law of Eq.~\eqref{eq:gcT} takes its constants from Ruan et al.~\cite{ruan2023thermomechanical}. The regularization
    length is chosen here, against $0.005$~mm in Tangella et al.}
    \label{tab:mat-plates}
    \footnotesize
    \begin{tabular}{lcccc}
        \toprule
        & & Homogeneous plate & \multicolumn{2}{c}{Graded plate} \\
        \cmidrule(lr){4-5}
        Property & Symbol & & $x = 0$ & $x = 1$~mm \\
        \midrule
        Young's modulus (GPa) & $E$ & 340 & 340 & 450 \\
        Poisson's ratio & $\nu$ & 0.22 & 0.22 & 0.22 \\
        Toughness (J\,m$^{-2}$) & $\Gcz$ & 42.47 & 42.47 & 120 \\
        Expansion coefficient (K$^{-1}$) & $\alphaT$ & $8.0 \times 10^{-6}$ & $8.0 \times 10^{-6}$ & $8.0 \times 10^{-6}$ \\
        Conductivity (W\,m$^{-1}$\,K$^{-1}$) & $\kcond$ & 300 & 300 & 350 \\
        Density (kg\,m$^{-3}$) & $\rho$ & 2450 & 2450 & 4000 \\
        Specific heat (J\,kg$^{-1}$\,K$^{-1}$) & $c$ & 775 & 775 & 775 \\
        Toughness law constants & $b_{1}$, $b_{2}$ & 1.80, 1.10 & 1.80, 1.10 & 1.80, 1.10 \\
        Toughness law temperatures (K) & $T_{\mathrm{ref}}$, $T_{\mathrm{max}}$ & 300, 1000 & 300, 1000 & 300, 1000 \\
        Regularization length (mm) & $\lreg$ & 0.01 & 0.01 & 0.01 \\
        \bottomrule
    \end{tabular}
\end{table}

\begin{figure}[htbp]
    \centering
    \begin{subfigure}[t]{0.56\linewidth}
        \centering
        \begin{tikzpicture}
            \begin{axis}[paperaxis, scale only axis, width=0.80\linewidth, height=6.2cm,
                         xlabel={$\delta$ ($10^{-3}$ mm)}, ylabel={$F$ (N)},
                         xmin=0, xmax=0.82, ymin=0, ymax=110,
                         legend columns=3, legend to name={leg:sent-fd},
                         legend style={/tikz/every even column/.append style={column sep=1.5ex}}]
                \addplot[cblue, thick, dashed, forget plot]
                    table[col sep=comma, x expr=\thisrow{delta_mm}*1000, y=F]
                    {figures/data/sent_ref_m50.csv};
                \addplot[black, thick, dashed, forget plot]
                    table[col sep=comma, x expr=\thisrow{delta_mm}*1000, y=F]
                    {figures/data/sent_ref_iso.csv};
                \addplot[cred, thick, dashed, forget plot]
                    table[col sep=comma, x expr=\thisrow{delta_mm}*1000, y=F]
                    {figures/data/sent_ref_p50.csv};
                \addplot[cblue, thick]
                    table[col sep=comma, x expr=\thisrow{delta_mm}*1000, y=F]
                    {figures/data/sent_prop_m50.csv};
                \addlegendentry{$\dT = -50$~K}
                \addplot[black, thick]
                    table[col sep=comma, x expr=\thisrow{delta_mm}*1000, y=F]
                    {figures/data/sent_prop_iso.csv};
                \addlegendentry{$\dT = 0$~K}
                \addplot[cred, thick]
                    table[col sep=comma, x expr=\thisrow{delta_mm}*1000, y=F]
                    {figures/data/sent_prop_p50.csv};
                \addlegendentry{$\dT = +50$~K}
            \end{axis}
        \end{tikzpicture}
        \caption{}
    \end{subfigure}\hfill
    \begin{subfigure}[t]{0.40\linewidth}
        \centering
        \begin{tikzpicture}
            \begin{axis}[paperaxis, scale only axis, width=0.72\linewidth, height=6.2cm,
                         xlabel={$\delta$ ($10^{-3}$ mm)}, ylabel={$F$ (N)},
                         xmin=0.30, xmax=0.76, ymin=85, ymax=105]
                \addplot[cblue, thick, dashed]
                    table[col sep=comma, x expr=\thisrow{delta_mm}*1000, y=F]
                    {figures/data/sent_ref_m50.csv};
                \addplot[black, thick, dashed]
                    table[col sep=comma, x expr=\thisrow{delta_mm}*1000, y=F]
                    {figures/data/sent_ref_iso.csv};
                \addplot[cred, thick, dashed]
                    table[col sep=comma, x expr=\thisrow{delta_mm}*1000, y=F]
                    {figures/data/sent_ref_p50.csv};
                \addplot[cblue, thick]
                    table[col sep=comma, x expr=\thisrow{delta_mm}*1000, y=F]
                    {figures/data/sent_prop_m50.csv};
                \addplot[black, thick]
                    table[col sep=comma, x expr=\thisrow{delta_mm}*1000, y=F]
                    {figures/data/sent_prop_iso.csv};
                \addplot[cred, thick]
                    table[col sep=comma, x expr=\thisrow{delta_mm}*1000, y=F]
                    {figures/data/sent_prop_p50.csv};
            \end{axis}
        \end{tikzpicture}
        \caption{}
    \end{subfigure}

    \vspace{2pt}
    \pgfplotslegendfromname{leg:sent-fd}
    \caption{Notched plate under tension. Load-displacement curves of the
    proposed method against the finite element reference at three thermal
    loads. (a)~Full response. (b)~Detail of the initiation peak. The
    cooled case is blue, the isothermal case black and the heated case
    red, the reference dashed and the proposed method solid.}
    \label{fig:sent-fd}
\end{figure}
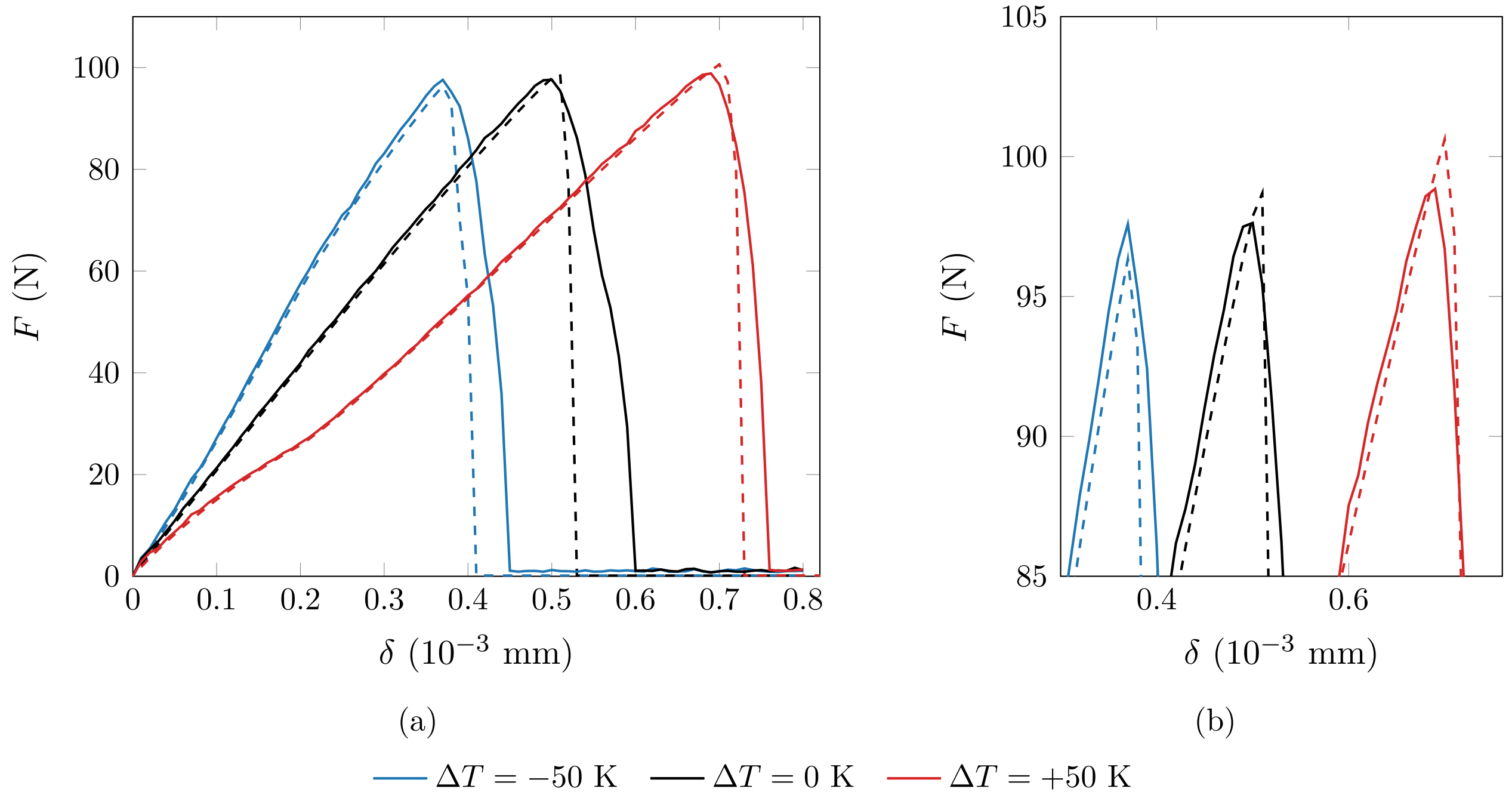

Fig.~\ref{fig:sent-fd} compares the computed load-displacement curves
with the reference at the three thermal loads, and
Fig.~\ref{fig:sent-fd}(b) enlarges the initiation peak. The peak loads
agree with the reference to within $1.30\%$, $1.08\%$ and $1.76\%$ in
the cooled, isothermal and heated cases, and the displacement at the
peak to within one load step. The isothermal case, which has no
thermal load, agrees about as closely as the purely elastic computation
of~\paperone, which agreed to within about $1\%$ on the same test and material.

The reference peaks rise by $4.5\%$ from the cooled to the heated case,
a rise that the proposed method, whose peaks lie within $1.3\%$ of one
another, does not reproduce. The toughness law that the reference omits
does not explain this difference, since it changes the reference peak
of the cooled case by $0.23\%$ (\ref{app:reference}). The thermal load changes the reference peak far less on this plate than on the graded plate, by $4.5\%$ against $28\%$. The thermal solve is therefore verified here through the temperature field, and the coupling through the peak load on the graded plate of Section~\ref{subsec:sensfgm}.

\begin{figure}[htbp]
    \centering
    \includegraphics[width=0.86\textwidth]{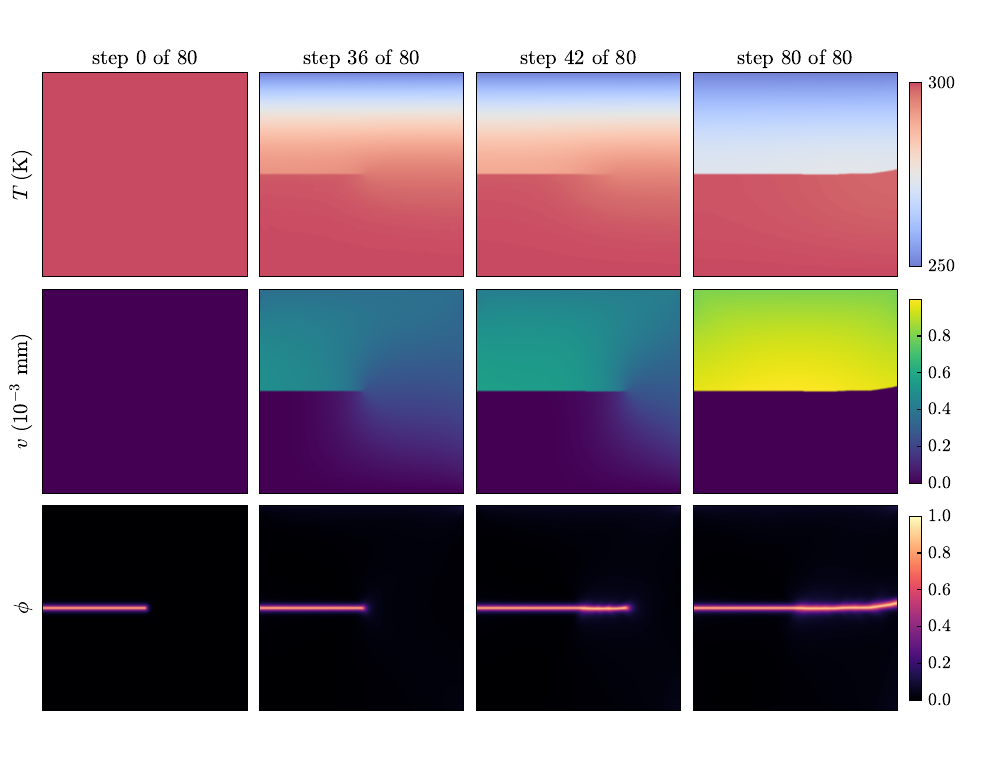}
    \caption{Notched plate under tension at $\dT = -50$~K, the cooled
    edge at the top. Temperature, vertical displacement and phase field
    of the proposed method at load steps $0$, $36$, $42$ and $80$ of
    $80$, the initial state, the last step before the peak load, a step during the
    propagation and the final state. Each row shares one color scale.
    The region below the crack stays at $\Tz$ once the crack has
    crossed, since the crack faces are nearly adiabatic.}
    \label{fig:sent-fields}
\end{figure}

Fig.~\ref{fig:sent-fields} shows the temperature, the vertical
displacement and the phase field of the proposed method from the start
to failure. In the cooled case the crack bends slightly toward the
cooled edge near the right side, as in the reference, whereas the
isothermal crack runs straight.

\begin{figure}[!htb]
    \centering
    \includegraphics[width=0.40\textwidth]{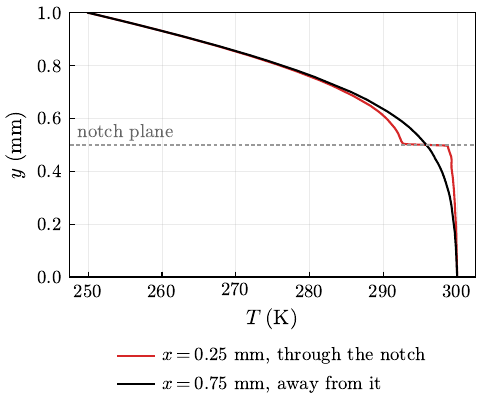}
    \caption{Notched plate under tension at $\dT = -50$~K,
    temperature of the proposed method along two vertical lines at the
    step before the peak load, through the notch and away from it. The
    jump on the notch plane is the effect of the nearly adiabatic notch faces.}
    \label{fig:notch-profile}
\end{figure}

Away from the notch the temperature depends on $y$ and $t$ alone, and
along a vertical line it agrees with an implicit finite difference
solution of the one-dimensional transient to within $0.44$~K until the
crack crosses the plate (\ref{app:hyper}). Across the notch it changes
by a few kelvin over the width of the phase-field band,
Fig.~\ref{fig:notch-profile}, since the damage-dependent conductivity of Eq.~\eqref{eq:kdeg} makes the notch faces nearly adiabatic, and the temperature
network represents this jump with no crack geometry in its inputs.

\subsection{Single-edge-notched shear test with functionally graded material}
\label{subsec:sensfgm}

As the next example, the same notch is loaded in shear on a
functionally graded plate, Fig.~\ref{fig:geom-plates}(b). This example
verifies the solver where the material properties, and with them the
thermal stress, vary across the specimen. The top edge is
displaced horizontally in $160$ steps of $2.5 \times 10^{-5}$~mm while
its temperature is ramped linearly to $\Tz + \dT$ at step $12$, with $\dT = -30$, $0$ or $+30$~K. The properties vary linearly in $x$ to the stiffer
and tougher end values of Table~\ref{tab:mat-plates}, in the reference
as well. Equation~\eqref{eq:gcT} is applied to the graded $\Gcz$, and the remaining settings are those of Section~\ref{subsec:sent}.

\begin{figure}[htbp]
    \centering
    \begin{subfigure}[t]{0.56\linewidth}
        \centering
        \begin{tikzpicture}
            \begin{axis}[paperaxis, scale only axis, width=0.80\linewidth, height=6.2cm,
                         xlabel={$\delta$ ($10^{-3}$ mm)}, ylabel={$F$ (N)},
                         xmin=0, xmax=3.5, ymin=0, ymax=125,
                         legend columns=3, legend to name={leg:sensfgm-fd},
                         legend style={/tikz/every even column/.append style={column sep=1.5ex}}]
                \addplot[cblue, thick, dashed, forget plot]
                    table[col sep=comma, x expr=\thisrow{delta_mm}*1000, y=F]
                    {figures/data/sensfgm_ref_m30.csv};
                \addplot[black, thick, dashed, forget plot]
                    table[col sep=comma, x expr=\thisrow{delta_mm}*1000, y=F]
                    {figures/data/sensfgm_ref_iso.csv};
                \addplot[cred, thick, dashed, forget plot]
                    table[col sep=comma, x expr=\thisrow{delta_mm}*1000, y=F]
                    {figures/data/sensfgm_ref_p30.csv};
                \addplot[cblue, thick]
                    table[col sep=comma, x expr=\thisrow{delta_mm}*1000, y=F]
                    {figures/data/sensfgm_prop_m30.csv};
                \addlegendentry{$\dT = -30$~K}
                \addplot[black, thick]
                    table[col sep=comma, x expr=\thisrow{delta_mm}*1000, y=F]
                    {figures/data/sensfgm_prop_iso.csv};
                \addlegendentry{$\dT = 0$~K}
                \addplot[cred, thick]
                    table[col sep=comma, x expr=\thisrow{delta_mm}*1000, y=F]
                    {figures/data/sensfgm_prop_p30.csv};
                \addlegendentry{$\dT = +30$~K}
            \end{axis}
        \end{tikzpicture}
        \caption{}
    \end{subfigure}\hfill
    \begin{subfigure}[t]{0.40\linewidth}
        \centering
        \begin{tikzpicture}
            \begin{axis}[paperaxis, scale only axis, width=0.72\linewidth, height=6.2cm,
                         xlabel={$\delta$ ($10^{-3}$ mm)}, ylabel={$F$ (N)},
                         xmin=0.9, xmax=1.6, ymin=80, ymax=122]
                \addplot[cblue, thick, dashed]
                    table[col sep=comma, x expr=\thisrow{delta_mm}*1000, y=F]
                    {figures/data/sensfgm_ref_m30.csv};
                \addplot[black, thick, dashed]
                    table[col sep=comma, x expr=\thisrow{delta_mm}*1000, y=F]
                    {figures/data/sensfgm_ref_iso.csv};
                \addplot[cred, thick, dashed]
                    table[col sep=comma, x expr=\thisrow{delta_mm}*1000, y=F]
                    {figures/data/sensfgm_ref_p30.csv};
                \addplot[cblue, thick]
                    table[col sep=comma, x expr=\thisrow{delta_mm}*1000, y=F]
                    {figures/data/sensfgm_prop_m30.csv};
                \addplot[black, thick]
                    table[col sep=comma, x expr=\thisrow{delta_mm}*1000, y=F]
                    {figures/data/sensfgm_prop_iso.csv};
                \addplot[cred, thick]
                    table[col sep=comma, x expr=\thisrow{delta_mm}*1000, y=F]
                    {figures/data/sensfgm_prop_p30.csv};
            \end{axis}
        \end{tikzpicture}
        \caption{}
    \end{subfigure}

    \vspace{2pt}
    \pgfplotslegendfromname{leg:sensfgm-fd}
    \caption{Graded notched plate under shear. Load-displacement curves
    of the proposed method against the finite element reference at three
    thermal loads. (a)~Full response through initiation, arrest,
    reloading and failure. (b)~Detail of the initiation peak. The cooled
    case is blue, the isothermal case black and the heated case red,
    the reference dashed and the proposed method solid.}
    \label{fig:sensfgm-fd}
\end{figure}
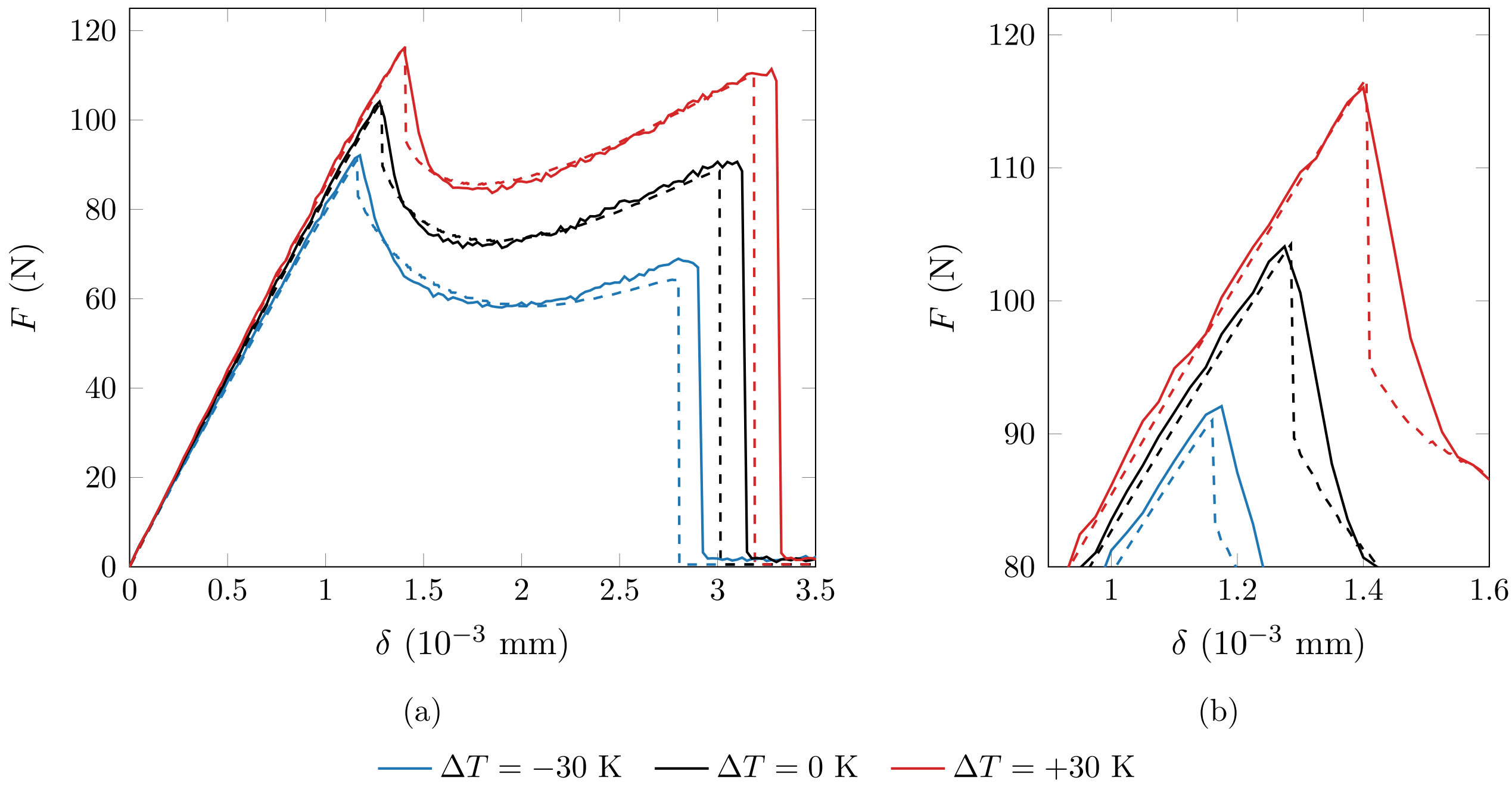

\begin{figure}[htbp]
    \centering
    \includegraphics[width=0.86\textwidth]{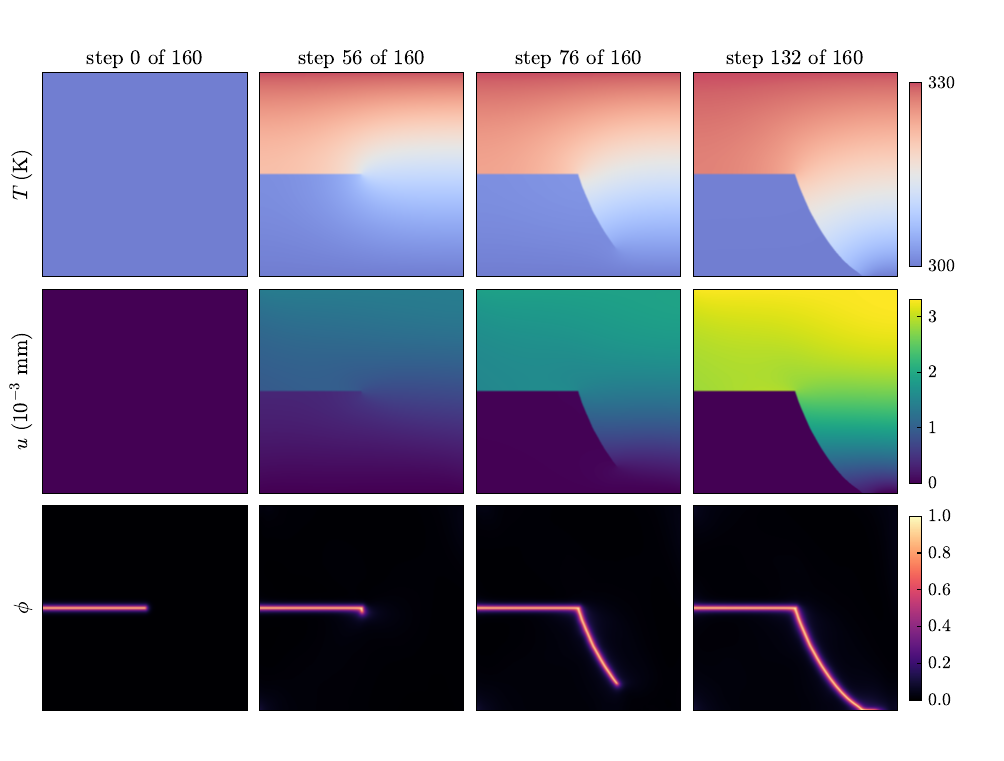}
    \caption{Graded notched plate under shear at $\dT = +30$~K, the
    heated edge at the top. Temperature, horizontal displacement and
    phase field of the proposed method at load steps $0$, $56$, $76$
    and $132$ of $160$, which are the start, the initiation peak, the
    step near the arrest trough and the step after the second maximum. Each row
    shares one color scale.}
    \label{fig:sensfgm-fields}
\end{figure}

Fig.~\ref{fig:sensfgm-fd} compares the load-displacement curves. The
crack initiates at the notch at the first peak, arrests as it reaches
the stiffer and tougher material, and the load rises to a second
maximum before the plate fails. Both solvers pass through these four
stages. The initiation peak, the global maximum of every curve,
agrees to within $1.18\%$, $0.14\%$ and $0.58\%$ in the cooled,
isothermal and heated cases, with the peak displacement within one load
step. The second maximum differs by at most $7.3\%$, the plate fails at a
displacement about $4\%$ larger than in the reference, and on the
reloading branch the force of the proposed method scatters by up to $2\%$ about a smooth curve, since the energy is estimated stochastically.
Fig.~\ref{fig:sensfgm-fields} shows the fields of the heated case at
these stages, with a downward kink of the crack at the notch tip.

The initiation peaks of the proposed method spread over $26\%$, from
$92.1$~N at $-30$~K to $116.0$~N at $+30$~K, against $28\%$ in the
reference. Since the displacement of the driven edge normal to the
notch is held at zero, the thermal expansion of the boundary layer
compresses the notch plane in the heated case and opens it in the
cooled one, which raises the shear load at initiation in the first and
lowers it in the second.

\subsection{Directionally quenched glass strip}
\label{subsec:strip}

With this problem we test the evolution of a crack path under an
analytical temperature field. A hot glass strip drawn at constant speed
into a cold bath carries a crack that follows the immersion front, and
as the temperature jump grows the path changes from straight to
oscillatory and then branches, while at larger jumps several cracks
start along the leading edge~\cite{yuse1993,yuse1997}. The
strip of Zeng and Dolbow~\cite{zeng2026}, Fig.~\ref{fig:geom-strip}, is
$30$ by $16$~mm with half-width $b = 8$~mm and made of the glass in Table~\ref{tab:mat-strip}. It is computed under plane stress and drawn
at $V = 0.4$~mm\,s$^{-1}$ into a bath at $\Tenv = 296$~K. Their model
carries a tensile strength of $80$~MPa independent of the
regularization length, whereas the AT2 model at $\lreg = 0.1$~mm has
the homogeneous peak stress
$\sigma_{\mathrm{c}} = \sqrt{27 E \Gc / (256\, \lreg)} = 24.7$~MPa~\cite{tanne2018}. The comparison is therefore between two models on one problem, and Zeng and Dolbow show that the strength also affects the path. The control
parameters are the P\'eclet number $P = bV/D = 6.81$ and the
thermal-shock number
$R = \alphaT E \dT \sqrt{b} / K_{\mathrm{c}} = 0.0634\,\dT$, with
$\dT$ in kelvin, where $\dT = \Tz - \Tenv$ is the jump from the
stress-free hot region to the bath and $K_{\mathrm{c}} = \sqrt{E \Gc}$.
This $R$ is not the resistance parameter of
Hasselman~\cite{hasselman1969}. The notch of Zeng and Dolbow, of length
$5$~mm at the leading edge and offset $0.25$~mm from the centerline, is
retained. The trailing edge carries zero horizontal
displacement and its midpoint zero vertical displacement, and all other
edges are traction free, so the thermal strain is the only load.

The profile of the front is steady in the frame of the bath, in which advection balances diffusion ahead of the front, and reads in the coordinates of the strip
\begin{equation}
    \Tf(x, t)
    =
    \Tenv + \dT
    \Bigl[ 1 - \exp\Bigl( -\frac{V}{D}\, \pospart{x - V t} \Bigr) \Bigr] ,
    \label{eq:front}
\end{equation}
with $x$ the coordinate along the strip. It is set as $T_{\mathrm{D}}$ in Eq.~\eqref{eq:T-ansatz} with $\omega_{\mathrm{T}} = 0$, so no thermal problem is solved, and it neglects the effect of the crack on the temperature.
Zeng and Dolbow and Corson et al.~\cite{corson2009} use the same profile. The
temperature jumps are $130$, $145$, $155$, $165$, $170$, $175$, $190$,
$217$ and $307$~K, and at each of them the front travels $24$~mm in
$240$ load steps, each advancing it by one regularization length. The increment and the
iteration budget are chosen in \ref{app:increment}.

The oscillatory path is measured by two order parameters, the amplitude
$A$ and the wavelength $\Lambda$, read from the ridge of the phase
field traced from the notch to the tip. With $(x_{q}, y_{q})$ the
successive turning points of this path, $Q$ in number,
\begin{equation}
    \Lambda
    =
    \frac{2}{Q - 1} \sum_{q=1}^{Q-1} (x_{q+1} - x_{q}) ,
    \qquad
    A
    =
    \tfrac{1}{2} \max_{q} \, |y_{q+1} - y_{q}| ,
    \label{eq:order-params}
\end{equation}
so that $\Lambda$ is the period of a nearly periodic path and $A$ half
the largest half swing $|y_{q+1} - y_{q}|$ between two successive
turning points. Both are read on the developed path, up to
$x = 18$~mm, which excludes the last quarter of the travel, where the
tip approaches the end of the strip (\ref{app:orderparams}).

\begin{table}[htb]
    \centering
    \caption{Material data of the glass strip, after Zeng and
    Dolbow~\cite{zeng2026}. The strip is computed under plane stress. The
    regularization length is chosen here, against $0.03$~mm in Zeng and
    Dolbow.}
    \label{tab:mat-strip}
    \footnotesize
    \begin{tabular}{lcc}
        \toprule
        Property & Symbol & Value \\
        \midrule
        Young's modulus (GPa) & $E$ & 70 \\
        Poisson's ratio & $\nu$ & 0.22 \\
        Toughness (J\,m$^{-2}$) & $\Gc$ & 8.25 \\
        Expansion coefficient (K$^{-1}$) & $\alphaT$ & $7.7 \times 10^{-6}$ \\
        Diffusivity (mm$^{2}$\,s$^{-1}$) & $D$ & 0.47 \\
        Regularization length (mm) & $\lreg$ & 0.1 \\
        \bottomrule
    \end{tabular}
\end{table}

\begin{figure}[htbp]
    \geomstripline
    \caption{Geometry of the glass strip, drawn at speed $V$ from the
    hot region into the bath. The shading and the curve are the analytical
    temperature of Eq.~\eqref{eq:front} at one instant, and the leading
    notch of length $5$~mm is offset $0.25$~mm from the centerline. The
    transition length, the offset and the width are not to scale.}
    \label{fig:geom-strip}
\end{figure}

\begin{figure}[!tbp]
    \centering
    \begin{subfigure}[t]{0.32\linewidth}
        \includegraphics[width=\linewidth]{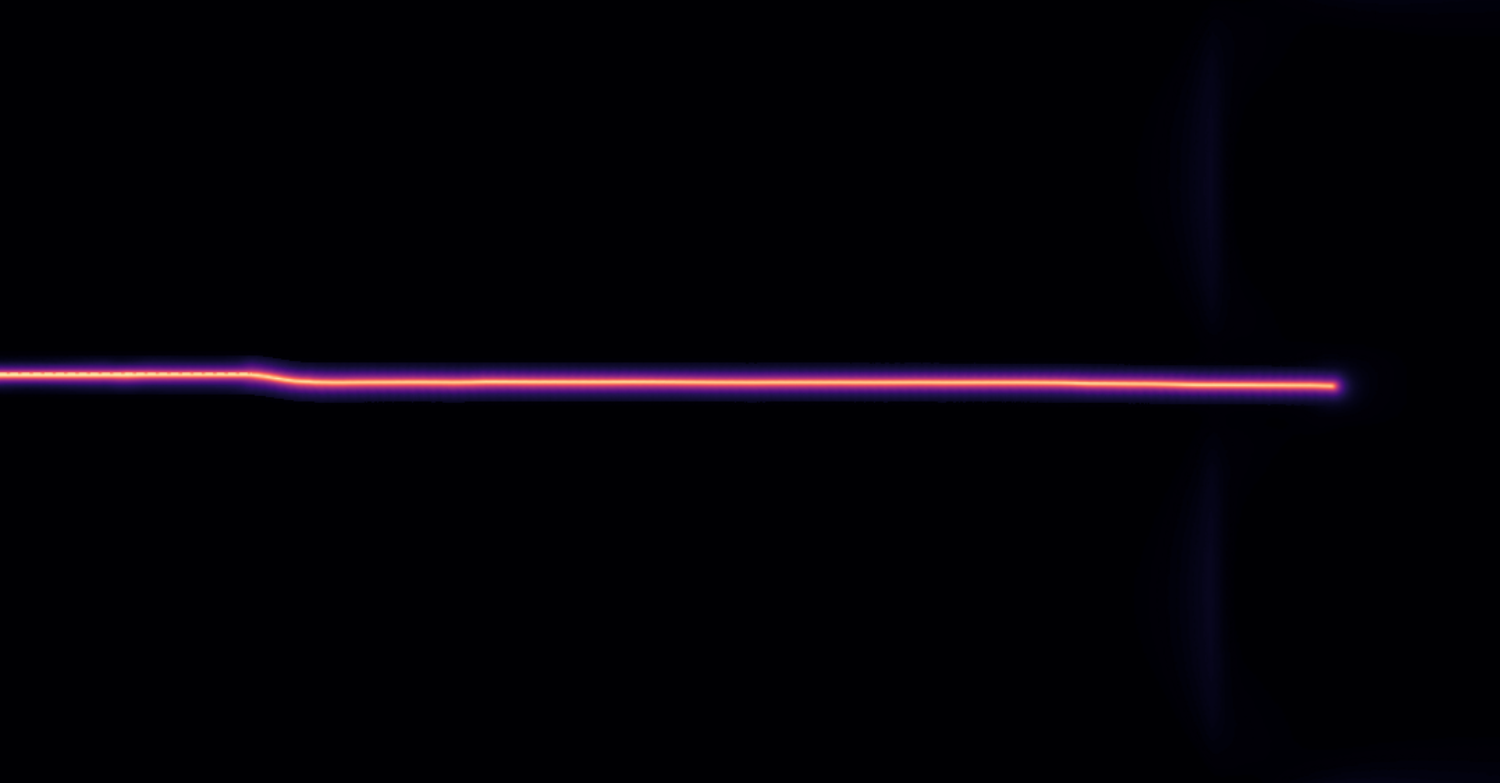}
        \caption{}
    \end{subfigure}\hfill
    \begin{subfigure}[t]{0.32\linewidth}
        \includegraphics[width=\linewidth]{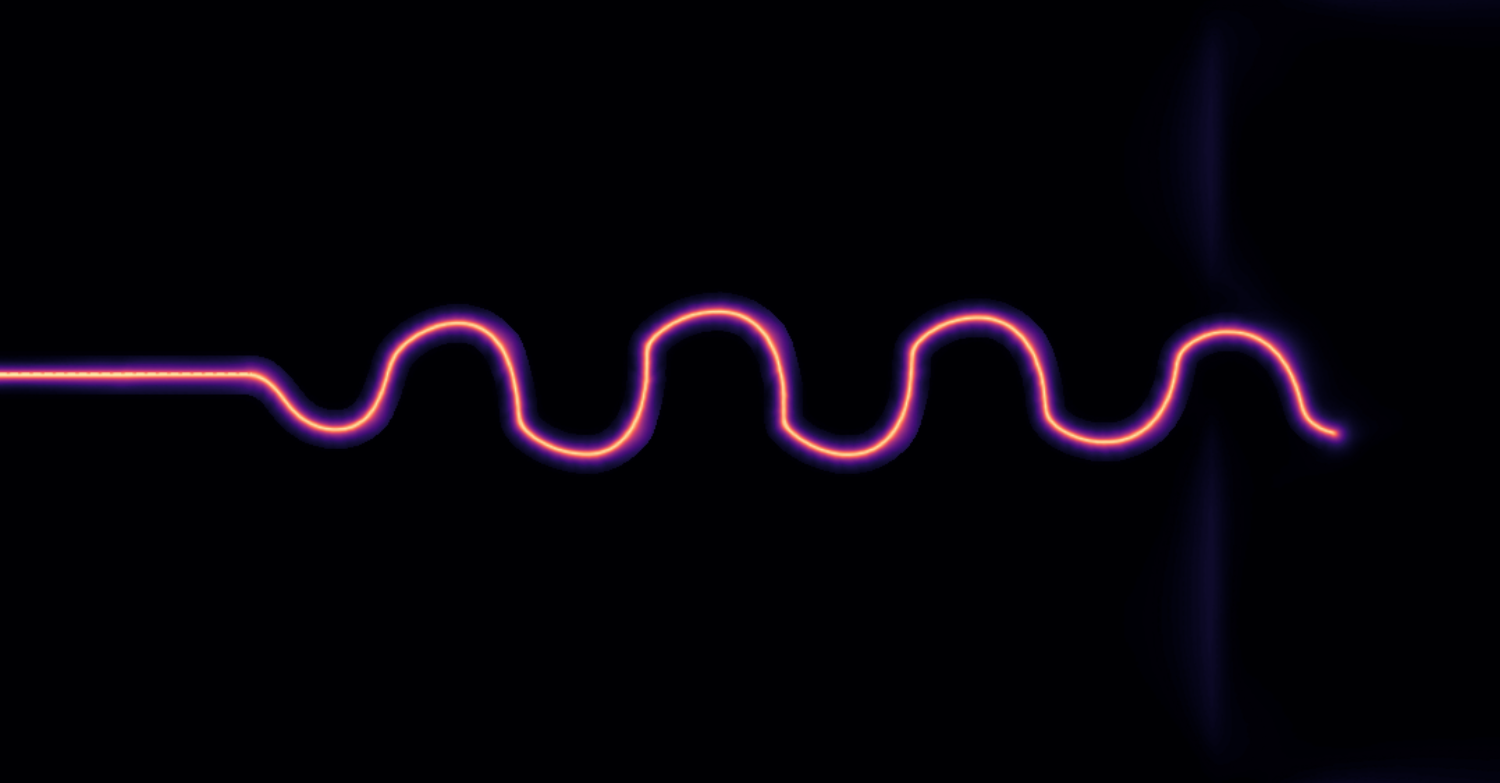}
        \caption{}
    \end{subfigure}\hfill
    \begin{subfigure}[t]{0.32\linewidth}
        \includegraphics[width=\linewidth]{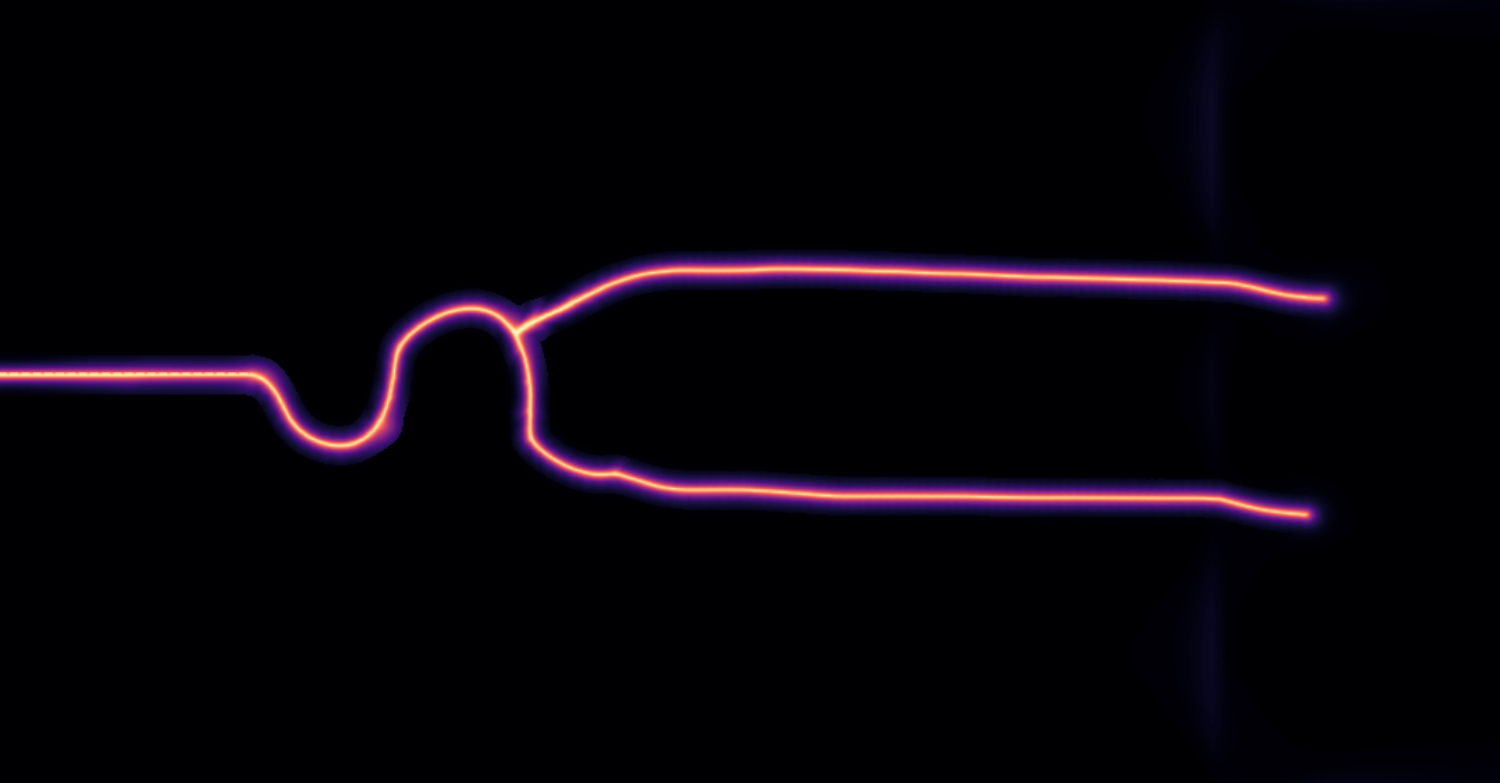}
        \caption{}
    \end{subfigure}

    \vspace{3pt}
    \begin{subfigure}[t]{0.32\linewidth}
        \includegraphics[width=\linewidth]{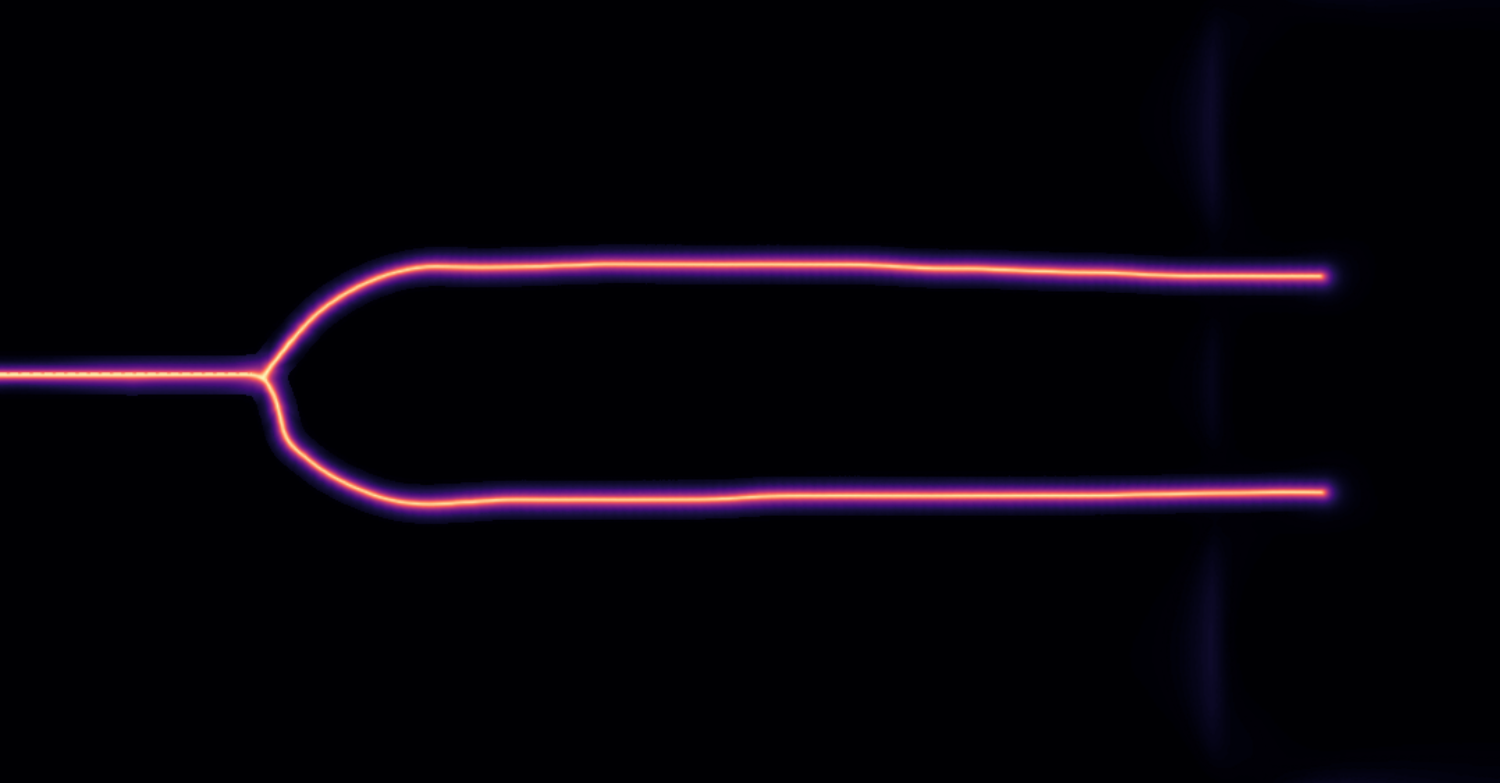}
        \caption{}
    \end{subfigure}\hfill
    \begin{subfigure}[t]{0.32\linewidth}
        \includegraphics[width=\linewidth]{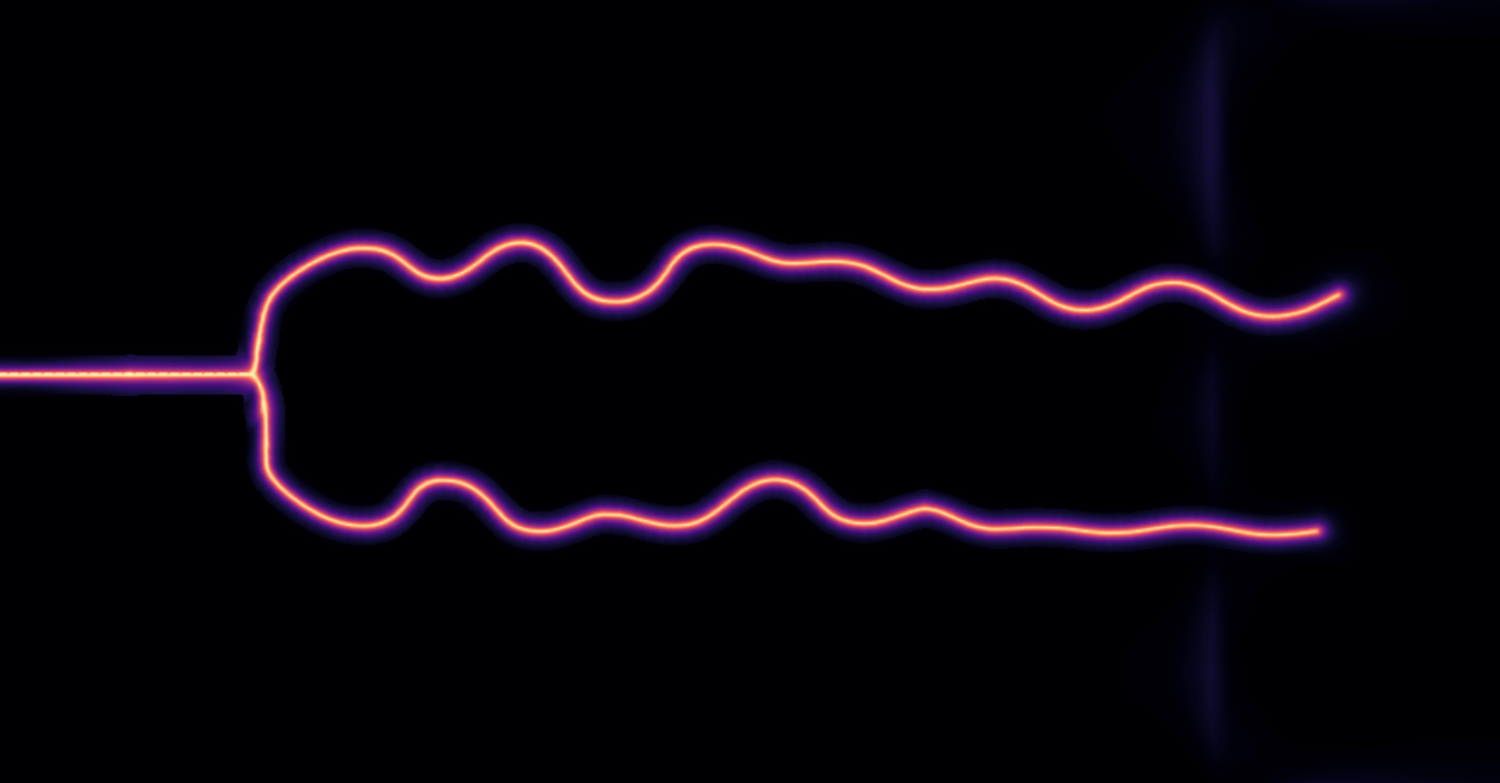}
        \caption{}
    \end{subfigure}\hfill
    \begin{subfigure}[t]{0.32\linewidth}
        \includegraphics[width=\linewidth]{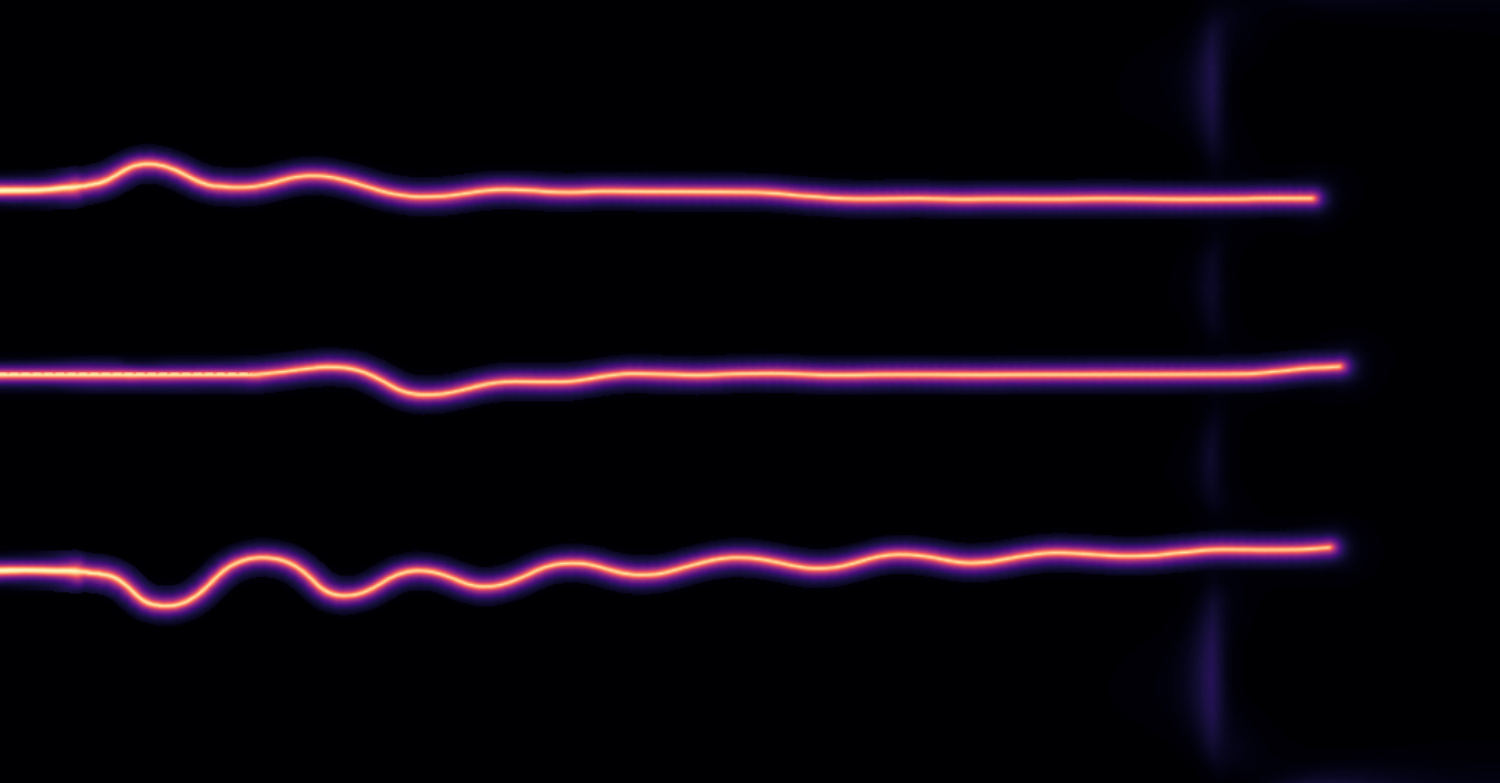}
        \caption{}
    \end{subfigure}
    \caption{Glass strip, phase fields at six temperature jumps,
    the bath on the left and the front advancing $\lreg$ per step.
    (a)~$\dT = 130$~K, straight. (b)~$165$~K, oscillatory.
    (c)~$175$~K, oscillatory over about one and a half periods, then
    branching. (d)~$190$~K, branching at the end of the notch.
    (e)~$217$~K, branching at the end of the notch, with oscillating
    branches. (f)~$307$~K, two further cracks nucleated along the
    leading edge beside the crack from the notch. The faint vertical band near the right end
    marks the immersion front at the last step. All panels share one color scale from
    intact (dark) to fully damaged (light).}
    \label{fig:strip-morph}
\end{figure}

Fig.~\ref{fig:strip-morph} shows the final paths at six of the nine
temperature jumps. The path is straight at $130$~K, oscillates from
$145$ to $165$~K with an amplitude that grows with the jump, and
branches from $170$ to $217$~K. At $170$ and $175$~K it first
oscillates and then forks, the fork moving toward the notch as the jump
grows, and from $190$~K the tip splits at the end of the notch. This is
the sequence observed in experiment, obtained here with one set of numerical parameters. Zeng and Dolbow~\cite{zeng2026} find the same order at this
speed, straight to $127$~K, oscillatory from $147$ to $167$~K and branched above. At $307$~K two further cracks nucleate along the leading edge, and at this jump the crack of Zeng and Dolbow turns back and follows an irregular wavy path. The finite element computation gives the same class of path at each jump at which it was run except $145$~K, where its path
stays straight. At $165$~K the wavelength and the amplitude of the proposed method are
$17\%$ and $59\%$ larger than those of the finite element computation
at $0.05$~mm per step, its finest increment (\ref{app:increment}). The faint band at
the immersion front is partial damage, $\pf < 0.15$, which recedes
behind the front since irreversibility acts on the crack set only.

\begin{figure}[!htbp]
    \centering
    \input{figures/data/strip_fix_fit.tex}
    \pgfplotsset{strippar/.style={paperaxis, scale only axis, width=0.74\linewidth,
                 height=5.0cm, xmin=7.8, xmax=11.6, xtick={8,9,10,11},
                 xtick pos=bottom, xlabel={$R$}},
                 stripdt/.style={paperaxis, scale only axis, width=0.74\linewidth,
                 height=5.0cm, xmin=123.0, xmax=183.0, xtick={130,150,170},
                 axis x line*=top, axis y line=none, xlabel={$\dT$ (K)},
                 ymin=0, ymax=1}}
    \tikzset{lsq/.style={black, thick}, rcline/.style={black!45, dashed}}
    \begin{subfigure}[t]{0.49\linewidth}
        \centering
        \begin{tikzpicture}
            \begin{axis}[stripdt]\end{axis}
            \begin{axis}[strippar, ylabel={$A^{2} / b^{2}$}, scaled y ticks=false,
                         ymin=-0.002, ymax=0.032, ytick={0,0.01,0.02,0.03},
                         yticklabel style={/pgf/number format/fixed, /pgf/number format/precision=2},
                         legend columns=4, legend to name={leg:strip-params},
                         legend style={/tikz/every even column/.append style={column sep=1.8ex}}]
                \draw[rcline] (axis cs:\stripfitRc,-0.002) -- (axis cs:\stripfitRc,0.032);
                \addplot[cblue, only marks, mark=*, mark size=2.2pt]
                    table[col sep=comma, x=R, y expr=\thisrow{A_b}^2] {figures/data/strip_fix_order.csv};
                \addlegendentry{oscillating path}
                \addplot[cblue, only marks, mark=triangle, mark size=2.6pt, thick]
                    table[col sep=comma, x=R, y=A_b] {figures/data/strip_fix_order_straight.csv};
                \addlegendentry{straight path}
                \addplot[lsq, domain=\stripfitRc:10.75, samples=2]
                    {\stripfitS*(x-\stripfitRc)};
                \addlegendentry{least-squares line}
                \addlegendimage{rcline}
                \addlegendentry{threshold $R_{\mathrm{c}}$}
            \end{axis}
        \end{tikzpicture}
        \caption{}
    \end{subfigure}\hfill
    \begin{subfigure}[t]{0.49\linewidth}
        \centering
        \begin{tikzpicture}
            \begin{axis}[stripdt]\end{axis}
            \begin{axis}[strippar, ylabel={$\Lambda / b$},
                         ymin=0.47, ymax=0.70, ytick={0.5,0.55,0.6,0.65,0.7},
                         yticklabel style={/pgf/number format/fixed, /pgf/number format/precision=2}]
                \draw[rcline] (axis cs:\stripfitRc,0.47) -- (axis cs:\stripfitRc,0.70);
                \addplot[lsq, domain=\stripfitRc:10.75, samples=2]
                    {\stripfitM*x + \stripfitQ};
                \addplot[cblue, only marks, mark=*, mark size=2.2pt]
                    table[col sep=comma, x=R, y=lam_b] {figures/data/strip_fix_order.csv};
            \end{axis}
        \end{tikzpicture}
        \caption{}
    \end{subfigure}

    \vspace{2pt}
    \pgfplotslegendfromname{leg:strip-params}
    \caption{Glass strip at $P = 6.81$, order parameters of the proposed
    method on the developed path against the thermal-shock number $R$,
    with the temperature jump on the upper axis. (a)~Squared amplitude.
    (b)~Wavelength. The lines are least-squares fits through the
    three oscillating cases. The line in (a) meets zero amplitude at the
    threshold $R_{\mathrm{c}}$, and the line in (b) is drawn from there.
    The cases from $170$~K on branch or form several cracks and have
    neither parameter.}
    \label{fig:strip-params}
\end{figure}
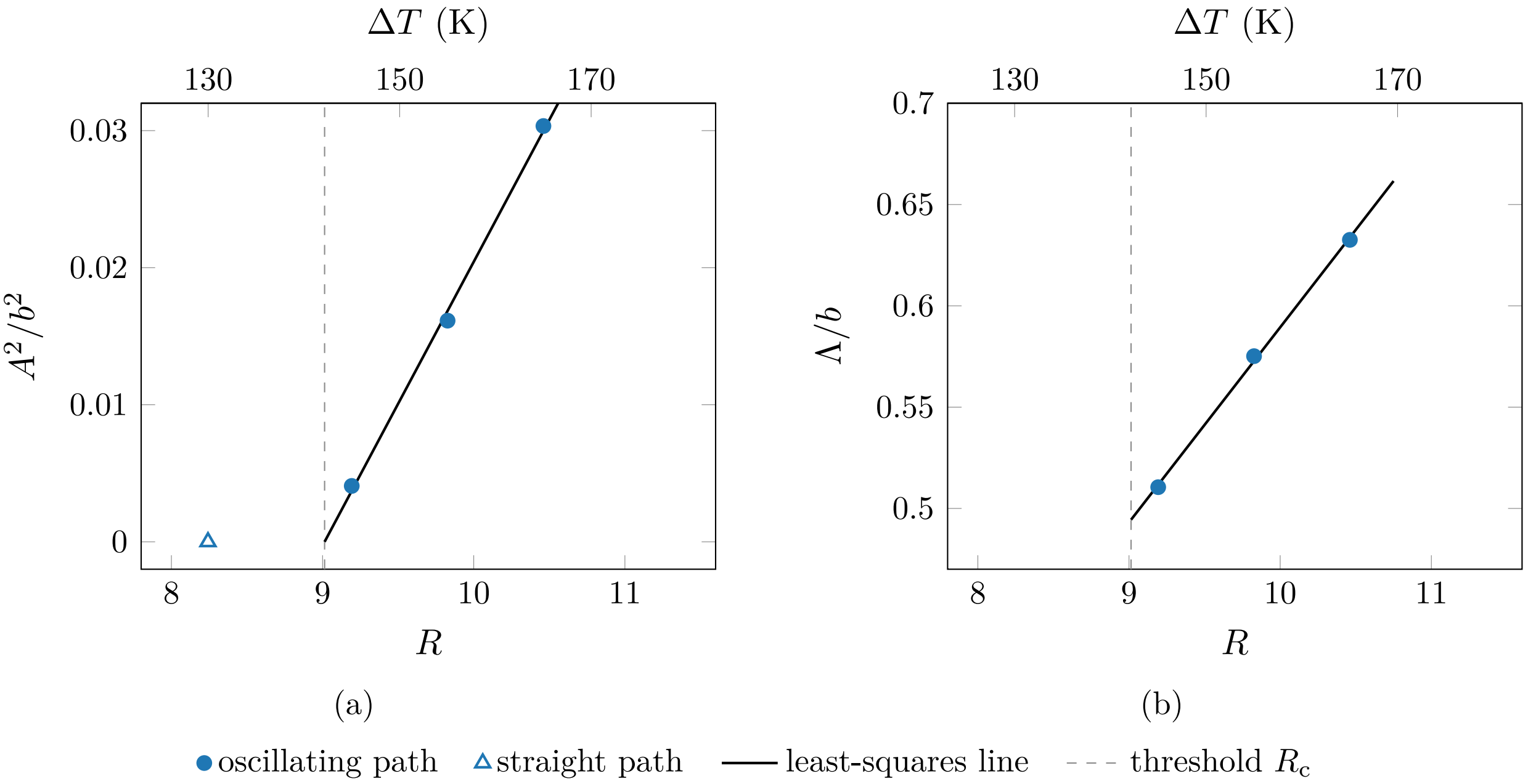

The path at $130$~K stays within $0.09$~mm of a straight line, and the
amplitude and the wavelength of the three oscillating cases rise with
$R$, Fig.~\ref{fig:strip-params}. In these cases the fourth half swing, centered between $x = 13.7$ and $15.5$~mm, is the largest, and the later ones decrease. At $145$~K the last half swing of the developed path is $0.94$ of the largest and the last of the whole path $0.28$ of it, so the oscillation there decays over the last quarter of the travel, which the developed path excludes. Yuse and
Sano~\cite{yuse1997}
reported the transition to oscillation as a supercritical Hopf
bifurcation, in which the oscillation grows continuously from zero and
its squared amplitude grows linearly with the distance from the
threshold $R_{\mathrm{c}}$, as in the phase-field computation of Corson
et al.~\cite{corson2009}. The three oscillating cases are consistent
with this linear growth, Fig.~\ref{fig:strip-params}(a), and the line
gives $R_{\mathrm{c}} = 9.01$, between the straight case at $R = 8.24$
and the lowest oscillating case at $R = 9.19$. Leaving out one case moves $R_{\mathrm{c}}$ by at most $0.1$ (\ref{app:orderparams}), and over the five
settings of \ref{app:increment} at which $145$ and $165$~K both oscillate, $R_{\mathrm{c}}$ lies between $8.97$ and $9.18$.
Three cases cannot distinguish this growth from the first-order
transition described by Bahr et al.~\cite{bahr1995}. The
wavelength rises linearly with $R$, Fig.~\ref{fig:strip-params}(b),
and its value at the threshold, $\Lambda / b = 0.494$, is compared with
experiment, which measures it at onset~\cite{ronsin1998}.

\begin{table}[!htbp]
    \centering
    \small
    \caption{Onset of oscillation on the glass strip at $P = 6.81$. Zeng
    and Dolbow give the interval of their computations at this speed,
    and their wavelength is read here from their lowest oscillating
    path, at $R = 9.32$. The values of Corson et al.\ are read from their
    Fig.~6 at $1/P = 0.147$. The finite element computation is that of
    \ref{app:increment} at $0.05$~mm per step, with its wavelength at
    $R = 10.46$. The experimental value is the fit of Ronsin and Perrin,
    $\Lambda/b = 0.30 + 2.5/P$ (\ref{app:orderparams}).}
    \label{tab:strip-onset}
    \begin{tabular}{@{}lcc@{}}
        \toprule
        & Threshold $R_{\mathrm{c}}$ & Wavelength $\Lambda / b$ \\
        \midrule
        Proposed method & $9.01$ & $0.494$ \\
        Zeng and Dolbow~\cite{zeng2026}, same strip & between $8.05$ and $9.32$ & $0.53$ (at $R = 9.32$) \\
        Linear stability analysis, Corson et al.~\cite{corson2009} & $8.43$ & $0.556$ \\
        Finite element, same model & between $9.19$ and $10.46$ & $0.542$ (at $R = 10.46$) \\
        Experiment, Ronsin and Perrin~\cite{ronsin1998} & & $0.667$ \\
        \bottomrule
    \end{tabular}
\end{table}

Table~\ref{tab:strip-onset} compares the onset with the other
computations and with experiment. The threshold lies inside the
interval bracketed by the straight and the oscillating paths of Zeng
and Dolbow~\cite{zeng2026} on the same strip, although in a run at the
slightly higher P\'eclet number $P = 6.9$ their computation already
oscillates at $R = 8.0$. The threshold of the proposed method lies
about $13\%$ above $R = 8.0$, $7\%$ above that of the linear stability analysis of Corson
et al.~\cite{corson2009} for a sharp crack and below the finite element threshold for the same model. The wavelength at onset lies
$11\%$ below that of Corson et al., and at $R = 9.32$ the line of
Fig.~\ref{fig:strip-params}(b) gives $\Lambda / b = 0.524$, about $1\%$
below the $0.53$ read from the path of Zeng and Dolbow. The onset wavelengths of the proposed method and of
Corson et al.\ lie $26\%$ and $17\%$ below the experimental value,
which is measured on the real temperature field rather than the
idealized front of Eq.~\eqref{eq:front}.

\FloatBarrier
\subsection{Quenched ceramic slab}
\label{subsec:slab}

\begin{figure}[htbp]
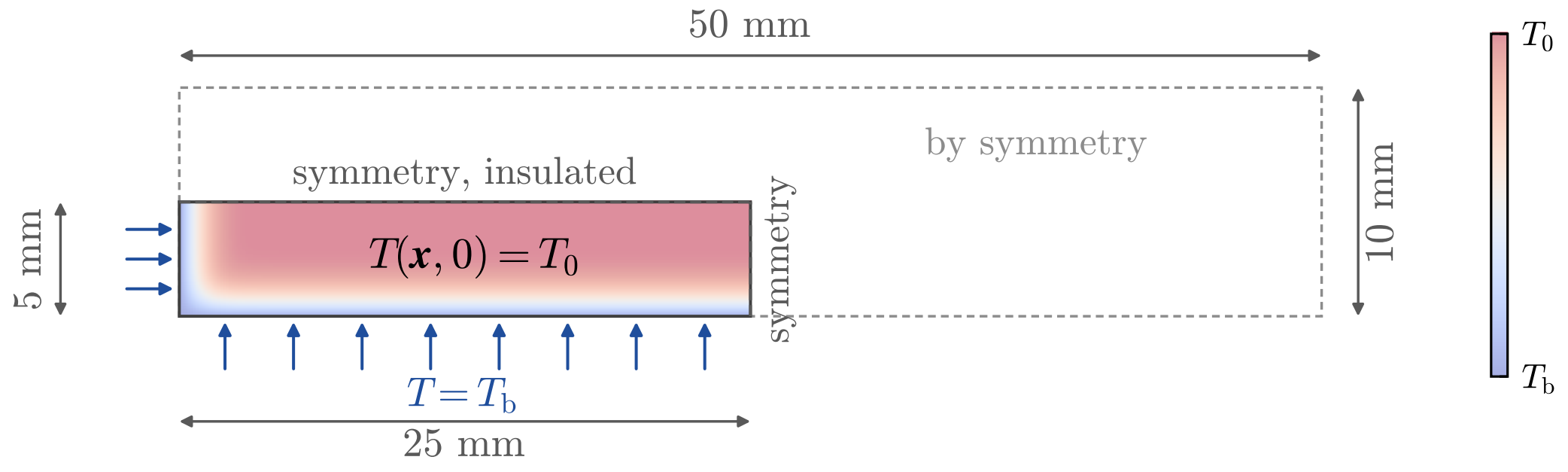

    \geomslabline
    \caption{Geometry and boundary conditions of the quenched slab.
    The computed quarter is shaded, held at the bath temperature along
    its two outer faces and bounded by the two insulated symmetry cuts. The
    temperature shading is schematic.}
    \label{fig:geom-slab}
\end{figure}

With this problem a thermal transient computed by the temperature network drives the
spontaneous nucleation and selection of a crack array. A ceramic slab
$50$~mm long and $10$~mm wide, at the uniform stress-free temperature
$\Tz$, is quenched into water at $\Tenv = 293$~K through all four faces, and the thermal load is $\dT = \Tz - \Tenv$. A periodic array of cracks nucleates
along the cooled faces, and a subset of them propagates into the depth. Geometry and material follow Tang et al.~\cite{tang2016}, who modeled the alumina of the experiments of Jiang et al.~\cite{jiang2012}, except that the toughness is that of Bourdin et al.~\cite{bourdin2014}. The benchmark is the computation of Bourdin et
al.\ on a slab of nearly the same proportions, which they compared with
the experiment of Shao et al.~\cite{shao2011}. One quarter,
$25$ by $5$~mm, is computed, Fig.~\ref{fig:geom-slab}, under plane
stress with the material of Table~\ref{tab:mat-quench},
$\lreg = 0.05$~mm and no mechanical load. The symmetry cuts carry zero
normal displacement and are insulated, and the cooled faces are
traction free.

A uniformly cooled slab in a homogeneous model develops a uniform
damage band that bifurcates to a periodic mode and localizes into
cracks~\cite{sicsic2014,bourdin2014}, and a computation leaves the
uniform state only through a perturbation of its symmetry. The quench
problems take this perturbation from the random material field of Tang
et al.~\cite{tang2016}, who regard the scatter of the local stiffness
and strength as intrinsic to the ceramic. Square cells of side
$\lreg / 2$ each receive independent Weibull draws of Young's modulus
$E_{i}$ and strength $\sigma_{\mathrm{t},i}$, of Weibull modulus
$m = 15$ with the nominal values of Table~\ref{tab:mat-quench} as scale
parameters. This Weibull modulus is the value that Tang et al.\ adopt,
not one measured on specimens.
Since the AT2 peak stress $\sigma_{\mathrm{c}}$ of
Section~\ref{subsec:strip} is fixed by $\Gc$, $E$ and $\lreg$, the drawn
strength enters through the toughness,
\begin{equation}
    G_{\mathrm{c},i}
    =
    \frac{256}{27}\, \frac{\lreg\, \sigma_{\mathrm{t},i}^{2}}{E_{i}} ,
    \label{eq:weibull-gc}
\end{equation}
which is rescaled by one constant per realization so that its mean equals the
nominal $\Gc$ and the Griffith length
$\lgrif = \Gc / (E \alphaT^{2} \dT^{2})$ is the same for every
realization (\ref{app:hyper}). The finite element computation evaluates
the same realization at its quadrature points.

The cooled surfaces are held at the bath temperature from $t = 0$ on,
the limit $\hconv \to \infty$ of the convective condition, so the
temperature is prescribed on the boundary and computed in the interior.
The conductivity is damage-independent since the cracks run along the
heat flux. Both choices follow Bourdin et al.~\cite{bourdin2014}, Chu et al.~\cite{chu2017}, Mandal et al.~\cite{mandal2021fracture} and Tangella et al.~\cite{tangella2022}.  The
temperature is advanced on a time grid that concentrates the load
steps in the early transient,
\begin{equation}
    t_{n}
    =
    t_{\mathrm{s}} + (t_{\mathrm{f}} - t_{\mathrm{s}})
    \Bigl( \frac{n-1}{n_{\max}-1} \Bigr)^{3} ,
    \qquad n = 1, \dots, n_{\max} ,
    \label{eq:clock}
\end{equation}
with $n_{\max} = 100$ load steps from $t_{\mathrm{s}} = 10^{-4}$~s to
$t_{\mathrm{f}} = 0.2$~s after the initial state at $t_{0} = 0$, so the
first step forms a cooled layer of thickness
$\sqrt{D t_{\mathrm{s}}} = 0.03$~mm. Four thermal loads, $\dT = 280$,
$380$, $480$ and $580$~K, are computed, each on one realization of the
field and each beside a finite element computation on that realization
(\ref{app:reference}). Crack counts and the spacing $d(a)$ are read along lines parallel to the cooled long face at depths $a$ from $0.1$ to $2$~mm (\ref{app:hyper}).

\begin{table}[htb]
    \centering
    \caption{Material data of the quenched ceramic, after Tang et
    al.~\cite{tang2016} with the toughness of Bourdin et
    al.~\cite{bourdin2014}, used by the slab and the annulus under plane
    stress. The regularization length is chosen here. The film
    coefficient, chosen within the measured range, enters only the
    computations under the convective condition, in
    \ref{app:arrays}, \ref{app:sampler} and \ref{app:convective}.}
    \label{tab:mat-quench}
    \footnotesize
    \begin{tabular}{lcc}
        \toprule
        Property & Symbol & Value \\
        \midrule
        Young's modulus (GPa) & $E$ & 370 \\
        Poisson's ratio & $\nu$ & 0.30 \\
        Toughness (J\,m$^{-2}$) & $\Gc$ & 42.47 \\
        Expansion coefficient (K$^{-1}$) & $\alphaT$ & $7.5 \times 10^{-6}$ \\
        Conductivity (W\,m$^{-1}$\,K$^{-1}$) & $\kcond$ & 31 \\
        Density (kg\,m$^{-3}$) & $\rho$ & 3980 \\
        Specific heat (J\,kg$^{-1}$\,K$^{-1}$) & $c$ & 880 \\
        Film coefficient (kW\,m$^{-2}$\,K$^{-1}$) & $\hconv$ & 50 \\
        Nominal strength (MPa) & $\bar{\sigma}_{\mathrm{t}}$ & 180 \\
        Weibull modulus & $m$ & 15 \\
        Regularization length (mm) & $\lreg$ & 0.05 \\
        \bottomrule
    \end{tabular}
\end{table}

Since the temperature does not depend on the cracks, halfway along the long face it obeys the one-dimensional heat equation normal to that face. At $\dT = 380$~K
the temperature network agrees with a backward Euler solution of this
transient on the same time steps to within $1.6\%$ of $\dT$ at every
load step, and at $480$~K to within $1.4\%$ (\ref{app:hyper}).

\begin{figure}[htb]
    \centering
    \setlength{\tabcolsep}{0pt}
    \renewcommand{\arraystretch}{0.6}
    \newcommand{\slablab}[1]{\raisebox{0.036\textwidth}{\scriptsize #1}}
    \newcommand{\slabpanel}[1]{\includegraphics[width=0.425\textwidth]{#1}}
    \newcommand{\slabfe}[1]{\slabpanel{slabdir_plate_fe_T#1.png}}
    \begin{tabular}{@{}r@{\hspace{1.5mm}}c@{\hspace{3mm}}c@{}}
        & {\footnotesize $\dT = 280$~K} & {\footnotesize $\dT = 380$~K} \\[1pt]
        \slablab{proposed} & \slabpanel{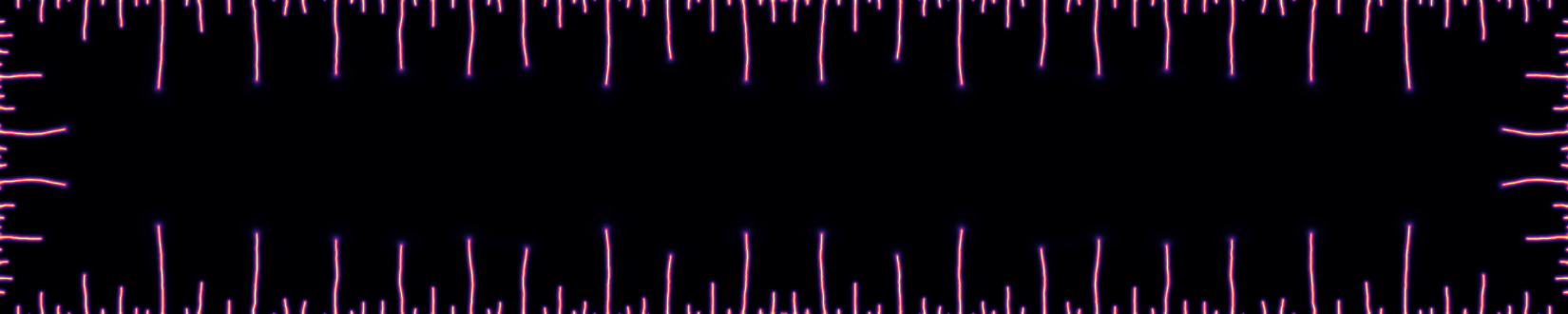} & \slabpanel{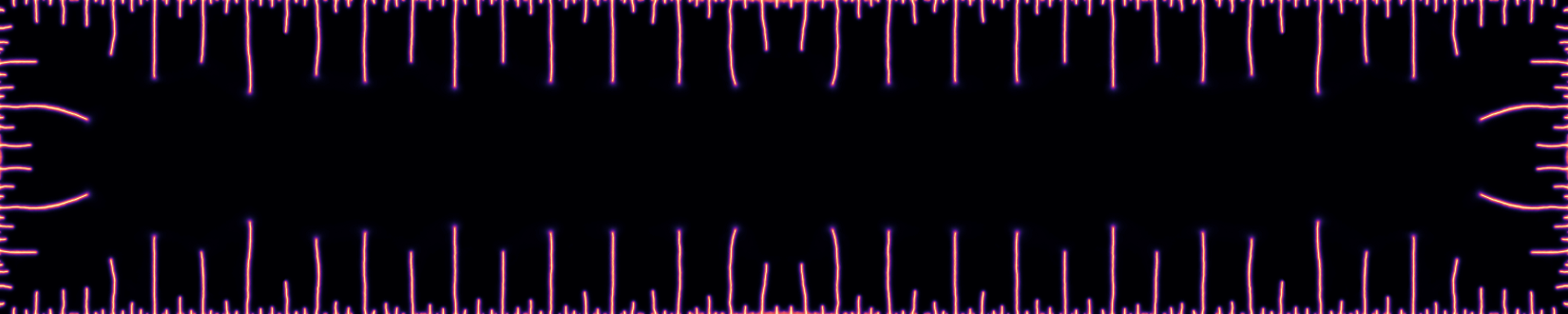} \\[2pt]
        \slablab{finite element} & \slabfe{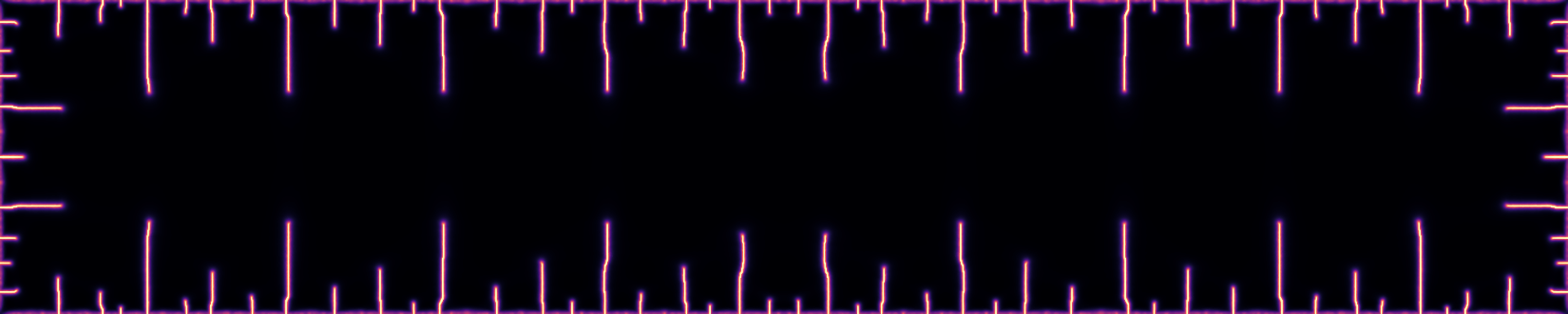} & \slabfe{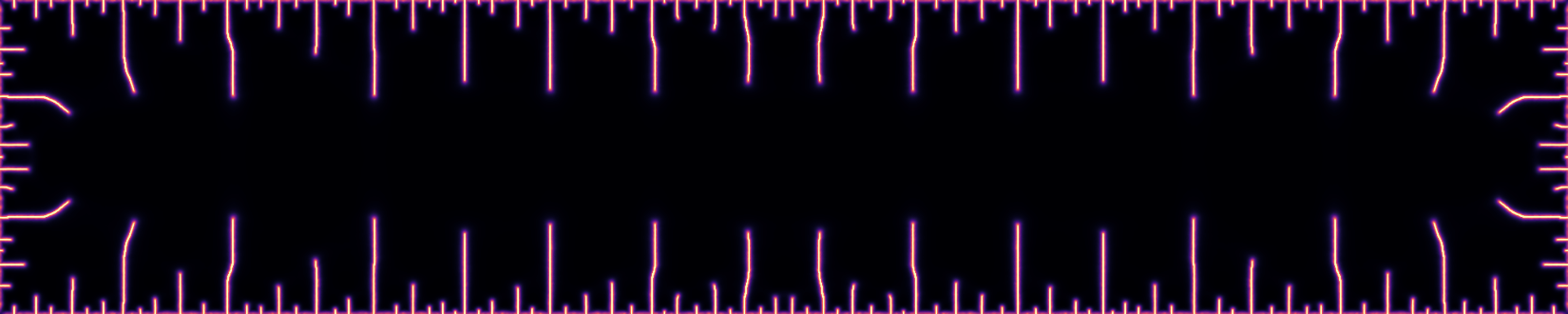} \\[2pt]
        \slablab{experiment} & \slabpanel{experiment/jiang2012_T280.png} & \slabpanel{experiment/jiang2012_T380.png} \\[6pt]
        & {\footnotesize $\dT = 480$~K} & {\footnotesize $\dT = 580$~K} \\[1pt]
        \slablab{proposed} & \slabpanel{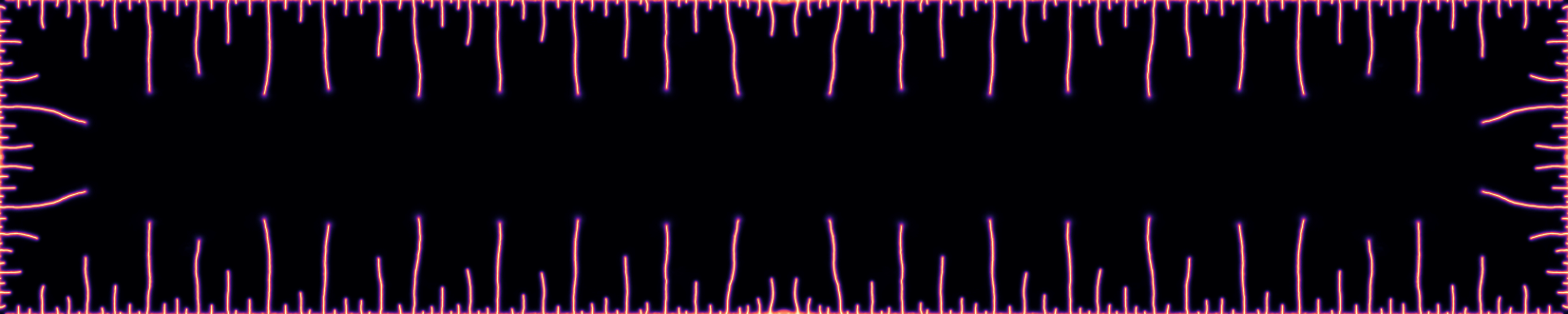} & \slabpanel{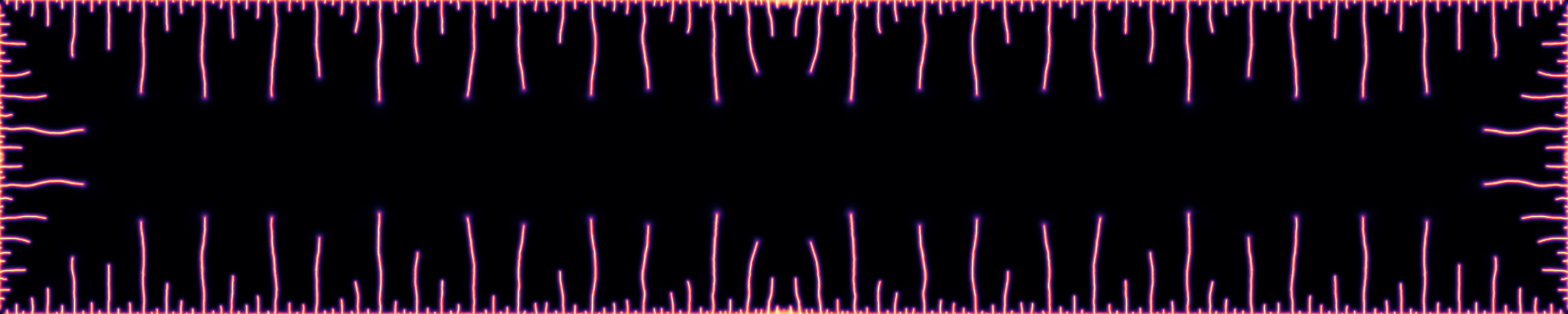} \\[2pt]
        \slablab{finite element} & \slabfe{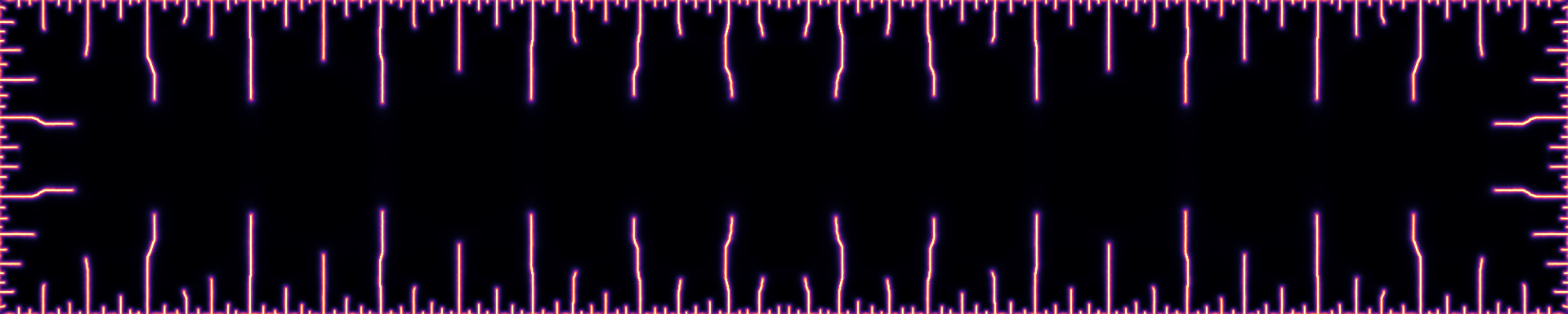} & \slabfe{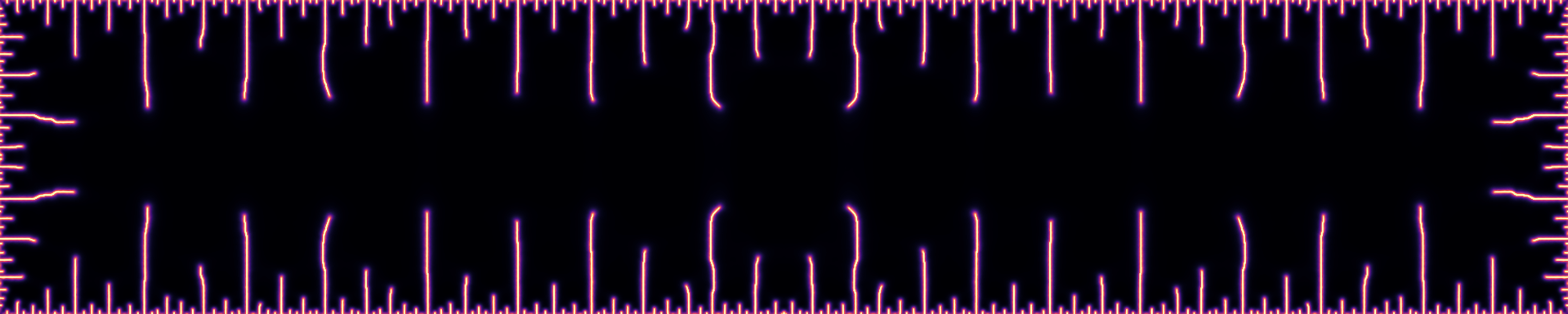} \\[2pt]
        \slablab{experiment} & \slabpanel{experiment/jiang2012_T480.png} & \slabpanel{experiment/jiang2012_T580.png} \\
    \end{tabular}
    \caption{Quenched slab under a prescribed boundary temperature, final
    crack patterns of the $50$ by $10$~mm slab at four thermal loads.
    In each group, the proposed method (top) and the finite element
    computation on the same realization (middle) show the computed quarter reflected about its two
    symmetry cuts, with the phase field from $0$ (black) to $1$ (light).
    The bottom row is the photograph of a plate quenched from
    $\Tz = \Tenv + \dT$ into a bath at $293$~K by Jiang et
    al.~\cite{jiang2012}, the cracks revealed by dye. Photographs
    reprinted from Acta Materialia, Vol.~60, C.P. Jiang et al., A study
    of the mechanism of formation and numerical simulations of crack
    patterns in ceramics subjected to thermal shock, Pages 4540--4550,
    Copyright (2012), with permission from Elsevier.}
    \label{fig:slab-patterns}
\end{figure}

Under the prescribed boundary temperature a uniform damage band forms in the
cooled layer in the first load steps and breaks into a dense array of
short cracks, whose number along the face stays fixed from load step
$20$ on, so the later steps select which cracks continue into the
depth. Fig.~\ref{fig:slab-patterns} shows the final states at the four
loads beside the finite element computations and the photographs of
plates of this ceramic quenched by Jiang et al.~\cite{jiang2012}. All
three show parallel cracks normal to the cooled long faces with a
hierarchy of lengths, a spacing of the long cracks that grows with the
depth and decreases with the load, and shorter cracks from the end
faces. The two computations differ in the positions of the cracks, and
both show short cracks between the long ones that are difficult to
observe in the photographs, as Tang et al.~\cite{tang2016} note for their
computation.

\begin{figure}[htb]
    \centering
    \begin{tabular}{@{}cc@{}}
    \begin{tikzpicture}
        \begin{axis}[paperaxis, xmode=log, ymode=log, width=0.46\linewidth, height=0.31\linewidth, xmin=1, xmax=600, ymin=15, ymax=1000, xtick={1,2,5,10,20,50,100,200,500}, ytick={20,50,100,200,500}, log ticks with fixed point, xlabel={$a / \lgrif$}, ylabel={$d / \lgrif$}, title={\footnotesize $\dT = 280$~K},
                     legend to name=spacinglegend, legend columns=4, legend style={/tikz/every even column/.append style={column sep=1.5ex}}]
            \addplot[black, thick] table[x=a_l0, y=d_l0, col sep=comma] {figures/data/slab_bourdin_numerics.csv};
            \addlegendentry{Bourdin et al.\ numerics}
            \addplot[only marks, mark=o, mark size=1.7pt, cgreen, thick] table[x=a_l0, y=d_l0, col sep=comma] {figures/data/slab_shao_experiment.csv};
            \addlegendentry{Shao et al.\ experiment}
            \addplot[cblue, line width=1.2pt] table[x=a_l0, y=d_l0, col sep=comma] {figures/data/slabdir_dem_T280.csv};
            \addlegendentry{proposed method}
            \addplot[corange, thick, dashed] table[x=a_l0, y=d_l0, col sep=comma] {figures/data/slabdir_fe_T280.csv};
            \addlegendentry{finite element}
        \end{axis}
    \end{tikzpicture} &
    \begin{tikzpicture}
        \begin{axis}[paperaxis, xmode=log, ymode=log, width=0.46\linewidth, height=0.31\linewidth, xmin=1, xmax=600, ymin=15, ymax=1000, xtick={1,2,5,10,20,50,100,200,500}, ytick={20,50,100,200,500}, log ticks with fixed point, xlabel={$a / \lgrif$}, ylabel={$d / \lgrif$}, title={\footnotesize $\dT = 380$~K}]
            \addplot[black, thick] table[x=a_l0, y=d_l0, col sep=comma] {figures/data/slab_bourdin_numerics.csv};
            \addplot[only marks, mark=o, mark size=1.7pt, cgreen, thick] table[x=a_l0, y=d_l0, col sep=comma] {figures/data/slab_shao_experiment.csv};
            \addplot[cblue, line width=1.2pt] table[x=a_l0, y=d_l0, col sep=comma] {figures/data/slabdir_dem_T380.csv};
            \addplot[corange, thick, dashed] table[x=a_l0, y=d_l0, col sep=comma] {figures/data/slabdir_fe_T380.csv};
        \end{axis}
    \end{tikzpicture} \\
    \begin{tikzpicture}
        \begin{axis}[paperaxis, xmode=log, ymode=log, width=0.46\linewidth, height=0.31\linewidth, xmin=1, xmax=600, ymin=15, ymax=1000, xtick={1,2,5,10,20,50,100,200,500}, ytick={20,50,100,200,500}, log ticks with fixed point, xlabel={$a / \lgrif$}, ylabel={$d / \lgrif$}, title={\footnotesize $\dT = 480$~K}]
            \addplot[black, thick] table[x=a_l0, y=d_l0, col sep=comma] {figures/data/slab_bourdin_numerics.csv};
            \addplot[only marks, mark=o, mark size=1.7pt, cgreen, thick] table[x=a_l0, y=d_l0, col sep=comma] {figures/data/slab_shao_experiment.csv};
            \addplot[cblue, line width=1.2pt] table[x=a_l0, y=d_l0, col sep=comma] {figures/data/slabdir_dem_T480.csv};
            \addplot[corange, thick, dashed] table[x=a_l0, y=d_l0, col sep=comma] {figures/data/slabdir_fe_T480.csv};
        \end{axis}
    \end{tikzpicture} &
    \begin{tikzpicture}
        \begin{axis}[paperaxis, xmode=log, ymode=log, width=0.46\linewidth, height=0.31\linewidth, xmin=1, xmax=600, ymin=15, ymax=1000, xtick={1,2,5,10,20,50,100,200,500}, ytick={20,50,100,200,500}, log ticks with fixed point, xlabel={$a / \lgrif$}, ylabel={$d / \lgrif$}, title={\footnotesize $\dT = 580$~K}]
            \addplot[black, thick] table[x=a_l0, y=d_l0, col sep=comma] {figures/data/slab_bourdin_numerics.csv};
            \addplot[only marks, mark=o, mark size=1.7pt, cgreen, thick] table[x=a_l0, y=d_l0, col sep=comma] {figures/data/slab_shao_experiment.csv};
            \addplot[cblue, line width=1.2pt] table[x=a_l0, y=d_l0, col sep=comma] {figures/data/slabdir_dem_T580.csv};
            \addplot[corange, thick, dashed] table[x=a_l0, y=d_l0, col sep=comma] {figures/data/slabdir_fe_T580.csv};
        \end{axis}
    \end{tikzpicture}
    \end{tabular}

    \ref*{spacinglegend}
    \caption{Quenched slab under a prescribed boundary temperature, crack
    spacing $d$ against the depth $a$ below the cooled long face at the
    four thermal loads, both normalized by the Griffith length $\lgrif$
    (\ref{app:hyper}). The proposed method (blue, solid) and
    the finite element computation (orange, dashed)
    are shown beside the benchmark of Bourdin et al.~\cite{bourdin2014} (black) and
    the experiment of Shao et al.~\cite{shao2011} (green circles), both
    digitized from the figure of Bourdin et al.}
    \label{fig:slab-spacing}
\end{figure}
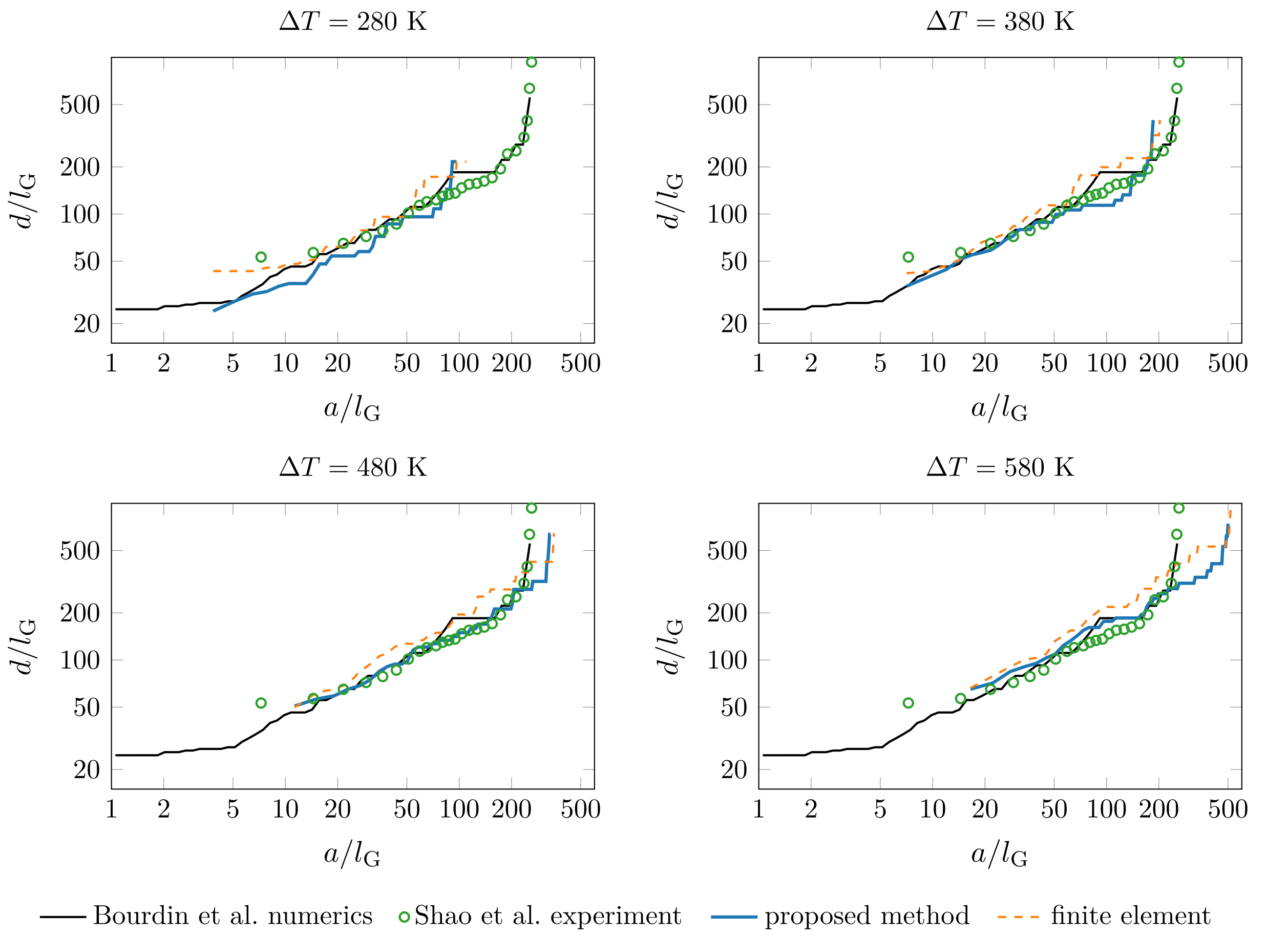

Bourdin et al.~\cite{bourdin2014} report that the developed cracks
propagate under the control of the Griffith length alone, so the
spacing against the depth in units of $\lgrif$ should follow one curve
at every load. Fig.~\ref{fig:slab-spacing} compares it with their curve
and with the experiment of Shao et al.~\cite{shao2011}. Averaged
geometrically over the depths covered by both curves, the spacing of
the proposed method lies within $14\%$ of the benchmark at every load,
smaller at $280$~K and larger at $580$~K (\ref{app:hyper}). That of the
finite element computation lies $11\%$ to $23\%$ above the benchmark,
and that of the proposed method $17\%$ to $25\%$ below that of the
finite element computation. Since the
benchmark uses a gradient damage model with dissipation linear in the
damage on a homogeneous material, the agreement of the proposed method
with the benchmark does not show which solver is more accurate. The
benchmark tests the scaling with the Griffith length, and the finite
element computation tests the solution of the same AT2 model up to the treatment of irreversibility. At
$380$~K three solver seeds on the same field, which differ in the
sampler and the network initialization, give spacings from $5\%$ to $11\%$ below the benchmark. With every iteration budget doubled the
spacing lies $1.8\%$ below it, outside the range of the seeds and
toward the finite element computation (\ref{app:arrays}), so part of
the difference from the finite element computation at this load depends
on the budget.

\begin{table}[htb]
    \centering
    \caption{Quenched slab under a prescribed boundary temperature,
    number of cracks along the cooled long face that reach the depth
    $a$, the proposed method against the finite element computation on
    the same realization of the field. A
    crack is a local maximum of the
    phase field above $0.7$, counted on the computed quarter over the
    $22.5$~mm of the face more than $2.5$~mm from the end face
    (\ref{app:hyper}).}
    \label{tab:slab-counts}
    \footnotesize
    \begin{tabular}{@{}lccccccccc@{}}
        \toprule
        $a$ (mm) & $0.1$ & $0.2$ & $0.3$ & $0.4$ & $0.5$ & $0.75$ & $1.0$ & $1.5$ & $2.0$ \\
        \midrule
        $280$~K, proposed & 36 & 27 & 24 & 19 & 16 & 15 & 10 & 9 & 8 \\
        $280$~K, finite element & 20 & 19 & 18 & 16 & 14 & 11 & 9 & 6 & 5 \\
        \addlinespace
        $380$~K, proposed & 46 & 31 & 27 & 22 & 20 & 16 & 15 & 14 & 9 \\
        $380$~K, finite element & 38 & 31 & 23 & 19 & 17 & 14 & 9 & 8 & 7 \\
        \addlinespace
        $480$~K, proposed & 50 & 39 & 30 & 27 & 21 & 19 & 17 & 12 & 9 \\
        $480$~K, finite element & 51 & 36 & 24 & 20 & 19 & 16 & 13 & 9 & 7 \\
        \addlinespace
        $580$~K, proposed & 57 & 41 & 34 & 27 & 23 & 20 & 19 & 13 & 11 \\
        $580$~K, finite element & 56 & 37 & 30 & 24 & 19 & 17 & 13 & 9 & 8 \\
        \bottomrule
    \end{tabular}
\end{table}

Table~\ref{tab:slab-counts} lists the counts against the depth. The counts of the two computations differ by at most $35\%$ down to $0.5$~mm at the three higher loads,
while at $280$~K the proposed method has nearly twice as many of the
shallowest cracks, and from $1$~mm on it has $1$ to $6$ more cracks at
every load. With the bound of the finite element computation at
$380$~K restricted to the crack set $\{\pf_{n-1} \geq 1/2\}$, the rule
of the proposed method, the finite element counts change by at most $3$ and the
difference remains, $2$ to $7$ cracks from $1$~mm on.

The two experiments differ at depth. On the photographs of Jiang et
al., Ruan et al.~\cite{ruan2023thermomechanical} counted $32$ cracks
longer than $1$~mm on the central $30$~mm of a long face at $380$~K.
The spacing measured by Shao et al.\ corresponds to $17$ such cracks,
and under the same rule the proposed method counts $20$, between the
two experiments, and the finite element computation $12$. The agreement
with experiment is qualitative.

Under the convective condition with a film coefficient of the measured magnitude~\cite{becher1981,kim1991,zhou2012}, both computations nucleate later and form sparser arrays near the surface (\ref{app:convective}). Under the convective condition the proposed method forms $16$ to $27$ cracks near the surface at the four loads against $3$ to $9$ for the finite element computation, a difference that arises at nucleation. This condition is closer to the experiments, since the Biot number $\hconv \sqrt{D t} / \kcond$ is only about $0.2$ at load step $20$, $t \approx 1.5$~ms, when the array under the prescribed boundary temperature has formed. The comparison with experiment above therefore rests on the idealization of a prescribed boundary temperature.

\FloatBarrier
\subsection{Quenched ceramic annulus}
\label{subsec:annulus}

\begin{figure}[!htbp]
    \centering
    \includegraphics[width=0.5\textwidth]{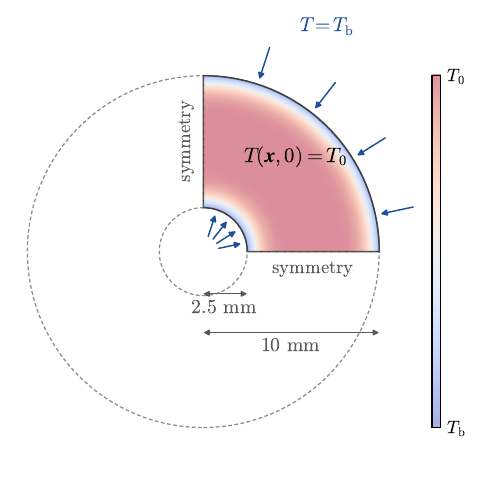}
    \caption{Geometry and boundary conditions of the quenched annulus.
    The computed quarter is shaded, held at the bath temperature along
    both rims and bounded by the two insulated radial symmetry cuts. The
    temperature shading is schematic.}
    \label{fig:geom-annulus}
\end{figure}

With this problem we apply the quench to a curved domain. Quenched
circular ceramic specimens develop periodic and hierarchical crack
patterns in experiment~\cite{liu2015circular}. An annulus with a bore
of radius $2.5$~mm and an outer radius of $10$~mm, of the ceramic of
Table~\ref{tab:mat-quench}, uniform at $\Tz$, is quenched through both
rims, Fig.~\ref{fig:geom-annulus}. One quarter is computed on a single
quadratic NURBS patch that represents both rims exactly, with no mesh.
The radial cuts carry zero normal displacement and are insulated, and
both rims are traction free and held at the bath temperature. The
material field, the damage-independent conductivity, the time grid, the loads
and the finite element computation on the same realization follow the
slab, and counts are
read along arcs at the depth $a$ from each rim (\ref{app:hyper}).

\begin{table}[!htbp]
    \centering
    \caption{Quenched annulus, number of cracks cut by the arc at the depth $a$ from each rim, counted at
    the level $0.7$ as on the slab, the proposed
    method against the finite element computation on the same
    realization of the field. At $580$~K the arc at $0.1$~mm from the bore lies inside the fully damaged layer of the proposed
    method.}
    \label{tab:ring-counts}
    \footnotesize
    \begin{tabular}{@{}lcccccccccc@{}}
        \toprule
        & \multicolumn{5}{c}{bore} & \multicolumn{5}{c}{outer rim} \\
        \cmidrule(lr){2-6} \cmidrule(l){7-11}
        $a$ (mm) & $0.1$ & $0.5$ & $1.0$ & $2.0$ & $3.0$ & $0.1$ & $0.5$ & $1.0$ & $2.0$ & $3.0$ \\
        \midrule
        $280$~K, proposed       & 7  & 5 & 3 & 1 & 0 & 22 & 11 & 7  & 2 & 0 \\
        $280$~K, finite element & 3  & 3 & 3 & 0 & 0 & 14 & 9  & 5  & 2 & 0 \\
        \addlinespace
        $380$~K, proposed       & 16 & 4 & 4 & 1 & 0 & 28 & 14 & 8  & 4 & 0 \\
        $380$~K, finite element & 8  & 4 & 3 & 2 & 0 & 24 & 9  & 7  & 3 & 0 \\
        \addlinespace
        $480$~K, proposed       & 19 & 5 & 4 & 2 & 0 & 37 & 14 & 10 & 5 & 0 \\
        $480$~K, finite element & 10 & 5 & 4 & 1 & 0 & 35 & 11 & 8  & 4 & 1 \\
        \addlinespace
        $580$~K, proposed       & 17 & 6 & 5 & 3 & 0 & 46 & 16 & 10 & 6 & 0 \\
        $580$~K, finite element & 12 & 5 & 4 & 2 & 0 & 39 & 14 & 8  & 4 & 2 \\
        \bottomrule
    \end{tabular}
\end{table}

\begin{figure}[!htbp]
    \centering
    \begin{subfigure}[t]{0.48\linewidth}
        \centering
        \includegraphics[width=\linewidth]{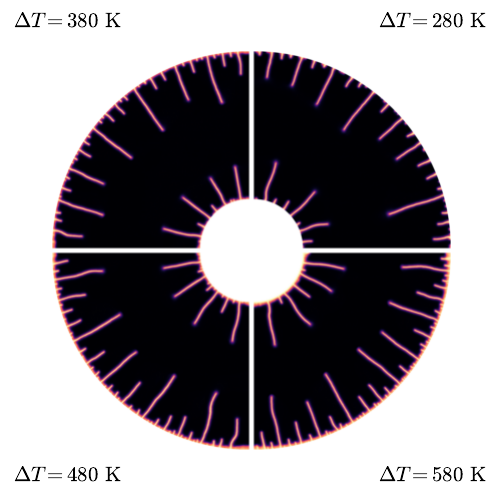}
        \caption{}
    \end{subfigure}\hfill
    \begin{subfigure}[t]{0.48\linewidth}
        \centering
        \includegraphics[width=\linewidth]{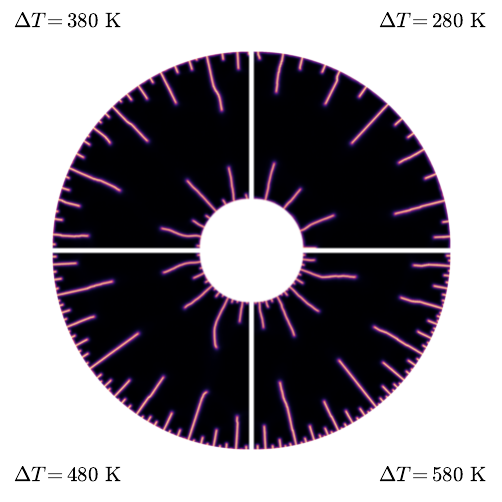}
        \caption{}
    \end{subfigure}
    \caption{Quenched annulus, final phase fields at the four thermal loads assembled into one ring, one
    load per quadrant counterclockwise from the top right. (a)~The
    proposed method. (b)~The finite element computation on the same
    realization. The phase field
    ranges from $0$ (black) to $1$ (light).}
    \label{fig:ring}
\end{figure}

\begin{figure}[!htbp]
    \centering
    \setlength{\tabcolsep}{0pt}
    \begin{tabular}{cccc}
    \begin{tikzpicture}
        \begin{axis}[paperaxis, xmode=log, ymode=log, width=0.30\linewidth, height=0.30\linewidth, xmin=3, xmax=600, ymin=15, ymax=2000, xtick={10,100}, ytick={100,1000}, log ticks with fixed point, /pgf/number format/1000 sep={}, title={\footnotesize $\dT = 280$~K}, ylabel={bore, $d / \lgrif$}, xticklabels={}, legend to name=ringlegend, legend columns=2, legend style={/tikz/every even column/.append style={column sep=1.5ex}}]
            \addplot[cblue, line width=1.2pt] table[x=a_l0, y=d_l0, col sep=comma] {figures/data/ringdir_dem_bore_T280.csv};
            \addlegendentry{proposed method}
            \addplot[corange, thick, dashed] table[x=a_l0, y=d_l0, col sep=comma] {figures/data/ringdir_fe_bore_T280.csv};
            \addlegendentry{finite element}
        \end{axis}
    \end{tikzpicture} &
    \begin{tikzpicture}
        \begin{axis}[paperaxis, xmode=log, ymode=log, width=0.30\linewidth, height=0.30\linewidth, xmin=3, xmax=600, ymin=15, ymax=2000, xtick={10,100}, ytick={100,1000}, log ticks with fixed point, /pgf/number format/1000 sep={}, title={\footnotesize $\dT = 380$~K}, yticklabels={}, xticklabels={}]
            \addplot[cblue, line width=1.2pt] table[x=a_l0, y=d_l0, col sep=comma] {figures/data/ringdir_dem_bore_T380.csv};
            \addplot[corange, thick, dashed] table[x=a_l0, y=d_l0, col sep=comma] {figures/data/ringdir_fe_bore_T380.csv};
        \end{axis}
    \end{tikzpicture} &
    \begin{tikzpicture}
        \begin{axis}[paperaxis, xmode=log, ymode=log, width=0.30\linewidth, height=0.30\linewidth, xmin=3, xmax=600, ymin=15, ymax=2000, xtick={10,100}, ytick={100,1000}, log ticks with fixed point, /pgf/number format/1000 sep={}, title={\footnotesize $\dT = 480$~K}, yticklabels={}, xticklabels={}]
            \addplot[cblue, line width=1.2pt] table[x=a_l0, y=d_l0, col sep=comma] {figures/data/ringdir_dem_bore_T480.csv};
            \addplot[corange, thick, dashed] table[x=a_l0, y=d_l0, col sep=comma] {figures/data/ringdir_fe_bore_T480.csv};
        \end{axis}
    \end{tikzpicture} &
    \begin{tikzpicture}
        \begin{axis}[paperaxis, xmode=log, ymode=log, width=0.30\linewidth, height=0.30\linewidth, xmin=3, xmax=600, ymin=15, ymax=2000, xtick={10,100}, ytick={100,1000}, log ticks with fixed point, /pgf/number format/1000 sep={}, title={\footnotesize $\dT = 580$~K}, yticklabels={}, xticklabels={}]
            \addplot[cblue, line width=1.2pt] table[x=a_l0, y=d_l0, col sep=comma] {figures/data/ringdir_dem_bore_T580.csv};
            \addplot[corange, thick, dashed] table[x=a_l0, y=d_l0, col sep=comma] {figures/data/ringdir_fe_bore_T580.csv};
        \end{axis}
    \end{tikzpicture} \\
    \begin{tikzpicture}
        \begin{axis}[paperaxis, xmode=log, ymode=log, width=0.30\linewidth, height=0.30\linewidth, xmin=3, xmax=600, ymin=15, ymax=2000, xtick={10,100}, ytick={100,1000}, log ticks with fixed point, /pgf/number format/1000 sep={}, xlabel={$a / \lgrif$}, ylabel={outer rim, $d / \lgrif$}]
            \addplot[cblue, line width=1.2pt] table[x=a_l0, y=d_l0, col sep=comma] {figures/data/ringdir_dem_outer_T280.csv};
            \addplot[corange, thick, dashed] table[x=a_l0, y=d_l0, col sep=comma] {figures/data/ringdir_fe_outer_T280.csv};
        \end{axis}
    \end{tikzpicture} &
    \begin{tikzpicture}
        \begin{axis}[paperaxis, xmode=log, ymode=log, width=0.30\linewidth, height=0.30\linewidth, xmin=3, xmax=600, ymin=15, ymax=2000, xtick={10,100}, ytick={100,1000}, log ticks with fixed point, /pgf/number format/1000 sep={}, xlabel={$a / \lgrif$}, yticklabels={}]
            \addplot[cblue, line width=1.2pt] table[x=a_l0, y=d_l0, col sep=comma] {figures/data/ringdir_dem_outer_T380.csv};
            \addplot[corange, thick, dashed] table[x=a_l0, y=d_l0, col sep=comma] {figures/data/ringdir_fe_outer_T380.csv};
        \end{axis}
    \end{tikzpicture} &
    \begin{tikzpicture}
        \begin{axis}[paperaxis, xmode=log, ymode=log, width=0.30\linewidth, height=0.30\linewidth, xmin=3, xmax=600, ymin=15, ymax=2000, xtick={10,100}, ytick={100,1000}, log ticks with fixed point, /pgf/number format/1000 sep={}, xlabel={$a / \lgrif$}, yticklabels={}]
            \addplot[cblue, line width=1.2pt] table[x=a_l0, y=d_l0, col sep=comma] {figures/data/ringdir_dem_outer_T480.csv};
            \addplot[corange, thick, dashed] table[x=a_l0, y=d_l0, col sep=comma] {figures/data/ringdir_fe_outer_T480.csv};
        \end{axis}
    \end{tikzpicture} &
    \begin{tikzpicture}
        \begin{axis}[paperaxis, xmode=log, ymode=log, width=0.30\linewidth, height=0.30\linewidth, xmin=3, xmax=600, ymin=15, ymax=2000, xtick={10,100}, ytick={100,1000}, log ticks with fixed point, /pgf/number format/1000 sep={}, xlabel={$a / \lgrif$}, yticklabels={}]
            \addplot[cblue, line width=1.2pt] table[x=a_l0, y=d_l0, col sep=comma] {figures/data/ringdir_dem_outer_T580.csv};
            \addplot[corange, thick, dashed] table[x=a_l0, y=d_l0, col sep=comma] {figures/data/ringdir_fe_outer_T580.csv};
        \end{axis}
    \end{tikzpicture}
    \end{tabular}

    \ref*{ringlegend}
    \caption{Quenched annulus, crack spacing $d$ against the depth $a$ from the bore (top) and from the
    outer rim (bottom) at the four thermal loads, both normalized by the
    Griffith length $\lgrif$ (\ref{app:hyper}). The proposed method
    (blue, solid) is compared with the finite element computation
    (orange, dashed).}
    \label{fig:ring-spacing}
\end{figure}
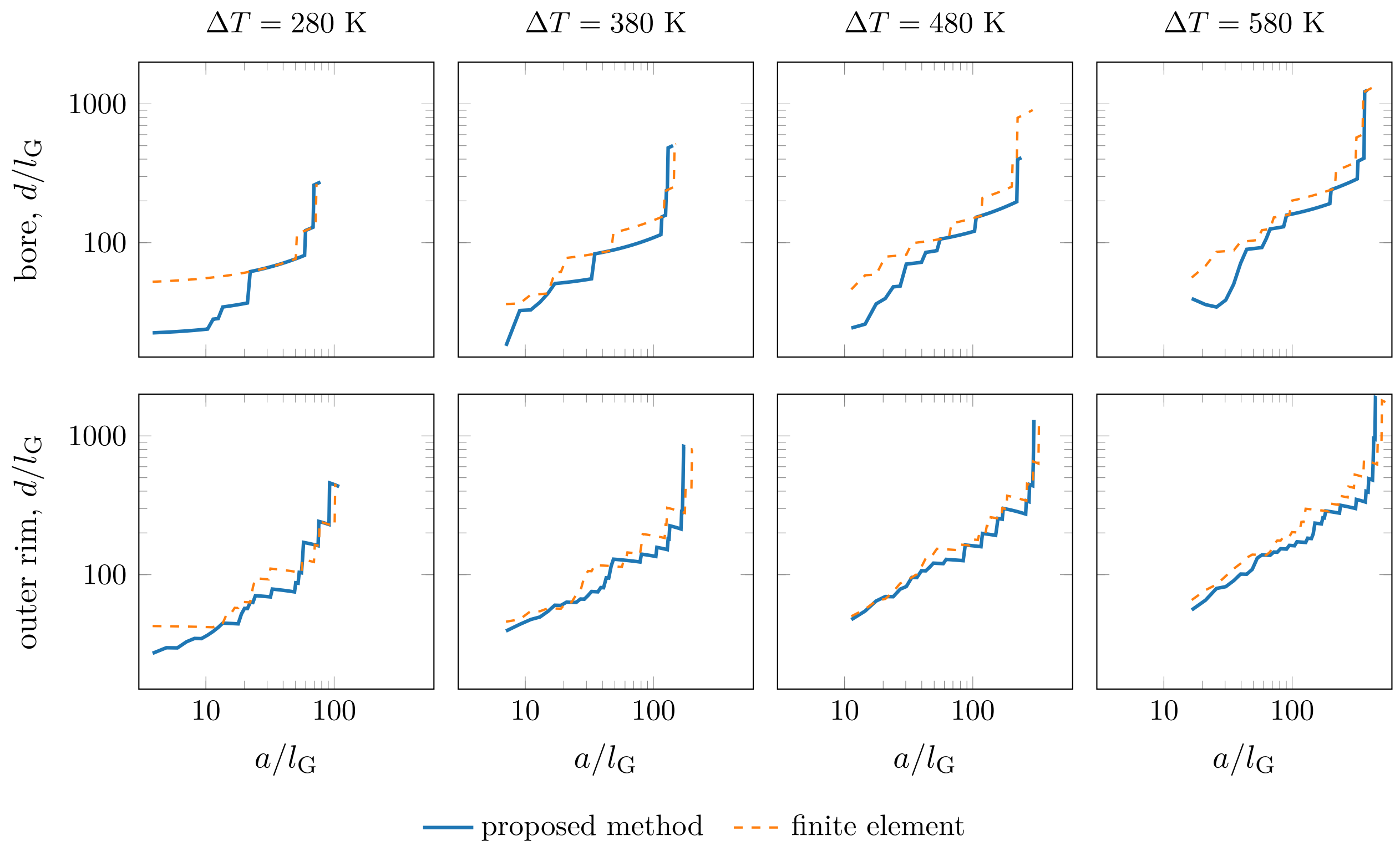

Fig.~\ref{fig:ring} shows the final states, one thermal load per
quadrant. Both computations form an array on each rim, with long
cracks running inward from the outer rim and outward from the bore and
shorter cracks between them, and several long cracks lie at the same
angles with similar lengths in the two. At depths of $1$~mm and more
the counts agree within $2$ cracks on both rims at every load
(Table~\ref{tab:ring-counts}). Nearer the surface, where nucleation
sets the spacing, the proposed method forms more cracks, about twice as
many at $0.1$~mm from the bore below $580$~K. At $580$~K the proposed method leaves a fully damaged layer about $0.25$~mm thick along the bore, which the finite element computation does not form (\ref{app:arrays}). The spacing of the proposed method averaged
over the depth lies $10\%$ to $17\%$ below that of the finite element
computation on the outer rim and $19\%$ to $36\%$ below at the bore,
Fig.~\ref{fig:ring-spacing}. The longest outer cracks of the finite
element computation reach $3$~mm at the two higher loads, beyond those
of the proposed method.

%% file: figures/geom_layout.tex
\newcommand{\geomstripline}{\makebox[\linewidth][l]{\hspace*{0.0788\linewidth}%
  \includegraphics[width=0.7000\linewidth]{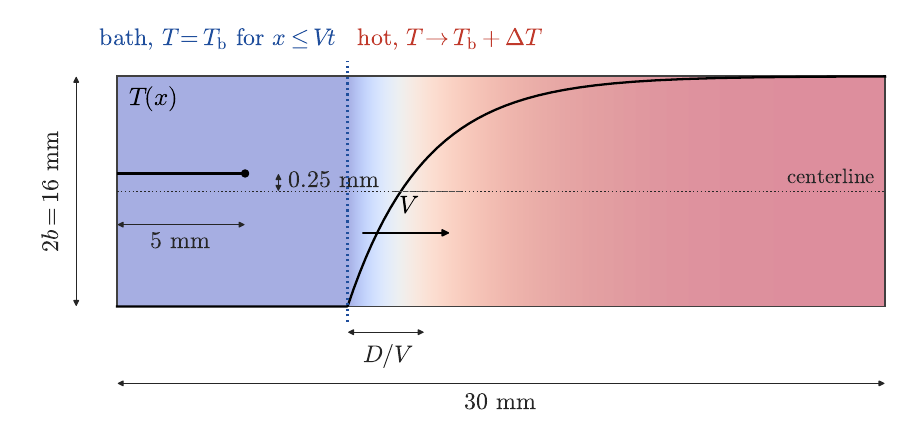}\hspace*{0.0640\linewidth}%
  \raisebox{0.0776\linewidth}{\includegraphics[width=0.0891\linewidth]{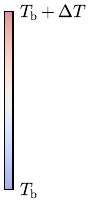}}}}
\newcommand{\geomslabline}{\makebox[\linewidth][l]{\hspace*{0.0636\linewidth}%
  \includegraphics[width=0.7657\linewidth]{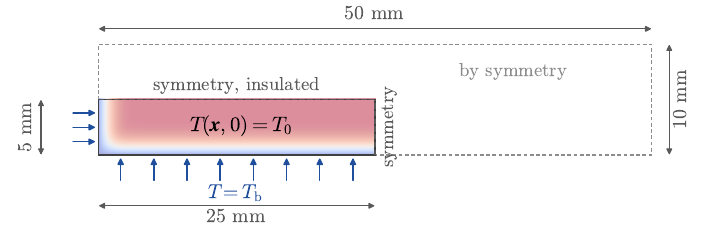}\hspace*{0.0134\linewidth}%
  \raisebox{0.0415\linewidth}{\includegraphics[width=0.0434\linewidth]{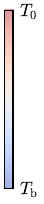}}}}

%% file: figures/data/strip_fix_fit.tex
\def\stripfitS{0.020710}\def\stripfitRc{9.013438}\def\stripfitM{0.096269}\def\stripfitQ{-0.373297}\def\stripfitLc{0.494416}

%% file: sections/S6_conclusions.tex
\let\OrigFloatBarrier\FloatBarrier
\let\FloatBarrier\relax
\section{Conclusions}
\label{sec:conclusions}
\let\FloatBarrier\OrigFloatBarrier

This paper proposes a mesh-free deep energy method for phase-field
modeling of thermally driven brittle fracture, built on the purely
elastic solver of~\paperone. Unlike the existing neural energy-based
approaches to thermal fracture, which start from defects placed in
advance, the proposed method computes the spontaneous nucleation of
cracks and the fracture patterns that develop under transient thermal
loading. It captures the emergence of crack arrays, their propagation
and arrest, and complex crack trajectories without prescribing crack paths or nucleation sites and without meshes adapted to the cracks. Where the temperature is not given as an analytical field, it is computed by a second network from an incremental functional of the heat equation, in which the convective condition is a natural boundary condition. Irreversibility on the crack set, together with an optimizer state carried across load steps and learning rates
that decay within each step, controls the spurious accumulation of
damage.

The performance of the proposed method was assessed on five numerical
examples. The temperature was computed with a damage-dependent conductivity on the notched plates, given as an analytical field on the strip, and computed with a damage-independent conductivity on the slab and the annulus. On the two notched plates the initiation peak loads agree with
an independent finite element reference to within $1.8\%$ at every
thermal load, and on the graded plate the initiation peak rises by $26\%$ from $-30$ to $+30$~K, against $28\%$ in the reference. On the glass strip the crack path changes from straight
to oscillatory and then to branched as the temperature jump grows, in
the order found by Zeng and Dolbow~\cite{zeng2026}, with an oscillation threshold close to the linear stability prediction. On the quenched slab under a prescribed boundary temperature the crack array is of the type observed in experiment, with a spacing averaged over the depth within $14\%$ of the benchmark computation of Bourdin et
al.~\cite{bourdin2014}. On the slab and the annulus the finite element
computation on the same material field forms a pattern of the same
type, and the proposed method forms a denser array.

Several limitations remain. The results depend on the resolution of the
neural representations and on the optimization procedure, and a
systematic convergence study has not been made. A larger iteration budget changes the crack spacing on the slab, and on the strip the paths near the transitions between propagation regimes change with every increase of the budget up to the one used. The settings
that control the spurious accumulation of damage were tested under
limited conditions and do not remove every such artifact. A damaged layer persists along the bore of the annulus at the largest thermal load. On the slab under the convective condition (\ref{app:convective}) the proposed method forms several times more cracks near the surface than the finite element computation, a difference that arises at nucleation. The results for the quench problems rest on
single realizations of the material field, and the strength of the AT2
model is tied to the regularization length. The formulation is
restricted to two-dimensional, quasi-static fracture, and the thermal problem of a step is solved before, and not iterated with, the
mechanical one. The thermo-elastic heat source, the heat released by fracture~\cite{miehe2015} and the contact of the crack faces are not modeled, although a heat source from the rates of the volumetric strain and of the damage changes the crack spacing in a gradient damage model of a quenched ceramic plate~\cite{huang2025gradient}. The crack band is either nearly adiabatic or transmits heat as intact material, without a model of the effective conductivity of a cracking body~\cite{svolos2020}. The proposed method also takes longer than the
finite element computation on the notched plates (Table~\ref{tab:hyper}).

Future work should improve the numerical convergence, the robustness
and the computational efficiency, in particular for problems with the
spontaneous nucleation of many cracks. Further developments include a
fuller thermo-mechanical coupling, the heat transfer across opening cracks and the extension to thermo-hydro-mechanical
fracture. Since the material field, the network fields and the energy enter one
differentiable computation, a spatially varying material property could
also be identified from an observed fracture pattern, which requires
reliable sensitivities of the computed fracture evolution to that
property.

\vspace{\baselineskip}
\section*{Data availability}

Data will be made available on request.

\section*{Code availability}

The solver, the staggered finite element reference implementation and
the scripts reproducing all examples of this paper are available at
\url{https://github.com/HannnZH/Meshfree-DEM-ThermoMechanical-PhaseField}.

\section*{Acknowledgments}

This research was undertaken with the assistance of resources from the
Gadi supercomputer~\cite{gadi} at the National Computational
Infrastructure (NCI Australia), an NCRIS enabled capability supported by
the Australian Government, and from the Katana cluster~\cite{katana} at
UNSW Sydney, supported by Research Technology Services.

%% file: sections/SA_appendix.tex
\section{Numerical parameters and verification of the reported runs}
\label{app:hyper}

Table~\ref{tab:hyper} collects the numerical parameters of every run
reported in Section~\ref{sec:results}. All runs use the multilayer
perceptrons of Eq.~\eqref{eq:net-te} with GELU activations, $4$ hidden
layers of width $128$ in the mechanical network and $3$ of width $64$
in the temperature network. Every level of the feature grids
stores $2$ feature values at each node. Adam is used with the moment
parameters $0.9$ and $0.999$, the stability constant $10^{-8}$ and a learning rate of
$5 \times 10^{-4}$ on the perceptron and $2 \times 10^{-3}$ on the
grids. Where the optimizer state is carried across load steps, the
step counter of Adam is carried with it, so the bias correction is not
restarted. The thermal solve starts each step with a new Adam and no
decay of the learning rates. The integration points are redrawn at
every iteration, each stratum receiving a fixed share of them, and
the stratum weights $w_{\mathrm{u}}$, $w_{\mathrm{c}}$ and
$w_{\mathrm{p}}$ of Eq.~\eqref{eq:mixture-te} are $0.4$, $0.3$ and $0.3$ for the uniform,
crack and process-zone strata on the notched plates and the strip and
$0.3$, $0.3$ and $0.4$ on the quench problems. The first thermal solve
has the largest budget, since it forms the boundary layer from the
initial state of the lift. All runs are in double precision on one
NVIDIA V100 GPU of Gadi, except the strip, which ran on one NVIDIA H200
GPU of Katana as consecutive jobs whose times are summed in the table.
The finite element references of the notched plates ran on $8$ cores
of a CPU node of Gadi.

The remaining numerical parameters are common to all runs unless
stated otherwise. The residual stiffness is $\kres = 10^{-6}$, the
displacement scale $U_{\mathrm{ref}} = 10^{-3}$~mm, and the temperature
scale $\dT_{\mathrm{ref}}$ of Eq.~\eqref{eq:T-ansatz} is the magnitude
of the thermal load, or $1$~K where it vanishes. A quadratic penalty
on the feature values, the sum of their squares multiplied by
$10^{-8}$, is added to each estimated loss. The bias of the phase-field
output is initialized at $-4$, so that $\pf = s(-4) \approx 0.018$ at
the start in material without a notch. The decay length of the temperature lift in Eq.~\eqref{eq:T-lift} is $r_{\mathrm{T}} = 0.005$~mm.
The finest mechanical level has a node spacing of about $\lreg / 4$ on
the plates, the slab and the strip, and on the annulus the physical
spacing grows with the radius.
On the notched plates the reaction force of Eq.~\eqref{eq:observables}
is estimated on $80\,000$ points.

Each solve stops when
\begin{equation}
    \frac{\max_{\tau' \in \mathcal{W}} \widehat{S}^{(\tau')}
          - \min_{\tau' \in \mathcal{W}} \widehat{S}^{(\tau')}}
         {\bigl| \tfrac{1}{n_{\mathrm{w}}} \sum_{\tau' \in \mathcal{W}}
          \widehat{S}^{(\tau')} \bigr|}
    < \epsilon_{\mathrm{tol}} ,
    \label{eq:plateau}
\end{equation}
or when its iteration budget is spent, with $\tau$ the current
iteration, the window
$\mathcal{W} = \{\tau - n_{\mathrm{w}} + 1, \dots, \tau\}$ of
$n_{\mathrm{w}} = 400$ iterations, $\widehat{S}^{(\tau')}$ the value of
$\widehat{\EneT}$ or of $\widehat{\mathcal{L}}$ at iteration $\tau'$
and the tolerance $\epsilon_{\mathrm{tol}} = 2 \times 10^{-4}$.

On the notched plates, whose parametric and physical coordinates
coincide, the uniform stratum of Eq.~\eqref{eq:mixture-te} is drawn
from a scrambled Sobol' sequence~\cite{sobol1967}, and on the mapped
patches from the auxiliary grid. That grid has $256 \times 256$ cells
on the notched plates, $384 \times 384$ on the slab and the strip and
$320 \times 320$ on the annulus. The crack and process-zone strata are
held constant within each cell and normalized by the sum over the
cells, and a point is drawn by choosing a cell with its probability and
a uniform position inside it. On the quench problems the crack stratum
is a Gaussian band along each cooled face or rim with a standard
deviation of $0.3$~mm normal to it. The process-zone stratum uses
$\beta_{0} = 0.3$ and $\beta_{\mathrm{d}} = 0.5$ on the driving-force
indicator
\begin{equation}
    Y_{\mathrm{proc}}
    =
    \dmg(\pf_{n-1})\,
    \psi^{+}_{\mathrm{e,vd}}\bigl(
        \straintensor(\ub_{n-1}) - \alphaT (\Tp - \Tz)\, \bm{I}
    \bigr) ,
    \label{eq:drive-te}
\end{equation}
divided by its maximum over the cells, with $\psi^{+}_{\mathrm{e,vd}}$
the tensile part of the volumetric-deviatoric split of Amor
et al.~\cite{amor2009}. This tensile part serves the sampling only and
is distinct from the spectral part $\edens^{+}$ that drives the crack.
Since the estimator of Eq.~\eqref{eq:mc-te} divides by the sampling
density, this choice changes the variance of the estimate and not its
expectation. As in~\paperone, the volumetric-deviatoric form is used
since it is smooth and, under shear, broader than the spectral one, so
that the process zone is not left without points.
On the slab under the convective condition of \ref{app:convective},
each of the two cooled faces has $M_{e} = 4000$ points of
Eq.~\eqref{eq:mcT}.

On the notched plates, with the bottom edge $y = 0$ fixed and the top
edge $y = H$ driven, the lifts read
\begin{equation}
    u = U_{\mathrm{ref}}\, \bar{y} (1 - \bar{y})\, \hat{u} + \bar{y}\, \delta_{u} ,
    \qquad
    v = U_{\mathrm{ref}}\, \bar{y} (1 - \bar{y})\, \hat{v} + \bar{y}\, \delta_{v} ,
    \qquad
    T_{\mathrm{D}} = \Tz + \bigl[\Tf_{\mathrm{edge}}(t) - \Tz\bigr]\, \bar{y} ,
    \label{eq:lift-te}
\end{equation}
with $\bar{y} = y / H$, $\omega_{\mathrm{T}} = \bar{y}(1 - \bar{y})$ and
$(\delta_{u}, \delta_{v}) = (0, \delta)$ under tension and $(\delta, 0)$
under shear, with the reaction force taken with respect to the nonzero
component.

In a profile of load step $50$ of the slab at $380$~K the backward pass
takes $71\%$ of a mechanical iteration, and at full budgets the thermal
solve takes about $8\%$ of a load step. A perceptron of $3$ hidden layers of width $64$
halves the time of a mechanical iteration on the slab. Its effect on the
accuracy was measured on the cooled tension plate, where the same
perceptron shortens the run by $3\%$ and raises the error of the peak
load from $1.30\%$ to $3.6\%$, so the reported runs keep the wider perceptron.

\begin{table}[htbp]
    \centering
    \scriptsize
    \caption{Numerical parameters of the reported runs. Levels are the
    numbers of nodes $n_{\ell}$ of the three levels of feature grids
    along $\eta$, from coarse to fine, and
    along $\xi$ each grid has as many nodes as the nearest integer to
    $n_{\ell}$ times the aspect ratio of the patch, so that the physical
    node spacing is about the same in the two directions. $M$
    is the number of integration points per iteration. The mechanical
    iteration budgets are those of the first step, the second step and
    every later step, and the thermal budgets those of the first
    transient solve and of every later step. $\gamma_{\mathrm{ir}}$ is
    the penalty weight of Eq.~\eqref{eq:penalty-te}, and the parameter
    counts of the two networks include the feature values. The
    temperature is computed or given as an analytical field, and the conductivity
    is damage-dependent, Eq.~\eqref{eq:kdeg}, or damage-independent. The
    strip has no temperature network, since its temperature is an
    analytical field. The wall time of the proposed method is that of one run, and that of the finite element computation the range over the thermal loads. The rows below the dashed line refer to the finite element computations.}
    \label{tab:hyper}
    \setlength{\tabcolsep}{2pt}
    \begin{tabular}{@{}lccccc@{}}
        \toprule
        & Tension & Shear, graded & Strip & Slab & Annulus \\
        \midrule
        Temperature              & computed & computed & analytical & computed & computed \\
        Conductivity             & $\kcond\,\dmg(\pf_{n-1})$ & $\kcond\,\dmg(\pf_{n-1})$ & & $\kcond$ & $\kcond$ \\
        $\lreg$ (mm)             & $0.01$ & $0.01$ & $0.1$ & $0.05$ & $0.05$ \\
        Levels, mechanical       & $32/128/384$ & $32/128/384$ & $64/256/624$ & $64/192/384$ & $64/256/1280$ \\
        Levels, thermal          & $32/128/384$ & $32/128/384$ & & $64/128/384$ & $64/256/320$ \\
        $M$                      & $12\,000$ & $12\,000$ & $100\,000$ & $80\,000$ & $80\,000$ \\
        Iterations, mechanical   & $1000/4000/2000$ & $1000/4000/2000$ & $10\,000/5000/5000$ & $1500/3000/2400$ & $1500/3000/2400$ \\
        Iterations, thermal      & $1200/400$ & $1200/400$ & & $1600/800$ & $1600/800$ \\
        $\gamma_{\mathrm{ir}}$ (MPa) & $10^{3}$ & $10^{3}$ & $20.6$ & $3.6 \times 10^{4}$ & $3.6 \times 10^{4}$ \\
        Load steps               & $80$ & $160$ & $240$ & $100$ & $100$ \\
        Parameters, mechanical   & $380\,803$ & $380\,803$ & $1\,772\,355$ & $1\,935\,235$ & $2\,661\,379$ \\
        Parameters, thermal      & $338\,689$ & $338\,689$ & & $1\,688\,321$ & $271\,745$ \\
        Wall time (h)            & $5.0$ & $10.4$ & $25.6$ & $16.8$ & $17.0$ \\[2pt]
        \cdashline{1-6}\noalign{\vskip 2pt}
        \multicolumn{6}{@{}l}{\textit{Finite element computation}} \\
        Element size             & $\lreg/5$ & $\lreg/5$ & $\lreg/4$ & $\lreg/2$ & $\lreg/2$ \\
        Wall time (h)            & $2.3$ to $2.6$ & $4.6$ to $6.3$ & & & \\
        \bottomrule
    \end{tabular}
\end{table}

\subsection{Settings, measurements and thermal verification}
\label{subsec:app-measure}

The notched plates and their finite element references were computed
with the volumetric heat capacity and the time scale both divided by
$10^{3}$, as reported by Tangella et al.~\cite{tangella2022}, which
leaves the product $D t$ and every field unchanged, and the times
reported are those of the physical problem.

On the quench problems the rescaling of the toughness field in Eq.~\eqref{eq:weibull-gc} multiplies the strength implied by each cell
by the square root of one constant, so the local strengths keep the
shape of the drawn distribution. With the nominal values at
$\lreg = 0.05$~mm the peak stress $\sigma_{\mathrm{c}}$ is $182$~MPa,
close to the nominal strength of $180$~MPa. The Griffith length
$\lgrif$ is written $l_{0}$ by Bourdin et al.~\cite{bourdin2014}, and at
$\dT = 380$~K the slab matches their setup to $4\%$ in
$\lreg / \lgrif$ and to $2\%$ in the half thickness over $\lgrif$.

On the slab, crack counts are the local maxima of the phase field above
the level $0.7$ along a line at the depth $a$ parallel to the cooled
long face, merged within $0.15$~mm, over the part of the face more than
$2.5$~mm from the cooled end face. The spacing $d(a)$ is the length of
that part divided by the count. On the annulus the same rule is
applied along the arc at the depth $a$ from each rim, merged within
$0.15$~mm of arc. The benchmark is the curve of Bourdin
et al.~\cite{bourdin2014} digitized from their Fig.~1(c). As a check on
the digitization, the semianalytical law of Bahr et al.~\cite{bahr2010},
digitized from the same axes, gives the exponent $0.745$ against its
$3/4$. A
computed spacing is compared with the benchmark, and with the finite
element computation, through the ratio of the two curves, averaged
geometrically over logarithmically spaced depths in the range that
both curves cover. The count under the rule of Ruan
et al.~\cite{ruan2023thermomechanical} is the number of cracks reaching
$1$~mm on the $15$~mm of the quarter next to its symmetry cut, doubled
for the central $30$~mm of the long face.

On the cooled tension plate at load steps $20$, $36$ and $37$, the last
at the peak, and on the isothermal plate at its peak, the resultant of
the hybrid stress $\sigma_{yy}$ along the loaded edge differs from the
reaction force of Eq.~\eqref{eq:observables} by $1.6\%$ to $4.0\%$. The reported force is the domain integral of
the energy derivative, and the edge resultant, which samples the stress
on one line, serves only as a check on equilibrium. The violation of irreversibility,
$\pospart{\pf_{n-1} - \pf_{n}}$, measured on a grid of $600 \times 600$
points, is at most $4 \times 10^{-3}$ on the tension plate at the same
steps and at most $1.1 \times 10^{-3}$ on the crack set of the slab at
$380$~K. Outside the crack set the phase field decreases by up to $0.11$ at nucleation, as the threshold allows.

The one-dimensional reference of the tension plate is an implicit
finite difference solution on $2001$ nodes with $200$ substeps per load
step. That of the slab is a backward Euler finite volume solution on
$20\,000$ cells on the time grid of Eq.~\eqref{eq:clock}, along the line normal to the long face halfway along it, over the $100$ load steps (Section~\ref{subsec:slab}). At $380$~K the first backward Euler step leaves the layer up to $47$~K warmer than the continuous transient, an error that decays to
$5.5$~K by load step $16$ and that the finite element computation
shares.

\section{Solver settings, resolution and budget on the crack arrays}
\label{app:arrays}

Table~\ref{tab:arrays} compares subsets of three settings on the slab of Section~\ref{subsec:slab} at $\dT = 480$~K, with all other settings as in Table~\ref{tab:hyper}. The settings are the crack-set threshold of Eq.~\eqref{eq:penalty-te}, at $\pf_{\mathrm{th}} = 1/2$, a value that was not varied, and the two options of the mechanical update in Section~\ref{subsec:staggered} (the optimizer state carried across load steps and the decay of the learning rates within a step). With none of the three, $84$ of the $100$ load steps start with a rise of
the loss, and a diffuse background and a damaged film accumulate along
the cooled face. The threshold alone, the carried optimizer alone and
the threshold with the decay still widen the bands or form a fully
damaged layer, and only the three together keep the bands at a nearly
constant width and the cooled face clean. The threshold with the
carried optimizer and without the decay, tried on the slab under the
convective condition of \ref{app:convective}, also widened the bands
and formed a film along the end face. The decay alone and the carried
optimizer with the decay were not computed.

\begin{table}[htbp]
    \centering
    \small
    \caption{Slab under a prescribed boundary temperature at
    $\dT = 480$~K, one realization, subsets of the crack-set threshold
    and the two options of the mechanical update. State of the cooled
    face at the last load step, width at half maximum of the bands at
    the depth of $1$~mm at load steps $55$ and $100$, count along the
    cooled face at the depth of $0.1$~mm at load steps $50$ and $100$,
    and number of load steps
    whose loss rises by more than $15\%$ in the first $300$ iterations.}
    \label{tab:arrays}
    \setlength{\tabcolsep}{4pt}
    \begin{tabular}{@{}lllcc@{}}
        \toprule
        Subset & Cooled face & Width (mm) & Count & Rising steps \\
        \midrule
        none                  & film and background           & $0.125$, $0.161$ & $50$, $50$ & $84$ \\
        threshold             & clean                         & $0.100$, $0.123$ & $48$, $50$ & $0$ \\
        carried optimizer  & layer of $0.2$~mm, background & $0.117$, $0.226$ & $48$, $63$ & $0$ \\
        threshold and decay   & fully damaged layer, bulges    & $0.127$, $0.237$ & $58$, $59$ & $74$ \\
        all three             & clean                         & $0.091$, $0.096$ & $50$, $50$ & $1$ \\
        \bottomrule
    \end{tabular}
\end{table}

On the annulus of Section~\ref{subsec:annulus} at $\dT = 580$~K the
bore retains a fully damaged layer, about $0.25$~mm thick on the
reported realization of the field and about $0.1$~mm on a second one.
A linear ramp of the learning rates over the first $200$ iterations of
every load step thins it to about $0.1$~mm and changes the counts by at
most $2$ cracks.

The resolution and the iteration budget were tested on the slab at
$\dT = 380$~K, on the reported realization. The three solver seeds of
Section~\ref{subsec:slab}, which measure the scatter of the solver and
not that of the material field, differ in their counts by at most $5$
cracks at any depth. With the number of integration points and the node counts on every mechanical level and on the two finer thermal levels doubled, the crack counts stay within the range of the seeds except for one
crack more at $2$~mm, and the spacing lies $8.8\%$ below the benchmark
of Bourdin et al.~\cite{bourdin2014}. With every iteration budget
doubled, the counts lie one or two cracks below that range between
$0.3$ and $0.75$~mm, and the spacing lies $1.8\%$ below the benchmark,
against $5.0\%$ to $10.5\%$ for the three seeds.

\section{Sampler and representation controls}
\label{app:sampler}

These controls, on the slab of Section~\ref{subsec:slab}, test two
choices of Section~\ref{sec:method}, the redrawing of the points at
every iteration and the feature grids of the temperature network, and
check that the process-zone stratum, which follows the mechanical state, does not bias the thermal solve. On the
slab the finest thermal level has a node
spacing of $0.013$~mm to resolve the layer of the first time step.
That is finer than the mean distance between the $80\,000$ points of a
fixed set even in the band along the cooled face, where the crack stratum
concentrates them to about $0.02$~mm, so a fixed set leaves that level unconstrained
between the points.

\begin{table}[htbp]
    \centering
    \small
    \caption{Thermal controls on the quenched slab under the convective
    boundary condition, $\dT = 380$~K, against the backward Euler finite
    volume solution. For each variant, the range over the load steps of
    three quantities, the largest departure (load steps $1$ to $30$ and
    $31$ to $100$), the root mean square weighted by area over the
    cooled layer and the departure of the temperature drop integrated over that layer (all $100$ load steps). The
    thermo-mechanical run is a computation on the realization of
    Table~\ref{tab:slab-counts} at this load under the same boundary
    condition.}
    \label{tab:thermal-controls}
    \setlength{\tabcolsep}{4pt}
    \begin{tabular}{@{}lccccc@{}}
        \toprule
        & & \multicolumn{2}{c}{Largest departure (K)} & & \\
        & & \multicolumn{2}{c}{at load steps} & & \\
        \cmidrule(lr){3-4}
        Variant & Parameters & $1$ to $30$ & $31$ to $100$
        & Layer rms (K) & Layer integral (\%) \\
        \midrule
        \begin{tabular}[t]{@{}l@{}}Thermo-mechanical\\ run\end{tabular} & $1\,688\,321$ & $1.35$ to $3.30$ & $1.67$ to $3.29$
        & $0.22$ to $0.69$ & $-0.37$ to $+0.50$ \\
        Redrawn points & $1\,688\,321$ & $1.75$ to $3.32$ & $1.58$ to $3.06$
        & $0.25$ to $0.69$ & $-0.44$ to $+2.19$ \\
        Fixed set per step & $1\,688\,321$ & $150$ to $1434$ & $121$ to $203$
        & $30$ to $158$ & $-10.3$ to $+1804$ \\
        No feature grids & $8577$ & $11.3$ to $61.5$ & $1.66$ to $8.79$
        & $0.54$ to $9.29$ & $-13.7$ to $+0.40$ \\
        \bottomrule
    \end{tabular}
\end{table}

The thermal problem of the slab alone, whose temperature does not
depend on the cracks since the conductivity is damage-independent, was
computed over the $100$ load steps at $\dT = 380$~K under the
convective condition of \ref{app:convective}, with the budgets, the
learning rates and the point count of a thermo-mechanical run,
and compared with a backward Euler finite volume solution along the
line normal to the long face halfway along it,
Table~\ref{tab:thermal-controls}. On points redrawn at every
iteration with no mechanical state in the sampler, the network
reproduces the temperature of the thermo-mechanical run, so the process-zone stratum does not bias the thermal solve. On one point set per time step
the temperature departs by at least $121$~K at every load step, while
the functional estimated at the fixed points lies below that of the
redrawn run at $93$ of the $100$ steps, the mechanism of~\paperone.
Without feature grids the departure reaches $16\%$ of the thermal load
while the cooled layer is thinner than about $0.3$~mm. The effect of
the representation on the crack pattern was not computed.

\section{Verification of the finite element reference}
\label{app:reference}

The reference for the two notched plates (Sections~\ref{subsec:sent} and~\ref{subsec:sensfgm}) is a staggered phase-field
solver built on scikit-fem~\cite{gustafsson2020scikitfem}. It solves the plane-strain problem with
the AT2 density and the hybrid formulation, in which the equilibrium
equation uses the degraded isotropic energy and the phase field is driven by the
tensile part of the spectral decomposition. Irreversibility is
enforced through the history field, and the notch enters as a
Dirichlet condition on the phase field. Each load step is solved by
alternate minimization to a staggered tolerance of $10^{-3}$ on the
relative change of the nodal phase field in the Euclidean norm, with a
cap of $2000$ staggered iterations on the notched plates, which was not
reached. The mesh is a uniform grid of bilinear quadrilaterals, $512$
on a side after $9$ refinements of the unit square, an element size of
$\lreg / 5$. The heat conduction
problem is advanced by backward Euler on the time grid of
Section~\ref{subsec:sent} with a lumped heat capacity, which keeps
the temperature inside the range of its boundary values at the steep
front where the consistent capacity does not. The conductivity is damage-dependent, Eq.~\eqref{eq:kdeg}. The thermal
problem is solved first at every step and the mechanical problem on
the frozen temperature, which is the order of the proposed method.
The thermal strain enters through the elastic strain, and with
$\varepsilon_{\mathrm{e},zz} = -\ath$ and
$3\lambda + 2\mu = (1 + \nu)(2\lambda + 2\mu)$ the in-plane stress of
the undamaged material is
\begin{equation}
    \stresstensor
    =
    \lambda \bigl( \trace_{2}\straintensor - 2\varepsilon^{*} \bigr)\,
    \bm{I}
    + 2\mu \bigl( \straintensor - \varepsilon^{*}\, \bm{I} \bigr) ,
    \qquad
    \varepsilon^{*} = (1 + \nu)\, \alphaT\, (\Tf - \Tz) ,
    \label{eq:plane-strain}
\end{equation}
the plane-strain law of a body carrying the isotropic in-plane
eigenstrain $\varepsilon^{*}$. The spectral decomposition is the one of
Section~\ref{subsec:thermoelastic} under plane strain and plane stress
alike, and the
graded plate carries the linear material fields of
Table~\ref{tab:mat-plates} at the quadrature points. The mesh was
checked on the homogeneous plate under shear at $+30$~K of Tangella
et al.~\cite{tangella2022}, where the peak falls from $79.9$~N at $7$ uniform refinements (elements of $\lreg / 1.3$) to $76.8$~N at $8$ ($\lreg / 2.6$) and to $75.9$~N at $9$ ($\lreg / 5$), the refinement of the reference. At $9$ refinements the crack initiates one load step earlier than at $8$. At
$7$ refinements the plate separates at a displacement $2\%$ below their
published value, with $\lreg = 0.01$~mm against their $0.005$~mm.

Added to the reference as an option,
evaluated on the frozen temperature, the toughness law of
Eq.~\eqref{eq:gcT} raises the peak of the cooled tension case by
$0.23\%$, from $96.33$ to $96.55$~N.

\begin{figure}[!htb]
    \centering
    \includegraphics[width=\textwidth]{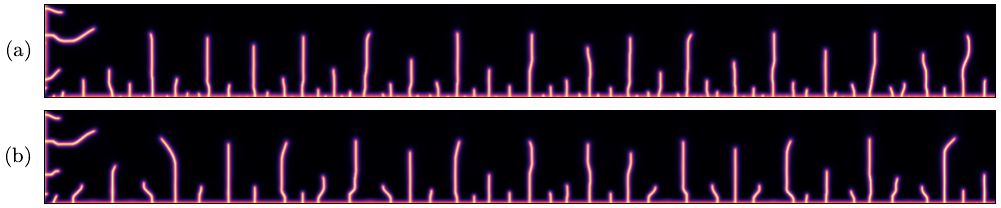}
    \caption{Finite element reproduction of the ceramic quenching test
    of Tangella et al.~\cite{tangella2022} at $\dT = 380$~K and
    $50$~\si{\micro\second}, quarter model with the cooled long face at the bottom
    and the cooled end face at the left, with the phase field from $0$
    (black) to $1$ (light). (a)~Their staggered loop.
    (b)~Alternate minimization.}
    \label{fig:fe-tangella}
\end{figure}

\begin{table}[!tbp]
    \centering
    \small
    \caption{Reproduction of the ceramic quenching test of Tangella et
    al.~\cite{tangella2022} at $\dT = 380$~K and $50$~\si{\micro\second}. Number of
    cracks along the cooled long face that reach the given fraction of
    the plate height, read from their Fig.~11(c) and from the finite
    element fields of Fig.~\ref{fig:fe-tangella} by one rule.}
    \label{tab:fe-tangella}
    \begin{tabular}{@{}lccccccc@{}}
        \toprule
        Fraction of the height & $0.10$ & $0.15$ & $0.20$ & $0.30$ & $0.40$ & $0.50$ & $0.60$ \\
        \midrule
        Tangella et al., Fig.~11(c) & $54$ & $41$ & $32$ & $22$ & $16$ & $14$ & $13$ \\
        Finite element, their staggered loop & $39$ & $30$ & $23$ & $19$ & $16$ & $14$ & $11$ \\
        Finite element, alternate minimization & $36$ & $29$ & $21$ & $18$ & $15$ & $14$ & $11$ \\
        \bottomrule
    \end{tabular}
\end{table}

The same solver was run on quenched ceramic plates in two roles. The first is a reproduction of the ceramic quenching test of Tangella et al.~\cite{tangella2022} at $\dT = 380$~K. It uses their nearly homogeneous material field, time step of $10^{-8}$~s up to $t = 50$~\si{\micro\second} and regularization length of $0.06$~mm, with the bath temperature prescribed on the cooled faces, a damage-independent conductivity and elements of side $\lreg / 2$, Fig.~\ref{fig:fe-tangella} and
Table~\ref{tab:fe-tangella}. Their staggered loop, following their
released code~\cite{tangella2022code}, iterates the displacement and
the phase field until both change by less than $10^{-2}$, with no cap.
The cracks reaching $0.4$ of the height and more agree with their
Fig.~11(c) in number to within $2$ and in depth. The end face has $4$ short cracks in their field and in both finite element fields, while along the long face the finite element fields have fewer shallow cracks, $39$ and $36$ against $54$ at $10\%$ of the height. The
total count depends on the time step, and no mesh refinement was made
on this problem.

The second role is the computation of the same model on the slab and
the annulus, under plane stress on the realization of the proposed
method, with the bath temperature prescribed on the cooled boundary, a
damage-independent conductivity and the time grid of
Eq.~\eqref{eq:clock}. It enforces the bound on the phase field by an
active set and uses bilinear quadrilaterals of side $\lreg / 2$, on the
annulus a tensor grid in radius and angle. Each step is solved by
alternate minimization to a staggered tolerance of $10^{-3}$, with a cap of $50$ iterations on the
slab and $1000$ on the annulus. On the slab at $380$~K elements of side
$\lreg / 4$ change the counts of Table~\ref{tab:slab-counts} by at most
$1$ crack and the spacing at $a / \lgrif = 10$, $20$, $30$, $50$ and
$70$ by at most $8\%$. At $480$~K a first step $100$ times shorter
changes the counts by at most $2$ cracks and the spacing at the same
depths by less than $6\%$.

\section{Loading increment and iteration budget on the glass strip}
\label{app:increment}

The loading increment on the strip (Section~\ref{subsec:strip}) was
chosen on the finite element computation of the same model. The solver
of \ref{app:reference} was run on the strip under plane stress with the
analytical temperature field of
Eq.~\eqref{eq:front} and elements of side $\lreg / 4$, at two increments
for each class of path, Table~\ref{tab:increment}. Between $0.1$ and
$0.05$~mm per step at $130$, $145$, $165$ and $190$~K, and between $0.2$
and $0.1$~mm per step at $217$ and $307$~K, no path changes class, the oscillation at $165$~K keeps its
order parameters to within $5\%$, and the forks and the cracks move by
at most $0.1$~mm. The finite element computation is therefore insensitive to the increment
at $0.1$~mm per step, one regularization length, the increment of
Section~\ref{subsec:strip}.

\begin{table}[!htb]
    \centering
    \footnotesize
    \caption{Loading increment on the glass strip, $\lreg = 0.1$~mm,
    travel $24$~mm, both computations under the analytical temperature
    field of Eq.~\eqref{eq:front}. Path at the last load step of the finite element
    computation at two increments and of the proposed method at $0.1$~mm
    per step with $5000$ iterations. The entries give for straight paths
    the largest transverse departure of the ridge from its position at
    $x = 6$~mm, taken from there to the crack tip, for oscillating paths
    $A / b$ and $\Lambda / b$ on the developed path, for branched paths
    the position $x$ of the fork, and for multiple cracks their number.}
    \label{tab:increment}
    \setlength{\tabcolsep}{4pt}
    \begin{tabular}{@{}lccccc@{}}
        \toprule
        & & \multicolumn{3}{c}{Finite element, mm per step} & Proposed method \\
        \cmidrule(lr){3-5}
        $\dT$ (K) & $R$ & $0.2$ & $0.1$ & $0.05$ & $0.1$~mm per step \\
        \midrule
        $130$ & $8.24$ & & straight, $0.11$~mm & straight, $0.04$~mm & straight, $0.09$~mm \\
        $145$ & $9.19$ & & straight, $0.06$~mm & straight, $0.10$~mm & $0.0638$, $0.510$ \\
        $165$ & $10.46$ & & $0.1047$, $0.540$ & $0.1094$, $0.542$ & $0.1742$, $0.633$ \\
        $190$ & $12.05$ & & fork, $x = 5.2$~mm & fork, $x = 5.2$~mm & fork, $x = 5.5$~mm \\
        $217$ & $13.76$ & fork, $x = 5.0$~mm & fork, $x = 5.0$~mm & & fork, $x = 5.1$~mm \\
        $307$ & $19.46$ & $3$ cracks & $3$ cracks & & $3$ cracks \\
        \bottomrule
    \end{tabular}
\end{table}

\begin{table}[!htb]
    \centering
    \footnotesize
    \caption{Iteration budget and loading increment of the proposed
    method on the glass strip, ordered by the number of iterations per
    millimeter of travel, the first step carrying twice the budget
    in every run. The entries give for straight paths the largest
    transverse departure defined in Table~\ref{tab:increment}, for
    oscillating paths $A / b$ and $\Lambda / b$ on the developed path and
    for branched paths the position $x$ of the fork, all lengths in mm.
    At $145$~K the ratio of the last half swing on the developed path to
    the largest follows in parentheses. At
    $307$~K every setting forms three cracks. The reported runs are
    those at $0.1$~mm per step with $5000$ iterations.}
    \label{tab:budget}
    \setlength{\tabcolsep}{3pt}
    \begin{tabular}{@{}cccccccc@{}}
        \toprule
        mm per step & Iterations & Iterations per mm & $130$~K & $145$~K & $165$~K & $190$~K & $217$~K \\
        \midrule
        $0.2$  & $3000$ & $15\,000$ & $0.04$ & $0.08$ & $0.1129$, $0.587$ & $10.8$ & $5.4$ \\
        $0.1$  & $1500$ & $15\,000$ & $0.10$ & $0.0142$, $0.498$ ($0.43$) & $0.1271$, $0.613$ & $8.0$ & $5.3$ \\
        $0.1$  & $3000$ & $30\,000$ & $0.07$ & $0.0250$, $0.471$ ($0.58$) & $0.1598$, $0.617$ & $7.9$ & $5.2$ \\
        $0.05$ & $2500$ & $50\,000$ & $0.05$ & $0.0556$, $0.504$ ($1.00$) & $0.1732$, $0.626$ & $5.4$ & $5.1$ \\
        $0.1$  & $5000$ & $50\,000$ & $0.09$ & $0.0638$, $0.510$ ($0.94$) & $0.1742$, $0.633$ & $5.5$ & $5.1$ \\
        $0.05$ & $3000$ & $60\,000$ & $0.07$ & $0.0681$, $0.518$ ($0.93$) & $0.1773$, $0.630$ & $5.4$ & $5.1$ \\
        \bottomrule
    \end{tabular}
\end{table}

Unlike the finite element computation, which is solved to a staggered
tolerance of $10^{-3}$ at every step with a cap of $200$ iterations,
the proposed method uses its full budget at every step of the strip,
since the noise of the estimate exceeds the tolerance of
Eq.~\eqref{eq:plateau}, so its path depends on the budget as well as on
the increment. Table~\ref{tab:budget} lists the paths at three budgets
for $0.1$~mm per step and at the increments of $0.05$ and $0.2$~mm per
step. Budget and increment act mainly through the number of iterations per unit of travel, and up to $50\,000$ iterations per millimeter a larger number raises the amplitude at $165$~K, sustains the oscillation at $145$~K on the developed path
and moves the fork at $190$~K to the end of the notch. At $50\,000$
iterations per millimeter the two increments give the same class of
path and forks within $0.1$~mm, with order parameters within about
$1\%$ at $165$~K and an amplitude $13\%$ smaller at $0.05$~mm per step
at $145$~K. The most expensive setting, $0.05$~mm per step with $3000$
iterations, changes no class of path and raises the amplitude by at
most $7\%$, so the reported runs use $0.1$~mm per step with $5000$ iterations. Over the five settings at which both $145$ and $165$~K oscillate, the line through their squared amplitudes gives a threshold $R_{\mathrm{c}}$ between $8.97$ and $9.18$.

\section{Measurement of the order parameters}
\label{app:orderparams}

Table~\ref{tab:halfswings} lists the successive half swings of the
three oscillating cases on the strip (Section~\ref{subsec:strip})
over the whole path. The largest, the fourth, lies on the developed path in each case, and the
threshold is $9.01$ whether the amplitude is read on the developed or
on the whole path. Leaving out one case gives
$8.98$ to $9.11$, reading the amplitude as the mean of the three
largest half swings gives $9.03$, and as the last half swing of the whole path $9.18$.
The wavelength over the whole path differs from that on the developed
path by less than $1\%$ at $155$ and $165$~K and by $4.3\%$ at $145$~K.
The experimental wavelength of Table~\ref{tab:strip-onset} is the fit
$\lambda_{\mathrm{osc}} / L_{\mathrm{osc}} = 0.15 + 2.5\, d_{\mathrm{th}} / L_{\mathrm{osc}}$
of Ronsin and Perrin~\cite{ronsin1998}, which reads
$\Lambda / b = 0.30 + 2.5 / P$ with the width $L_{\mathrm{osc}} = 2b$
at onset and the thermal length $d_{\mathrm{th}} = D / V$, evaluated at
$P = 6.81$.

\begin{table}[htbp]
    \centering
    \small
    \caption{Successive half swings of the traced path (mm) of the three
    oscillating cases, $0.1$~mm per step with $5000$ iterations, over the
    whole path, and the last half swing over the largest.}
    \label{tab:halfswings}
    \begin{tabular}{@{}lccccccccccc@{}}
        \toprule
        $\dT$ (K) & \multicolumn{10}{c}{Half swings (mm)} & Last over largest \\
        \midrule
        $145$ & $0.75$ & $0.90$ & $0.99$ & $1.02$ & $0.96$ & $0.86$ & $0.71$ & $0.55$ & $0.39$ & $0.28$ & $0.28$ \\
        $155$ & $1.31$ & $1.71$ & $1.97$ & $2.03$ & $1.90$ & $1.75$ & $1.60$ & $1.49$ & & & $0.73$ \\
        $165$ & $2.06$ & $2.55$ & $2.78$ & $2.79$ & $2.67$ & $2.42$ & $2.15$ & & & & $0.77$ \\
        \bottomrule
    \end{tabular}
\end{table}

\section{Slab under a convective boundary}
\label{app:convective}

On the slab of Section~\ref{subsec:slab}, under the film coefficient of
Table~\ref{tab:mat-quench}, the cooled faces exchange heat with the bath by the convective condition of Eq.~\eqref{eq:heat-strong}, a natural boundary condition of the functional of Eq.~\eqref{eq:piT}, and the problems of
Section~\ref{subsec:slab} are otherwise unchanged. Table~\ref{tab:convective}
lists the counts. Film coefficients for alumina quenched into
water peak at $15$ to $60~\mathrm{kW\,m^{-2}\,K^{-1}}$~\cite{becher1981,kim1991,zhou2012},
and the coefficient here is held constant at a value in that range.
At $380$~K the finite element computation forms its first cracks at load step $44$, against step $11$ under the prescribed boundary temperature. At every load it forms a sparse array of nearly equal cracks, unchanged at
$380$~K when the cap of its staggered loop is raised from $50$ to
$1000$. Repeated at $380$~K with film coefficients of $200$, $1000$ and
$5000~\mathrm{kW\,m^{-2}\,K^{-1}}$, it forms an array that approaches
that of the prescribed boundary temperature as the film coefficient grows.

\begin{table}[htbp]
    \centering
    \small
    \caption{Quenched slab under the convective boundary condition,
    number of cracks along the cooled long face that reach the depth
    $a$, on the realizations of Table~\ref{tab:slab-counts}. The film
    coefficient is that of Table~\ref{tab:mat-quench} unless stated,
    and the last row of the sweep is the prescribed boundary temperature.}
    \label{tab:convective}
    \begin{tabular}{@{}llcccc@{}}
        \toprule
        & & \multicolumn{4}{c}{$a$ (mm)} \\
        \cmidrule(l){3-6}
        Computation & $\dT$ (K) & $0.1$ & $0.5$ & $1.0$ & $2.0$ \\
        \midrule
        finite element & $280$ & $3$ & $3$ & $3$ & $3$ \\
                       & $380$ & $5$ & $5$ & $5$ & $4$ \\
                       & $480$ & $6$ & $6$ & $6$ & $6$ \\
                       & $580$ & $9$ & $9$ & $9$ & $6$ \\
        \addlinespace
        proposed method & $280$ & $16$ & $16$ & $11$ & $5$ \\
                        & $380$ & $17$ & $13$ & $11$ & $7$ \\
                        & $480$ & $20$ & $14$ & $11$ & $7$ \\
                        & $580$ & $27$ & $17$ & $13$ & $9$ \\
        \addlinespace
        finite element, $\hconv = 200~\mathrm{kW\,m^{-2}\,K^{-1}}$  & $380$ & $12$ & $12$ & $11$ & $5$ \\
        finite element, $\hconv = 1000~\mathrm{kW\,m^{-2}\,K^{-1}}$ & $380$ & $27$ & $16$ & $13$ & $7$ \\
        finite element, $\hconv = 5000~\mathrm{kW\,m^{-2}\,K^{-1}}$ & $380$ & $35$ & $18$ & $10$ & $7$ \\
        finite element, prescribed boundary temperature                      & $380$ & $38$ & $17$ & $9$ & $7$ \\
        \bottomrule
    \end{tabular}
\end{table}

Under the convective condition both arrays are sparser near the surface, since the layer in tension is deeper when it
cracks. At $380$~K, when the cracks appear, the stress of a one-dimensional cooling falls to half its surface value at the depth of $0.05$~mm under the prescribed boundary temperature and at $0.29$~mm under the convective condition. The deeper layer releases its energy through fewer
and longer cracks.

At $380$~K the proposed method forms $17$ nuclei in load step $47$, in which its damage band localizes, and from $16$ to $20$ when the computation is restarted at load step $44$ with $2$ or $4$ times the iterations or other optimizer settings. Started from the state of the proposed method at load step $44$, before this localization, the finite element computation forms its own $5$ cracks, while started from a state with $16$ nuclei it retains them. The two solvers therefore differ at nucleation, and irreversibility then preserves the nuclei. With its bound applied on the crack set only, the finite
element computation still forms $5$ cracks near the surface, so within the variations tried none of the budget, the optimizer settings or the treatment of irreversibility removes the difference.

%% file: main.bbl
\begin{thebibliography}{134}
\providecommand{\natexlab}[1]{#1}
\providecommand{\url}[1]{\texttt{#1}}
\expandafter\ifx\csname urlstyle\endcsname\relax
  \providecommand{\doi}[1]{doi: #1}\else
  \providecommand{\doi}{doi: \begingroup \urlstyle{rm}\Url}\fi

\bibitem[Bahr et~al.(1986)Bahr, Fischer, and Weiss]{bahr1986}
H.-A. Bahr, G.~Fischer, and H.-J. Weiss.
\newblock Thermal-shock crack patterns explained by single and multiple crack
  propagation.
\newblock \emph{Journal of Materials Science}, 21\penalty0 (8):\penalty0
  2716--2720, 1986.
\newblock \doi{10.1007/BF00551478}.

\bibitem[Shao et~al.(2010)Shao, Xu, Meng, Bai, Jiang, and Song]{shao2010}
Y.~Shao, X.~Xu, S.~Meng, G.~Bai, C.~Jiang, and F.~Song.
\newblock Crack patterns in ceramic plates after quenching.
\newblock \emph{Journal of the American Ceramic Society}, 93\penalty0
  (10):\penalty0 3006--3008, 2010.
\newblock \doi{10.1111/j.1551-2916.2010.03971.x}.

\bibitem[Jiang et~al.(2012)Jiang, Wu, Li, Song, Shao, Xu, and Yan]{jiang2012}
C.~P. Jiang, X.~F. Wu, J.~Li, F.~Song, Y.~F. Shao, X.~H. Xu, and P.~Yan.
\newblock A study of the mechanism of formation and numerical simulations of
  crack patterns in ceramics subjected to thermal shock.
\newblock \emph{Acta Materialia}, 60\penalty0 (11):\penalty0 4540--4550, 2012.
\newblock \doi{10.1016/j.actamat.2012.05.020}.

\bibitem[Yuse and Sano(1993)]{yuse1993}
A.~Yuse and M.~Sano.
\newblock Transition between crack patterns in quenched glass plates.
\newblock \emph{Nature}, 362\penalty0 (6418):\penalty0 329--331, 1993.
\newblock \doi{10.1038/362329a0}.

\bibitem[Yuse and Sano(1997)]{yuse1997}
A.~Yuse and M.~Sano.
\newblock Instabilities of quasi-static crack patterns in quenched glass
  plates.
\newblock \emph{Physica D: Nonlinear Phenomena}, 108\penalty0 (4):\penalty0
  365--378, 1997.
\newblock \doi{10.1016/S0167-2789(97)00011-0}.

\bibitem[Goehring et~al.(2009)Goehring, Mahadevan, and Morris]{goehring2009}
L.~Goehring, L.~Mahadevan, and S.~W. Morris.
\newblock Nonequilibrium scale selection mechanism for columnar jointing.
\newblock \emph{Proceedings of the National Academy of Sciences}, 106\penalty0
  (2):\penalty0 387--392, 2009.
\newblock \doi{10.1073/pnas.0805132106}.

\bibitem[Jagla(2002)]{jagla2002}
E.~A. Jagla.
\newblock Stable propagation of an ordered array of cracks during directional
  drying.
\newblock \emph{Physical Review E}, 65\penalty0 (4):\penalty0 046147, 2002.
\newblock \doi{10.1103/PhysRevE.65.046147}.

\bibitem[Bahr et~al.(2010)Bahr, Weiss, Bahr, Hofmann, Fischer, Lampenscherf,
  and Balke]{bahr2010}
H.-A. Bahr, H.-J. Weiss, U.~Bahr, M.~Hofmann, G.~Fischer, S.~Lampenscherf, and
  H.~Balke.
\newblock Scaling behavior of thermal shock crack patterns and tunneling cracks
  driven by cooling or drying.
\newblock \emph{Journal of the Mechanics and Physics of Solids}, 58\penalty0
  (9):\penalty0 1411--1421, 2010.
\newblock \doi{10.1016/j.jmps.2010.05.005}.

\bibitem[Sicsic et~al.(2014)Sicsic, Marigo, and Maurini]{sicsic2014}
P.~Sicsic, J.-J. Marigo, and C.~Maurini.
\newblock Initiation of a periodic array of cracks in the thermal shock
  problem: a gradient damage modeling.
\newblock \emph{Journal of the Mechanics and Physics of Solids}, 63:\penalty0
  256--284, 2014.
\newblock \doi{10.1016/j.jmps.2013.09.003}.

\bibitem[Lu and Fleck(1998)]{lu1998}
T.~J. Lu and N.~A. Fleck.
\newblock The thermal shock resistance of solids.
\newblock \emph{Acta Materialia}, 46\penalty0 (13):\penalty0 4755--4768, 1998.
\newblock \doi{10.1016/S1359-6454(98)00127-X}.

\bibitem[Zeng and Dolbow(2026)]{zeng2026}
B.~Zeng and J.~E. Dolbow.
\newblock A complete phase-field fracture model for brittle materials subjected
  to thermal shocks.
\newblock arXiv preprint arXiv:2602.09031v2, 2026.

\bibitem[Hasselman(1969)]{hasselman1969}
D.~P.~H. Hasselman.
\newblock Unified theory of thermal shock fracture initiation and crack
  propagation in brittle ceramics.
\newblock \emph{Journal of the American Ceramic Society}, 52\penalty0
  (11):\penalty0 600--604, 1969.
\newblock \doi{10.1111/j.1151-2916.1969.tb15848.x}.

\bibitem[Nemat-Nasser et~al.(1978)Nemat-Nasser, Keer, and
  Parihar]{nematnasser1978}
S.~Nemat-Nasser, L.~M. Keer, and K.~S. Parihar.
\newblock Unstable growth of thermally induced interacting cracks in brittle
  solids.
\newblock \emph{International Journal of Solids and Structures}, 14\penalty0
  (6):\penalty0 409--430, 1978.
\newblock \doi{10.1016/0020-7683(78)90007-0}.

\bibitem[Ba{\v{z}}ant et~al.(1979)Ba{\v{z}}ant, Ohtsubo, and Aoh]{bazant1979}
Z.~P. Ba{\v{z}}ant, H.~Ohtsubo, and K.~Aoh.
\newblock Stability and post-critical growth of a system of cooling or
  shrinkage cracks.
\newblock \emph{International Journal of Fracture}, 15\penalty0 (5):\penalty0
  443--456, 1979.
\newblock \doi{10.1007/BF00023331}.

\bibitem[Geyer and Nemat-Nasser(1982)]{geyer1982}
J.~F. Geyer and S.~Nemat-Nasser.
\newblock Experimental investigation of thermally induced interacting cracks in
  brittle solids.
\newblock \emph{International Journal of Solids and Structures}, 18\penalty0
  (4):\penalty0 349--356, 1982.
\newblock \doi{10.1016/0020-7683(82)90059-2}.

\bibitem[Griffith(1921)]{griffith1921}
A.~A. Griffith.
\newblock The phenomena of rupture and flow in solids.
\newblock \emph{Philosophical Transactions of the Royal Society of London A},
  221:\penalty0 163--198, 1921.
\newblock \doi{10.1098/rsta.1921.0006}.

\bibitem[Irwin(1957)]{irwin1957}
G.~R. Irwin.
\newblock Analysis of stresses and strains near the end of a crack traversing a
  plate.
\newblock \emph{Journal of Applied Mechanics}, 24\penalty0 (3):\penalty0
  361--364, 1957.
\newblock \doi{10.1115/1.4011547}.

\bibitem[Bittencourt et~al.(1996)Bittencourt, Wawrzynek, Ingraffea, and
  Sousa]{bittencourt1996}
T.~N. Bittencourt, P.~A. Wawrzynek, A.~R. Ingraffea, and J.~L. Sousa.
\newblock Quasi-automatic simulation of crack propagation for {2D} {LEFM}
  problems.
\newblock \emph{Engineering Fracture Mechanics}, 55\penalty0 (2):\penalty0
  321--334, 1996.
\newblock \doi{10.1016/0013-7944(95)00247-2}.

\bibitem[Mo{\"e}s et~al.(1999)Mo{\"e}s, Dolbow, and Belytschko]{moes1999}
N.~Mo{\"e}s, J.~Dolbow, and T.~Belytschko.
\newblock A finite element method for crack growth without remeshing.
\newblock \emph{International Journal for Numerical Methods in Engineering},
  46\penalty0 (1):\penalty0 131--150, 1999.
\newblock
  \doi{10.1002/(SICI)1097-0207(19990910)46:1<131::AID-NME726>3.0.CO;2-J}.

\bibitem[Belytschko et~al.(1994)Belytschko, Lu, and Gu]{belytschko1994}
T.~Belytschko, Y.~Y. Lu, and L.~Gu.
\newblock Element-free {Galerkin} methods.
\newblock \emph{International Journal for Numerical Methods in Engineering},
  37\penalty0 (2):\penalty0 229--256, 1994.
\newblock \doi{10.1002/nme.1620370205}.

\bibitem[Rabczuk and Belytschko(2004)]{rabczuk2004}
T.~Rabczuk and T.~Belytschko.
\newblock Cracking particles: a simplified meshfree method for arbitrary
  evolving cracks.
\newblock \emph{International Journal for Numerical Methods in Engineering},
  61\penalty0 (13):\penalty0 2316--2343, 2004.
\newblock \doi{10.1002/nme.1151}.

\bibitem[Duflot(2008)]{duflot2008}
M.~Duflot.
\newblock The extended finite element method in thermoelastic fracture
  mechanics.
\newblock \emph{International Journal for Numerical Methods in Engineering},
  74\penalty0 (5):\penalty0 827--847, 2008.
\newblock \doi{10.1002/nme.2197}.

\bibitem[Menouillard and Belytschko(2011)]{menouillard2011}
T.~Menouillard and T.~Belytschko.
\newblock Analysis and computations of oscillating crack propagation in a
  heated strip.
\newblock \emph{International Journal of Fracture}, 167\penalty0 (1):\penalty0
  57--70, 2011.
\newblock \doi{10.1007/s10704-010-9519-0}.

\bibitem[Chiaramonte et~al.(2020)Chiaramonte, Grossman-Ponemon, Keer, and
  Lew]{chiaramonte2020}
M.~M. Chiaramonte, B.~E. Grossman-Ponemon, L.~M. Keer, and A.~J. Lew.
\newblock Numerical analyses of crack path instabilities in quenched plates.
\newblock \emph{Extreme Mechanics Letters}, 40:\penalty0 100878, 2020.
\newblock \doi{10.1016/j.eml.2020.100878}.

\bibitem[Xu and Needleman(1994)]{xu1994}
X.-P. Xu and A.~Needleman.
\newblock Numerical simulations of fast crack growth in brittle solids.
\newblock \emph{Journal of the Mechanics and Physics of Solids}, 42\penalty0
  (9):\penalty0 1397--1434, 1994.
\newblock \doi{10.1016/0022-5096(94)90003-5}.

\bibitem[Kilic and Madenci(2009)]{kilic2009}
B.~Kilic and E.~Madenci.
\newblock Prediction of crack paths in a quenched glass plate by using
  peridynamic theory.
\newblock \emph{International Journal of Fracture}, 156\penalty0 (2):\penalty0
  165--177, 2009.
\newblock \doi{10.1007/s10704-009-9355-2}.

\bibitem[Xu et~al.(2018)Xu, Zhang, Chen, and Bobaru]{xu2018peri}
Z.~Xu, G.~Zhang, Z.~Chen, and F.~Bobaru.
\newblock Elastic vortices and thermally-driven cracks in brittle materials
  with peridynamics.
\newblock \emph{International Journal of Fracture}, 209\penalty0
  (1--2):\penalty0 203--222, 2018.
\newblock \doi{10.1007/s10704-017-0256-5}.

\bibitem[Li et~al.(2025)Li, Zhang, Wang, Gu, Zhang, and Shao]{li2025pddl}
H.~Li, Z.~Zhang, L.~Wang, X.~Gu, Y.~Zhang, and X.~Shao.
\newblock Integration of peridynamics and deep learning for efficient and
  accurate thermomechanical modeling.
\newblock \emph{Applied Sciences}, 15\penalty0 (18):\penalty0 10032, 2025.
\newblock \doi{10.3390/app151810032}.

\bibitem[Hayakawa(1994)]{hayakawa1994}
Y.~Hayakawa.
\newblock Numerical study of oscillatory crack propagation through a
  two-dimensional crystal.
\newblock \emph{Physical Review E}, 49\penalty0 (3):\penalty0 R1804--R1807,
  1994.
\newblock \doi{10.1103/PhysRevE.49.R1804}.

\bibitem[Francfort and Marigo(1998)]{francfort1998}
G.~A. Francfort and J.-J. Marigo.
\newblock Revisiting brittle fracture as an energy minimization problem.
\newblock \emph{Journal of the Mechanics and Physics of Solids}, 46\penalty0
  (8):\penalty0 1319--1342, 1998.
\newblock \doi{10.1016/S0022-5096(98)00034-9}.

\bibitem[Bourdin et~al.(2000)Bourdin, Francfort, and Marigo]{bourdin2000}
B.~Bourdin, G.~A. Francfort, and J.-J. Marigo.
\newblock Numerical experiments in revisited brittle fracture.
\newblock \emph{Journal of the Mechanics and Physics of Solids}, 48\penalty0
  (4):\penalty0 797--826, 2000.
\newblock \doi{10.1016/S0022-5096(99)00028-9}.

\bibitem[Bourdin et~al.(2008)Bourdin, Francfort, and Marigo]{bourdin2008}
B.~Bourdin, G.~A. Francfort, and J.-J. Marigo.
\newblock The variational approach to fracture.
\newblock \emph{Journal of Elasticity}, 91\penalty0 (1--3):\penalty0 5--148,
  2008.
\newblock \doi{10.1007/s10659-007-9107-3}.

\bibitem[Ambrosio and Tortorelli(1990)]{ambrosio1990}
L.~Ambrosio and V.~M. Tortorelli.
\newblock Approximation of functionals depending on jumps by elliptic
  functionals via {$\Gamma$}-convergence.
\newblock \emph{Communications on Pure and Applied Mathematics}, 43\penalty0
  (8):\penalty0 999--1036, 1990.
\newblock \doi{10.1002/cpa.3160430805}.

\bibitem[Miehe et~al.(2010{\natexlab{a}})Miehe, Hofacker, and
  Welschinger]{miehe2010cmame}
C.~Miehe, M.~Hofacker, and F.~Welschinger.
\newblock A phase field model for rate-independent crack propagation: robust
  algorithmic implementation based on operator splits.
\newblock \emph{Computer Methods in Applied Mechanics and Engineering},
  199\penalty0 (45--48):\penalty0 2765--2778, 2010{\natexlab{a}}.
\newblock \doi{10.1016/j.cma.2010.04.011}.

\bibitem[Amor et~al.(2009)Amor, Marigo, and Maurini]{amor2009}
H.~Amor, J.-J. Marigo, and C.~Maurini.
\newblock Regularized formulation of the variational brittle fracture with
  unilateral contact: numerical experiments.
\newblock \emph{Journal of the Mechanics and Physics of Solids}, 57\penalty0
  (8):\penalty0 1209--1229, 2009.
\newblock \doi{10.1016/j.jmps.2009.04.011}.

\bibitem[Ambati et~al.(2015)Ambati, Gerasimov, and De~Lorenzis]{ambati2015}
M.~Ambati, T.~Gerasimov, and L.~De~Lorenzis.
\newblock A review on phase-field models of brittle fracture and a new fast
  hybrid formulation.
\newblock \emph{Computational Mechanics}, 55\penalty0 (2):\penalty0 383--405,
  2015.
\newblock \doi{10.1007/s00466-014-1109-y}.

\bibitem[Pham et~al.(2011)Pham, Amor, Marigo, and Maurini]{pham2011}
K.~Pham, H.~Amor, J.-J. Marigo, and C.~Maurini.
\newblock Gradient damage models and their use to approximate brittle fracture.
\newblock \emph{International Journal of Damage Mechanics}, 20\penalty0
  (4):\penalty0 618--652, 2011.
\newblock \doi{10.1177/1056789510386852}.

\bibitem[Wu(2017)]{wu2017}
J.-Y. Wu.
\newblock A unified phase-field theory for the mechanics of damage and
  quasi-brittle failure.
\newblock \emph{Journal of the Mechanics and Physics of Solids}, 103:\penalty0
  72--99, 2017.
\newblock \doi{10.1016/j.jmps.2017.03.015}.

\bibitem[Tann{\'e} et~al.(2018)Tann{\'e}, Li, Bourdin, Marigo, and
  Maurini]{tanne2018}
E.~Tann{\'e}, T.~Li, B.~Bourdin, J.-J. Marigo, and C.~Maurini.
\newblock Crack nucleation in variational phase-field models of brittle
  fracture.
\newblock \emph{Journal of the Mechanics and Physics of Solids}, 110:\penalty0
  80--99, 2018.
\newblock \doi{10.1016/j.jmps.2017.09.006}.

\bibitem[Kumar et~al.(2020)Kumar, Bourdin, Francfort, and
  Lopez-Pamies]{kumar2020}
A.~Kumar, B.~Bourdin, G.~A. Francfort, and O.~Lopez-Pamies.
\newblock Revisiting nucleation in the phase-field approach to brittle
  fracture.
\newblock \emph{Journal of the Mechanics and Physics of Solids}, 142:\penalty0
  104027, 2020.
\newblock \doi{10.1016/j.jmps.2020.104027}.

\bibitem[Borden et~al.(2014)Borden, Hughes, Landis, and Verhoosel]{borden2014}
M.~J. Borden, T.~J.~R. Hughes, C.~M. Landis, and C.~V. Verhoosel.
\newblock A higher-order phase-field model for brittle fracture: formulation
  and analysis within the isogeometric analysis framework.
\newblock \emph{Computer Methods in Applied Mechanics and Engineering},
  273:\penalty0 100--118, 2014.
\newblock \doi{10.1016/j.cma.2014.01.016}.

\bibitem[Gerasimov and De~Lorenzis(2019)]{gerasimov2019}
T.~Gerasimov and L.~De~Lorenzis.
\newblock On penalization in variational phase-field models of brittle
  fracture.
\newblock \emph{Computer Methods in Applied Mechanics and Engineering},
  354:\penalty0 990--1026, 2019.
\newblock \doi{10.1016/j.cma.2019.05.038}.

\bibitem[Wu et~al.(2020)Wu, Nguyen, Nguyen, Sutula, Sinaie, and Bordas]{wu2020}
J.-Y. Wu, V.~P. Nguyen, C.~T. Nguyen, D.~Sutula, S.~Sinaie, and S.~P.~A.
  Bordas.
\newblock Phase-field modeling of fracture.
\newblock \emph{Advances in Applied Mechanics}, 53:\penalty0 1--183, 2020.
\newblock \doi{10.1016/bs.aams.2019.08.001}.

\bibitem[Bourdin et~al.(2014)Bourdin, Marigo, Maurini, and Sicsic]{bourdin2014}
B.~Bourdin, J.-J. Marigo, C.~Maurini, and P.~Sicsic.
\newblock Morphogenesis and propagation of complex cracks induced by thermal
  shocks.
\newblock \emph{Physical Review Letters}, 112\penalty0 (1):\penalty0 014301,
  2014.
\newblock \doi{10.1103/PhysRevLett.112.014301}.

\bibitem[Shao et~al.(2011)Shao, Zhang, Xu, Zhou, Li, and Liu]{shao2011}
Y.~Shao, Y.~Zhang, X.~Xu, Z.~Zhou, W.~Li, and B.~Liu.
\newblock Effect of crack pattern on the residual strength of ceramics after
  quenching.
\newblock \emph{Journal of the American Ceramic Society}, 94\penalty0
  (9):\penalty0 2804--2807, 2011.
\newblock \doi{10.1111/j.1551-2916.2011.04728.x}.

\bibitem[Lopez-Pamies et~al.(2025)Lopez-Pamies, Dolbow, Francfort, and
  Larsen]{lopezpamies2025}
O.~Lopez-Pamies, J.~E. Dolbow, G.~A. Francfort, and C.~J. Larsen.
\newblock Classical variational phase-field models cannot predict fracture
  nucleation.
\newblock \emph{Computer Methods in Applied Mechanics and Engineering},
  433:\penalty0 117520, 2025.
\newblock \doi{10.1016/j.cma.2024.117520}.

\bibitem[Kamarei et~al.(2026)Kamarei, Zeng, Dolbow, and
  Lopez-Pamies]{kamarei2026}
F.~Kamarei, B.~Zeng, J.~E. Dolbow, and O.~Lopez-Pamies.
\newblock Nine circles of elastic brittle fracture: a series of challenge
  problems to assess fracture models.
\newblock \emph{Computer Methods in Applied Mechanics and Engineering},
  448:\penalty0 118449, 2026.
\newblock \doi{10.1016/j.cma.2025.118449}.

\bibitem[Zeng et~al.(2025)Zeng, Guilleminot, and Dolbow]{zeng2025nucleation}
B.~Zeng, J.~Guilleminot, and J.~E. Dolbow.
\newblock Examining crack nucleation under spatially uniform stress states with
  a complete phase-field model for fracture.
\newblock \emph{Theoretical and Applied Fracture Mechanics}, 140:\penalty0
  105170, 2025.
\newblock \doi{10.1016/j.tafmec.2025.105170}.

\bibitem[Ronsin et~al.(1995)Ronsin, Heslot, and Perrin]{ronsin1995}
O.~Ronsin, F.~Heslot, and B.~Perrin.
\newblock Experimental study of quasistatic brittle crack propagation.
\newblock \emph{Physical Review Letters}, 75\penalty0 (12):\penalty0
  2352--2355, 1995.
\newblock \doi{10.1103/PhysRevLett.75.2352}.

\bibitem[Ronsin and Perrin(1998)]{ronsin1998}
O.~Ronsin and B.~Perrin.
\newblock Dynamics of quasistatic directional crack growth.
\newblock \emph{Physical Review E}, 58\penalty0 (6):\penalty0 7878--7886, 1998.
\newblock \doi{10.1103/PhysRevE.58.7878}.

\bibitem[Yang and Ravi-Chandar(2001)]{yang2001}
B.~Yang and K.~Ravi-Chandar.
\newblock Crack path instabilities in a quenched glass plate.
\newblock \emph{Journal of the Mechanics and Physics of Solids}, 49\penalty0
  (1):\penalty0 91--130, 2001.
\newblock \doi{10.1016/S0022-5096(00)00022-3}.

\bibitem[Marder(1994)]{marder1994}
M.~Marder.
\newblock Instability of a crack in a heated strip.
\newblock \emph{Physical Review E}, 49\penalty0 (1):\penalty0 R51--R54, 1994.
\newblock \doi{10.1103/PhysRevE.49.R51}.

\bibitem[Sasa et~al.(1994)Sasa, Sekimoto, and Nakanishi]{sasa1994}
S.-i. Sasa, K.~Sekimoto, and H.~Nakanishi.
\newblock Oscillatory instability of crack propagations in quasistatic
  fracture.
\newblock \emph{Physical Review E}, 50\penalty0 (3):\penalty0 R1733--R1736,
  1994.
\newblock \doi{10.1103/PhysRevE.50.R1733}.

\bibitem[Adda-Bedia and Pomeau(1995)]{adda1995}
M.~Adda-Bedia and Y.~Pomeau.
\newblock Crack instabilities of a heated glass strip.
\newblock \emph{Physical Review E}, 52\penalty0 (4):\penalty0 4105--4113, 1995.
\newblock \doi{10.1103/PhysRevE.52.4105}.

\bibitem[Corson et~al.(2009)Corson, Adda-Bedia, Henry, and Katzav]{corson2009}
F.~Corson, M.~Adda-Bedia, H.~Henry, and E.~Katzav.
\newblock Thermal fracture as a framework for quasi-static crack propagation.
\newblock \emph{International Journal of Fracture}, 158\penalty0 (1):\penalty0
  1--14, 2009.
\newblock \doi{10.1007/s10704-009-9361-4}.

\bibitem[Bouchbinder et~al.(2003)Bouchbinder, Hentschel, and
  Procaccia]{bouchbinder2003}
E.~Bouchbinder, H.~G.~E. Hentschel, and I.~Procaccia.
\newblock Dynamical instabilities of quasistatic crack propagation under
  thermal stress.
\newblock \emph{Physical Review E}, 68\penalty0 (3):\penalty0 036601, 2003.
\newblock \doi{10.1103/PhysRevE.68.036601}.

\bibitem[Bahr et~al.(1995)Bahr, Gerbatsch, Bahr, and Weiss]{bahr1995}
H.-A. Bahr, A.~Gerbatsch, U.~Bahr, and H.-J. Weiss.
\newblock Oscillatory instability in thermal cracking: a first-order
  phase-transition phenomenon.
\newblock \emph{Physical Review E}, 52\penalty0 (1):\penalty0 240--243, 1995.
\newblock \doi{10.1103/PhysRevE.52.240}.

\bibitem[Pan et~al.(2024)Pan, Abdelmoula, Li, and Cheng]{pan2024}
W.~Pan, R.~Abdelmoula, J.~Li, and C.~Cheng.
\newblock Analysis of crack path instabilities in a quenched glass plate using
  the phase-field cohesive zone model.
\newblock \emph{Journal of Mechanics of Materials and Structures}, 19\penalty0
  (2):\penalty0 235--250, 2024.
\newblock \doi{10.2140/jomms.2024.19.235}.

\bibitem[Miehe et~al.(2015)Miehe, Sch{\"a}nzel, and Ulmer]{miehe2015}
C.~Miehe, L.-M. Sch{\"a}nzel, and H.~Ulmer.
\newblock Phase field modeling of fracture in multi-physics problems. {Part}
  {I}. {Balance} of crack surface and failure criteria for brittle crack
  propagation in thermo-elastic solids.
\newblock \emph{Computer Methods in Applied Mechanics and Engineering},
  294:\penalty0 449--485, 2015.
\newblock \doi{10.1016/j.cma.2014.11.016}.

\bibitem[Chu et~al.(2017)Chu, Li, and Liu]{chu2017}
D.~Chu, X.~Li, and Z.~Liu.
\newblock Study the dynamic crack path in brittle material under thermal shock
  loading by phase field modeling.
\newblock \emph{International Journal of Fracture}, 208\penalty0
  (1--2):\penalty0 115--130, 2017.
\newblock \doi{10.1007/s10704-017-0220-4}.

\bibitem[Wang et~al.(2020)Wang, Ye, Liu, Liu, Chu, and Zhuang]{wang2020}
T.~Wang, X.~Ye, Z.~Liu, X.~Liu, D.~Chu, and Z.~Zhuang.
\newblock A phase-field model of thermo-elastic coupled brittle fracture with
  explicit time integration.
\newblock \emph{Computational Mechanics}, 65\penalty0 (5):\penalty0 1305--1321,
  2020.
\newblock \doi{10.1007/s00466-020-01820-6}.

\bibitem[Mandal et~al.(2021)Mandal, Nguyen, Wu, Nguyen-Thanh, and
  de~Vaucorbeil]{mandal2021fracture}
T.~K. Mandal, V.~P. Nguyen, J.-Y. Wu, C.~Nguyen-Thanh, and A.~de~Vaucorbeil.
\newblock Fracture of thermo-elastic solids: phase-field modeling and new
  results with an efficient monolithic solver.
\newblock \emph{Computer Methods in Applied Mechanics and Engineering},
  376:\penalty0 113648, 2021.
\newblock \doi{10.1016/j.cma.2020.113648}.

\bibitem[Tangella et~al.(2022{\natexlab{a}})Tangella, Kumbhar, and
  Annabattula]{tangella2022}
R.~G. Tangella, P.~Kumbhar, and R.~K. Annabattula.
\newblock Hybrid phase-field modeling of thermo-elastic crack propagation.
\newblock \emph{International Journal for Computational Methods in Engineering
  Science and Mechanics}, 23\penalty0 (1):\penalty0 29--44, 2022{\natexlab{a}}.
\newblock \doi{10.1080/15502287.2021.1904462}.

\bibitem[Ruan et~al.(2023)Ruan, Rezaei, Yang, Gross, and
  Xu]{ruan2023thermomechanical}
H.~Ruan, S.~Rezaei, Y.~Yang, D.~Gross, and B.-X. Xu.
\newblock A thermo-mechanical phase-field fracture model: application to hot
  cracking simulations in additive manufacturing.
\newblock \emph{Journal of the Mechanics and Physics of Solids}, 172:\penalty0
  105169, 2023.
\newblock \doi{10.1016/j.jmps.2022.105169}.

\bibitem[Li et~al.(2022)Li, Li, Li, Li, and Zhou]{li2022thermal}
D.~Li, P.~Li, W.~Li, W.~Li, and K.~Zhou.
\newblock Three-dimensional phase-field modeling of temperature-dependent
  thermal shock-induced fracture in ceramic materials.
\newblock \emph{Engineering Fracture Mechanics}, 268:\penalty0 108444, 2022.
\newblock \doi{10.1016/j.engfracmech.2022.108444}.

\bibitem[Wang et~al.(2024)Wang, Zhang, Han, Wang, Ye, and
  Zhuang]{wang2024thermal}
T.~Wang, Y.~Zhang, H.~Han, L.~Wang, X.~Ye, and Z.~Zhuang.
\newblock Phase field modeling of crack propagation in three-dimensional
  quasi-brittle materials under thermal shock.
\newblock \emph{Engineering Fracture Mechanics}, 302:\penalty0 110070, 2024.
\newblock \doi{10.1016/j.engfracmech.2024.110070}.

\bibitem[Badnava et~al.(2018)Badnava, Msekh, Etemadi, and Rabczuk]{badnava2018}
H.~Badnava, M.~A. Msekh, E.~Etemadi, and T.~Rabczuk.
\newblock An h-adaptive thermo-mechanical phase field model for fracture.
\newblock \emph{Finite Elements in Analysis and Design}, 138:\penalty0 31--47,
  2018.
\newblock \doi{10.1016/j.finel.2017.09.003}.

\bibitem[Ferro et~al.(2018)Ferro, Micheletti, and Perotto]{ferro2018}
N.~Ferro, S.~Micheletti, and S.~Perotto.
\newblock Anisotropic mesh adaptation for crack propagation induced by a
  thermal shock in {2D}.
\newblock \emph{Computer Methods in Applied Mechanics and Engineering},
  331:\penalty0 138--158, 2018.
\newblock \doi{10.1016/j.cma.2017.11.024}.

\bibitem[Chen et~al.(2024)Chen, Natarajan, Ooi, and Song]{chen2024}
H.~Chen, S.~Natarajan, E.~T. Ooi, and C.~Song.
\newblock Modeling of coupled thermo-mechanical crack propagation in brittle
  solids using adaptive phase field method with scaled boundary finite element
  method.
\newblock \emph{Theoretical and Applied Fracture Mechanics}, 129:\penalty0
  104158, 2024.
\newblock \doi{10.1016/j.tafmec.2023.104158}.

\bibitem[Ali et~al.(2025)Ali, Vuga, Mavri\v{c}, and \v{S}arler]{ali2025}
I.~Ali, G.~Vuga, B.~Mavri\v{c}, and B.~\v{S}arler.
\newblock A fourth-order phase-field model of crack initiation and propagation
  under thermomechanical loading solved with strong-form meshless method.
\newblock \emph{Theoretical and Applied Fracture Mechanics}, 139:\penalty0
  105078, 2025.
\newblock \doi{10.1016/j.tafmec.2025.105078}.

\bibitem[Si et~al.(2023)Si, Yu, Li, and Natarajan]{si2023}
Z.~Si, T.~Yu, Y.~Li, and S.~Natarajan.
\newblock Adaptive multi-patch isogeometric phase-field method for quasi-static
  brittle fracture based on {Nitsche}'s method.
\newblock \emph{Computer Methods in Applied Mechanics and Engineering},
  414:\penalty0 116154, 2023.
\newblock \doi{10.1016/j.cma.2023.116154}.

\bibitem[Zhang et~al.(2026{\natexlab{a}})Zhang, Makki~Alamdari, Shahbodagh,
  Vahab, Anitescu, Rabczuk, and Atroshchenko]{zhang2026meshfree}
H.~Zhang, M.~Makki~Alamdari, B.~Shahbodagh, M.~Vahab, C.~Anitescu, T.~Rabczuk,
  and E.~Atroshchenko.
\newblock A mesh-free multiresolution deep energy method with phase-field
  modeling of brittle fracture.
\newblock arXiv preprint arXiv:2608.24126, 2026{\natexlab{a}}.

\bibitem[Lagaris et~al.(1998)Lagaris, Likas, and Fotiadis]{lagaris1998}
I.~E. Lagaris, A.~Likas, and D.~I. Fotiadis.
\newblock Artificial neural networks for solving ordinary and partial
  differential equations.
\newblock \emph{IEEE Transactions on Neural Networks}, 9\penalty0 (5):\penalty0
  987--1000, 1998.
\newblock \doi{10.1109/72.712178}.

\bibitem[Raissi et~al.(2019)Raissi, Perdikaris, and Karniadakis]{raissi2019}
M.~Raissi, P.~Perdikaris, and G.~E. Karniadakis.
\newblock Physics-informed neural networks: a deep learning framework for
  solving forward and inverse problems involving nonlinear partial differential
  equations.
\newblock \emph{Journal of Computational Physics}, 378:\penalty0 686--707,
  2019.
\newblock \doi{10.1016/j.jcp.2018.10.045}.

\bibitem[Karniadakis et~al.(2021)Karniadakis, Kevrekidis, Lu, Perdikaris, Wang,
  and Yang]{karniadakis2021}
G.~E. Karniadakis, I.~G. Kevrekidis, L.~Lu, P.~Perdikaris, S.~Wang, and
  L.~Yang.
\newblock Physics-informed machine learning.
\newblock \emph{Nature Reviews Physics}, 3\penalty0 (6):\penalty0 422--440,
  2021.
\newblock \doi{10.1038/s42254-021-00314-5}.

\bibitem[E and Yu(2018)]{e2018deepritz}
W.~E and B.~Yu.
\newblock The deep {Ritz} method: a deep learning-based numerical algorithm for
  solving variational problems.
\newblock \emph{Communications in Mathematics and Statistics}, 6\penalty0
  (1):\penalty0 1--12, 2018.
\newblock \doi{10.1007/s40304-018-0127-z}.

\bibitem[Samaniego et~al.(2020)Samaniego, Anitescu, Goswami, Nguyen-Thanh, Guo,
  Hamdia, Zhuang, and Rabczuk]{samaniego2020}
E.~Samaniego, C.~Anitescu, S.~Goswami, V.~M. Nguyen-Thanh, H.~Guo, K.~Hamdia,
  X.~Zhuang, and T.~Rabczuk.
\newblock An energy approach to the solution of partial differential equations
  in computational mechanics via machine learning: concepts, implementation and
  applications.
\newblock \emph{Computer Methods in Applied Mechanics and Engineering},
  362:\penalty0 112790, 2020.
\newblock \doi{10.1016/j.cma.2019.112790}.

\bibitem[Nguyen-Thanh et~al.(2020)Nguyen-Thanh, Zhuang, and
  Rabczuk]{nguyenthanh2020}
V.~M. Nguyen-Thanh, X.~Zhuang, and T.~Rabczuk.
\newblock A deep energy method for finite deformation hyperelasticity.
\newblock \emph{European Journal of Mechanics - A/Solids}, 80:\penalty0 103874,
  2020.
\newblock \doi{10.1016/j.euromechsol.2019.103874}.

\bibitem[Sukumar and Srivastava(2022)]{sukumar2022}
N.~Sukumar and A.~Srivastava.
\newblock Exact imposition of boundary conditions with distance functions in
  physics-informed deep neural networks.
\newblock \emph{Computer Methods in Applied Mechanics and Engineering},
  389:\penalty0 114333, 2022.
\newblock \doi{10.1016/j.cma.2021.114333}.

\bibitem[Wu et~al.(2023)Wu, Zhu, Tan, Kartha, and Lu]{wu2023}
C.~Wu, M.~Zhu, Q.~Tan, Y.~Kartha, and L.~Lu.
\newblock A comprehensive study of non-adaptive and residual-based adaptive
  sampling for physics-informed neural networks.
\newblock \emph{Computer Methods in Applied Mechanics and Engineering},
  403:\penalty0 115671, 2023.
\newblock \doi{10.1016/j.cma.2022.115671}.

\bibitem[Nabian et~al.(2021)Nabian, Gladstone, and Meidani]{nabian2021}
M.~A. Nabian, R.~J. Gladstone, and H.~Meidani.
\newblock Efficient training of physics-informed neural networks via importance
  sampling.
\newblock \emph{Computer-Aided Civil and Infrastructure Engineering},
  36\penalty0 (8):\penalty0 962--977, 2021.
\newblock \doi{10.1111/mice.12685}.

\bibitem[Tancik et~al.(2020)Tancik, Srinivasan, Mildenhall, Fridovich-Keil,
  Raghavan, Singhal, Ramamoorthi, Barron, and Ng]{tancik2020}
M.~Tancik, P.~P. Srinivasan, B.~Mildenhall, S.~Fridovich-Keil, N.~Raghavan,
  U.~Singhal, R.~Ramamoorthi, J.~T. Barron, and R.~Ng.
\newblock Fourier features let networks learn high frequency functions in low
  dimensional domains.
\newblock In \emph{Advances in Neural Information Processing Systems},
  volume~33, pages 7537--7547, 2020.

\bibitem[M{\"u}ller et~al.(2022)M{\"u}ller, Evans, Schied, and
  Keller]{mueller2022}
T.~M{\"u}ller, A.~Evans, C.~Schied, and A.~Keller.
\newblock Instant neural graphics primitives with a multiresolution hash
  encoding.
\newblock \emph{ACM Transactions on Graphics}, 41\penalty0 (4):\penalty0
  102:1--102:15, 2022.
\newblock \doi{10.1145/3528223.3530127}.

\bibitem[Huang and Alkhalifah(2024)]{huang2024hash}
X.~Huang and T.~Alkhalifah.
\newblock Efficient physics-informed neural networks using hash encoding.
\newblock \emph{Journal of Computational Physics}, 501:\penalty0 112760, 2024.
\newblock \doi{10.1016/j.jcp.2024.112760}.

\bibitem[Goswami et~al.(2020{\natexlab{a}})Goswami, Anitescu, Chakraborty, and
  Rabczuk]{goswami2020tl}
S.~Goswami, C.~Anitescu, S.~Chakraborty, and T.~Rabczuk.
\newblock Transfer learning enhanced physics informed neural network for
  phase-field modeling of fracture.
\newblock \emph{Theoretical and Applied Fracture Mechanics}, 106:\penalty0
  102447, 2020{\natexlab{a}}.
\newblock \doi{10.1016/j.tafmec.2019.102447}.

\bibitem[Goswami et~al.(2020{\natexlab{b}})Goswami, Anitescu, and
  Rabczuk]{goswami2020fourth}
S.~Goswami, C.~Anitescu, and T.~Rabczuk.
\newblock Adaptive fourth-order phase field analysis using deep energy
  minimization.
\newblock \emph{Theoretical and Applied Fracture Mechanics}, 107:\penalty0
  102527, 2020{\natexlab{b}}.
\newblock \doi{10.1016/j.tafmec.2020.102527}.

\bibitem[Manav et~al.(2024)Manav, Molinaro, Mishra, and De~Lorenzis]{manav2024}
M.~Manav, R.~Molinaro, S.~Mishra, and L.~De~Lorenzis.
\newblock Phase-field modeling of fracture with physics-informed deep learning.
\newblock \emph{Computer Methods in Applied Mechanics and Engineering},
  429:\penalty0 117104, 2024.
\newblock \doi{10.1016/j.cma.2024.117104}.

\bibitem[Chakraborty et~al.(2022)Chakraborty, Anitescu, Goswami, Zhuang, and
  Rabczuk]{chakraborty2022}
A.~Chakraborty, C.~Anitescu, S.~Goswami, X.~Zhuang, and T.~Rabczuk.
\newblock Variational energy based {XPINNs} for phase field analysis in brittle
  fracture.
\newblock arXiv preprint arXiv:2207.02307, 2022.

\bibitem[Ghaffari~Motlagh et~al.(2023)Ghaffari~Motlagh, Jimack, and
  de~Borst]{motlagh2023}
Y.~Ghaffari~Motlagh, P.~K. Jimack, and R.~de~Borst.
\newblock Deep learning phase-field model for brittle fractures.
\newblock \emph{International Journal for Numerical Methods in Engineering},
  124\penalty0 (3):\penalty0 620--638, 2023.
\newblock \doi{10.1002/nme.7135}.

\bibitem[Feng and Zhou(2025)]{feng2025gnn}
B.~Feng and X.-P. Zhou.
\newblock The novel physics-enhanced graph neural network for phase-field
  fracture modelling.
\newblock \emph{Computer Methods in Applied Mechanics and Engineering},
  446:\penalty0 118284, 2025.
\newblock \doi{10.1016/j.cma.2025.118284}.

\bibitem[Zhang et~al.(2026{\natexlab{b}})Zhang, Tang, and
  Xu]{zhang2026chebykan}
P.~Zhang, K.~Tang, and B.~Xu.
\newblock {ChebyKAN}-enhanced deep energy method for phase field fracture: from
  homogeneous to heterogeneous materials.
\newblock \emph{Journal of the Mechanics and Physics of Solids}, 214:\penalty0
  106693, 2026{\natexlab{b}}.
\newblock \doi{10.1016/j.jmps.2026.106693}.

\bibitem[Plung{\.e} et~al.(2026)Plung{\.e}, Brommer, Edwards, and
  Kakouris]{plunge2026}
N.~Plung{\.e}, P.~Brommer, R.~S. Edwards, and E.~G. Kakouris.
\newblock Deep learning-based phase-field modelling of brittle fracture in
  anisotropic media.
\newblock \emph{International Journal for Numerical Methods in Engineering},
  127\penalty0 (13):\penalty0 e70381, 2026.
\newblock \doi{10.1002/nme.70381}.

\bibitem[Dean and Bahtiri(2026)]{dean2026}
A.~Dean and B.~Bahtiri.
\newblock A fracture-informed multiphysics self-sensing framework for
  phase-field fracture in piezoresistive materials using the deep energy
  method.
\newblock \emph{Theoretical and Applied Fracture Mechanics}, 148:\penalty0
  105890, 2026.
\newblock \doi{10.1016/j.tafmec.2026.105890}.

\bibitem[Yu et~al.(2026)Yu, Long, Jiang, Wang, and Liu]{yu2026pinn}
M.~Yu, X.~Long, C.~Jiang, X.~Wang, and K.~Liu.
\newblock Physics-informed neural networks for inverse length scale
  identification and forward phase-field fracture modeling.
\newblock \emph{Engineering Fracture Mechanics}, 337:\penalty0 111978, 2026.
\newblock \doi{10.1016/j.engfracmech.2026.111978}.

\bibitem[Wang et~al.(2026)Wang, Lin, Goswami, Zhao, Zhang, Bai, Anitescu,
  Eshaghi, Zhuang, Rabczuk, and Liu]{wang2026xdem}
Y.~Wang, Y.~Lin, S.~Goswami, L.~Zhao, H.~Zhang, J.~Bai, C.~Anitescu, M.~S.
  Eshaghi, X.~Zhuang, T.~Rabczuk, and Y.~Liu.
\newblock Towards unified {AI}-driven fracture mechanics: the extended deep
  energy method ({XDEM}).
\newblock \emph{Nature Communications}, 17:\penalty0 8492, 2026.
\newblock \doi{10.1038/s41467-026-76748-1}.

\bibitem[Kiyani et~al.(2025)Kiyani, Manav, Kadivar, De~Lorenzis, and
  Karniadakis]{kiyani2025}
E.~Kiyani, M.~Manav, N.~Kadivar, L.~De~Lorenzis, and G.~E. Karniadakis.
\newblock Predicting crack nucleation and propagation in brittle materials
  using deep operator networks with diverse trunk architectures.
\newblock \emph{Computer Methods in Applied Mechanics and Engineering},
  441:\penalty0 117984, 2025.
\newblock \doi{10.1016/j.cma.2025.117984}.

\bibitem[Hamdi and Lejeune(2026)]{hamdi2026}
E.~Hamdi and E.~Lejeune.
\newblock Towards robust surrogate models: benchmarking machine learning
  approaches to expediting phase field simulations of brittle fracture.
\newblock \emph{Computer Methods in Applied Mechanics and Engineering},
  449:\penalty0 118526, 2026.
\newblock \doi{10.1016/j.cma.2025.118526}.

\bibitem[Ani et~al.(2026)Ani, Nakka, Subhash, Molinari, and Ponnusami]{ani2026}
A.~Ani, R.~Nakka, G.~Subhash, J.-F. Molinari, and S.~A. Ponnusami.
\newblock Machine learning for computational fracture and damage
  mechanics---status and perspectives.
\newblock \emph{Engineering Fracture Mechanics}, 332:\penalty0 111778, 2026.
\newblock \doi{10.1016/j.engfracmech.2025.111778}.

\bibitem[Zhou and Yu(2024)]{zhouyu2024transfer}
X.-P. Zhou and X.-L. Yu.
\newblock Transfer learning enhanced nonlocal energy-informed neural network
  for quasi-static fracture in rock-like materials.
\newblock \emph{Computer Methods in Applied Mechanics and Engineering},
  430:\penalty0 117226, 2024.
\newblock \doi{10.1016/j.cma.2024.117226}.

\bibitem[Zhou and Ma(2026)]{zhouma2026nonlocal}
X.~Zhou and J.~Ma.
\newblock Nonlocal energy-encoded physics-informed neural networks for crack
  propagation in solid materials.
\newblock \emph{Journal of the Mechanics and Physics of Solids}, 216:\penalty0
  106770, 2026.
\newblock \doi{10.1016/j.jmps.2026.106770}.

\bibitem[Cai et~al.(2021)Cai, Wang, Wang, Perdikaris, and Karniadakis]{cai2021}
S.~Cai, Z.~Wang, S.~Wang, P.~Perdikaris, and G.~E. Karniadakis.
\newblock Physics-informed neural networks for heat transfer problems.
\newblock \emph{Journal of Heat Transfer}, 143\penalty0 (6):\penalty0 060801,
  2021.
\newblock \doi{10.1115/1.4050542}.

\bibitem[Lin et~al.(2024)Lin, Wang, and Hu]{lin2024}
K.-C. Lin, K.-C. Wang, and C.-H. Hu.
\newblock Investigating deep energy method applications in thermoelasticity.
\newblock \emph{Engineering Analysis with Boundary Elements}, 159:\penalty0
  302--314, 2024.
\newblock \doi{10.1016/j.enganabound.2023.12.012}.

\bibitem[Mojahedin et~al.(2022)Mojahedin, Hashemitaheri, Salavati, Fu, and
  Rabczuk]{mojahedin2022}
A.~Mojahedin, M.~Hashemitaheri, M.~Salavati, X.~Fu, and T.~Rabczuk.
\newblock A deep energy method for the analysis of thermoporoelastic
  functionally graded beams.
\newblock \emph{International Journal of Computational Methods}, 19\penalty0
  (8):\penalty0 2143020, 2022.
\newblock \doi{10.1142/S0219876221430209}.

\bibitem[Raj et~al.(2021)Raj, Kumbhar, and Annabattula]{raj2021pinnfgm}
M.~Raj, P.~Kumbhar, and R.~K. Annabattula.
\newblock Physics-informed neural networks for solving thermo-mechanics
  problems of functionally graded material.
\newblock arXiv preprint arXiv:2111.10751, 2021.

\bibitem[Harandi et~al.(2024)Harandi, Moeineddin, Kaliske, Reese, and
  Rezaei]{harandi2024}
A.~Harandi, A.~Moeineddin, M.~Kaliske, S.~Reese, and S.~Rezaei.
\newblock Mixed formulation of physics-informed neural networks for
  thermo-mechanically coupled systems and heterogeneous domains.
\newblock \emph{International Journal for Numerical Methods in Engineering},
  125\penalty0 (4):\penalty0 e7388, 2024.
\newblock \doi{10.1002/nme.7388}.

\bibitem[Yu and Zhou(2023)]{yuzhou2023nonlocal}
X.-L. Yu and X.-P. Zhou.
\newblock A nonlocal energy-informed neural network for isotropic elastic
  solids with cracks under thermomechanical loads.
\newblock \emph{International Journal for Numerical Methods in Engineering},
  124\penalty0 (18):\penalty0 3935--3963, 2023.
\newblock \doi{10.1002/nme.7296}.

\bibitem[Wu et~al.(2026)Wu, Liu, Zhang, Liu, Sun, Wang, and Gong]{wu2026}
Y.~Wu, W.~Liu, Z.~Zhang, B.~Liu, H.~Sun, G.~Wang, and X.~Gong.
\newblock Identification of rock fracture parameters under non-uniform
  temperature fields based on physics-informed neural networks.
\newblock \emph{Acta Mechanica Sinica}, 42\penalty0 (9):\penalty0 425672, 2026.
\newblock \doi{10.1007/s10409-025-25672-x}.

\bibitem[Yadav et~al.(2026)Yadav, Yadav, and Patil]{yadav2026mpinn}
S.~K. Yadav, V.~Yadav, and R.~Patil.
\newblock Fracture analysis of functionally graded plates under
  thermo-mechanical loading using mesh enriched physics-informed neural
  networks ({M-PINNs}).
\newblock Research Square preprint, \doi{10.21203/rs.3.rs-9295110/v1}, 2026.

\bibitem[He et~al.(2026)He, Li, Natarajan, and Yu]{he2026drm}
J.-N. He, H.~Li, S.~Natarajan, and T.~Yu.
\newblock Thermo-mechanical phase-field fracture modeling via deep {Ritz}
  method.
\newblock \emph{Engineering Fracture Mechanics}, 346:\penalty0 112651, 2026.
\newblock \doi{10.1016/j.engfracmech.2026.112651}.

\bibitem[Kong(2026)]{kong2026thesis}
H.-e. Kong.
\newblock \emph{A variational multiscale method for material heterogeneity of
  solid mechanics and an investigation of machine learning framework for
  fracture modeling}.
\newblock {PhD} dissertation, The Pennsylvania State University, 2026.
\newblock URL \url{https://etda.libraries.psu.edu/catalog/20420hqk5365}.

\bibitem[Zhang et~al.(2026{\natexlab{c}})Zhang, Makki~Alamdari, Shahbodagh,
  Vahab, Anitescu, Rabczuk, and Atroshchenko]{zhang2026xdem}
H.~Zhang, M.~Makki~Alamdari, B.~Shahbodagh, M.~Vahab, C.~Anitescu, T.~Rabczuk,
  and E.~Atroshchenko.
\newblock An extended deep energy method for thermo-mechanical crack
  propagation.
\newblock arXiv preprint arXiv:2610.09433, 2026{\natexlab{c}}.

\bibitem[Tang et~al.(2016)Tang, Zhang, Tang, and Liu]{tang2016}
S.~B. Tang, H.~Zhang, C.~A. Tang, and H.~Y. Liu.
\newblock Numerical model for the cracking behavior of heterogeneous brittle
  solids subjected to thermal shock.
\newblock \emph{International Journal of Solids and Structures}, 80:\penalty0
  520--531, 2016.
\newblock \doi{10.1016/j.ijsolstr.2015.10.012}.

\bibitem[Miehe et~al.(2010{\natexlab{b}})Miehe, Welschinger, and
  Hofacker]{miehe2010}
C.~Miehe, F.~Welschinger, and M.~Hofacker.
\newblock Thermodynamically consistent phase-field models of fracture:
  variational principles and multi-field {FE} implementations.
\newblock \emph{International Journal for Numerical Methods in Engineering},
  83\penalty0 (10):\penalty0 1273--1311, 2010{\natexlab{b}}.
\newblock \doi{10.1002/nme.2861}.

\bibitem[Linse et~al.(2017)Linse, Hennig, K{\"a}stner, and de~Borst]{linse2017}
T.~Linse, P.~Hennig, M.~K{\"a}stner, and R.~de~Borst.
\newblock A convergence study of phase-field models for brittle fracture.
\newblock \emph{Engineering Fracture Mechanics}, 184:\penalty0 307--318, 2017.
\newblock \doi{10.1016/j.engfracmech.2017.09.013}.

\bibitem[Yang et~al.(2006)Yang, Stainier, and Ortiz]{yang2006}
Q.~Yang, L.~Stainier, and M.~Ortiz.
\newblock A variational formulation of the coupled thermo-mechanical
  boundary-value problem for general dissipative solids.
\newblock \emph{Journal of the Mechanics and Physics of Solids}, 54\penalty0
  (2):\penalty0 401--424, 2006.
\newblock \doi{10.1016/j.jmps.2005.08.010}.

\bibitem[Kamagat{\'e} et~al.(2025)Kamagat{\'e}, Cheng, Abdelmoula, Danho, and
  Kondo]{kamagate2025}
B.~Kamagat{\'e}, L.~Cheng, R.~Abdelmoula, E.~Danho, and D.~Kondo.
\newblock An incremental variational method to the coupling between gradient
  damage, thermoelasticity and heat conduction.
\newblock \emph{Comptes Rendus. M{\'e}canique}, 353:\penalty0 1063--1084, 2025.
\newblock \doi{10.5802/crmeca.325}.

\bibitem[Georgoulis et~al.(2023)Georgoulis, Loulakis, and
  Tsiourvas]{georgoulis2023}
E.~H. Georgoulis, M.~Loulakis, and A.~Tsiourvas.
\newblock Discrete gradient flow approximations of high dimensional evolution
  partial differential equations via deep neural networks.
\newblock \emph{Communications in Nonlinear Science and Numerical Simulation},
  117:\penalty0 106893, 2023.
\newblock \doi{10.1016/j.cnsns.2022.106893}.

\bibitem[Park et~al.(2023)Park, Kim, Son, and Hwang]{park2023}
M.~S. Park, C.~Kim, H.~Son, and H.~J. Hwang.
\newblock The deep minimizing movement scheme.
\newblock \emph{Journal of Computational Physics}, 494:\penalty0 112518, 2023.
\newblock \doi{10.1016/j.jcp.2023.112518}.

\bibitem[Huang et~al.(2025)Huang, Yu, Xu, Wang, and Wu]{huang2025gradient}
H.~Huang, H.~Yu, W.~Xu, Q.~Wang, and H.~Wu.
\newblock A thermo-mechanical localizing gradient damage model with modified
  {Mazars} strain.
\newblock \emph{International Journal for Numerical Methods in Engineering},
  126\penalty0 (2):\penalty0 e7652, 2025.
\newblock \doi{10.1002/nme.7652}.

\bibitem[Kingma and Ba(2015)]{kingma2015}
D.~P. Kingma and J.~Ba.
\newblock Adam: a method for stochastic optimization.
\newblock In \emph{International Conference on Learning Representations}, 2015.

\bibitem[Hendrycks and Gimpel(2016)]{hendrycks2016gelu}
D.~Hendrycks and K.~Gimpel.
\newblock Gaussian error linear units ({GELUs}).
\newblock arXiv preprint arXiv:1606.08415, 2016.

\bibitem[Hughes et~al.(2005)Hughes, Cottrell, and Bazilevs]{hughes2005}
T.~J.~R. Hughes, J.~A. Cottrell, and Y.~Bazilevs.
\newblock Isogeometric analysis: {CAD}, finite elements, {NURBS}, exact
  geometry and mesh refinement.
\newblock \emph{Computer Methods in Applied Mechanics and Engineering},
  194\penalty0 (39--41):\penalty0 4135--4195, 2005.
\newblock \doi{10.1016/j.cma.2004.10.008}.

\bibitem[Rivera et~al.(2022)Rivera, Taylor, Omella, and Pardo]{rivera2022}
J.~A. Rivera, J.~M. Taylor, {\'A}.~J. Omella, and D.~Pardo.
\newblock On quadrature rules for solving partial differential equations using
  neural networks.
\newblock \emph{Computer Methods in Applied Mechanics and Engineering},
  393:\penalty0 114710, 2022.
\newblock \doi{10.1016/j.cma.2022.114710}.

\bibitem[Loshchilov and Hutter(2017)]{loshchilov2017}
I.~Loshchilov and F.~Hutter.
\newblock {SGDR}: stochastic gradient descent with warm restarts.
\newblock In \emph{International Conference on Learning Representations}, 2017.

\bibitem[Gustafsson and McBain(2020)]{gustafsson2020scikitfem}
T.~Gustafsson and G.~D. McBain.
\newblock scikit-fem: a {Python} package for finite element assembly.
\newblock \emph{Journal of Open Source Software}, 5\penalty0 (52):\penalty0
  2369, 2020.
\newblock \doi{10.21105/joss.02369}.

\bibitem[Tangella et~al.(2022{\natexlab{b}})Tangella, Kumbhar, and
  Annabattula]{tangella2022code}
R.~G. Tangella, P.~Kumbhar, and R.~K. Annabattula.
\newblock Thermo-elastic fracture using phase field method in {FEniCS}: source
  code for the single-edge notched specimen.
\newblock Mechanics of Materials Group, Indian Institute of Technology Madras,
  2022{\natexlab{b}}.
\newblock
  \url{https://home.iitm.ac.in/ratna/codes/thermo-elastic-fracture-fenics/index.html},
  accessed 2026-10-29.

\bibitem[Becher(1981)]{becher1981}
P.~F. Becher.
\newblock Effect of water bath temperature on the thermal shock of
  {Al$_2$O$_3$}.
\newblock \emph{Journal of the American Ceramic Society}, 64\penalty0
  (1):\penalty0 C17--C18, 1981.
\newblock \doi{10.1111/j.1151-2916.1981.tb09544.x}.

\bibitem[Kim et~al.(1991)Kim, Lee, and Case]{kim1991}
Y.~Kim, W.-J. Lee, and E.~D. Case.
\newblock The measurement of the surface heat transfer coefficient for ceramics
  quenched into a water bath.
\newblock \emph{Materials Science and Engineering: A}, 145\penalty0
  (1):\penalty0 L7--L11, 1991.
\newblock \doi{10.1016/0921-5093(91)90307-9}.

\bibitem[Zhou et~al.(2012)Zhou, Song, Shao, Meng, Jiang, and Li]{zhou2012}
Z.~Zhou, F.~Song, Y.~Shao, S.~Meng, C.~Jiang, and J.~Li.
\newblock Characteristics of the surface heat transfer coefficient for
  {Al$_2$O$_3$} ceramic in water quench.
\newblock \emph{Journal of the European Ceramic Society}, 32\penalty0
  (12):\penalty0 3029--3034, 2012.
\newblock \doi{10.1016/j.jeurceramsoc.2012.04.027}.

\bibitem[Liu et~al.(2015)Liu, Wu, Guo, Jiang, Song, and Li]{liu2015circular}
Y.~Liu, X.~Wu, Q.~Guo, C.~Jiang, F.~Song, and J.~Li.
\newblock Experiments and numerical simulations of thermal shock crack patterns
  in thin circular ceramic specimens.
\newblock \emph{Ceramics International}, 41\penalty0 (1):\penalty0 1107--1114,
  2015.
\newblock \doi{10.1016/j.ceramint.2014.09.036}.

\bibitem[Svolos et~al.(2020)Svolos, Bronkhorst, and Waisman]{svolos2020}
L.~Svolos, C.~A. Bronkhorst, and H.~Waisman.
\newblock Thermal-conductivity degradation across cracks in coupled
  thermo-mechanical systems modeled by the phase-field fracture method.
\newblock \emph{Journal of the Mechanics and Physics of Solids}, 137:\penalty0
  103861, 2020.
\newblock \doi{10.1016/j.jmps.2019.103861}.

\bibitem[{NCI Australia}(2019)]{gadi}
{NCI Australia}.
\newblock Gadi supercomputer.
\newblock \doi{10.25914/608bfd1838db2}, 2019.

\bibitem[{PVC (Research Infrastructure), UNSW Sydney}(2010)]{katana}
{PVC (Research Infrastructure), UNSW Sydney}.
\newblock Katana.
\newblock \doi{10.26190/669X-A286}, 2010.

\bibitem[Sobol'(1967)]{sobol1967}
I.~M. Sobol'.
\newblock On the distribution of points in a cube and the approximate
  evaluation of integrals.
\newblock \emph{USSR Computational Mathematics and Mathematical Physics},
  7\penalty0 (4):\penalty0 86--112, 1967.
\newblock \doi{10.1016/0041-5553(67)90144-9}.

\end{thebibliography}
